\documentclass{article} 

\usepackage[usenames,dvipsnames,table]{xcolor}
\usepackage[font=footnotesize]{caption}
\usepackage[hyperfootnotes=false]{hyperref}
\usepackage{url}
\usepackage{subcaption}
\usepackage{cuted}
\usepackage{xspace}
\usepackage{pifont}
\usepackage{lipsum}
\usepackage{wrapfig}
\usepackage{overpic}
\usepackage{overpic}          
\usepackage{capt-of}
\usepackage{multirow}
\usepackage{booktabs}
\usepackage{comment}
\usepackage{comment}
\usepackage{amsmath}
\usepackage{mathtools}
\usepackage{empheq}
\usepackage{amssymb}
\usepackage{adjustbox}
\usepackage[capitalize,noabbrev]{cleveref}
\usepackage{soul,color}
\usepackage{enumitem} 
\usepackage{changepage}
\usepackage{algorithm}
\usepackage[noend]{algorithmic}

\renewcommand{\algorithmiccomment}[1]{\bgroup\hfill $\triangleright$ ~#1\egroup}

\newcommand{\orthoadamcolor}{Salmon}

\newcommand{\orthoadamtext}[1]{\colorbox{\orthoadamcolor}{#1}}

\def\ie{\emph{i.e.}\@\xspace}
\def\eg{\emph{e.g.}\@\xspace}

\newcommand{\expect}{\mathbb{E}}
\newcommand{\real}{\mathbb{R}}
\newcommand{\var}{\text{Var}}

\usepackage{blindtext}

\renewcommand{\paragraph}[1]{\vspace{1em}\noindent\textbf{#1}.}

\newcommand{\xmark}{\ding{55}}%

\newcommand{\ismail}[1]{\textcolor{magenta}{[Ismail: #1]}}
\newcommand{\prannay}[1]{\textcolor{blue}{[Prannay: #1]}}

\newcommand{\mailtodomainpk}[1]{\href{mailto:#1@gmail.com}{\nolinkurl{#1}}}
\newcommand{\mailtodomainhw}[1]{\href{mailto:#1@huawei.com}{\nolinkurl{#1@huawei.com}}}

\newcommand{\parb}[1]{\par{\noindent \bf #1}}
\usepackage{iclr2025_conference,times}

\usepackage{amsmath,amsfonts,bm}

\def\eqref#1{equation~\ref{#1}}

\def\1{\bm{1}}

\DeclareMathAlphabet{\mathsfit}{\encodingdefault}{\sfdefault}{m}{sl}
\SetMathAlphabet{\mathsfit}{bold}{\encodingdefault}{\sfdefault}{bx}{n}

\newcommand{\R}{\mathbb{R}}

\title{From Attention to Activation: Unravelling the Enigmas of Large Language Models}

\author{Prannay Kaul\textsuperscript{1}\thanks{Work conducted during internship} \quad
Chengcheng Ma\textsuperscript{2} \quad
Ismail Elezi\textsuperscript{1}\thanks{Correspondence to \mailtodomainhw{ismail.elezi}} \quad
Jiankang Deng\textsuperscript{1} \\
\textsuperscript{1}Huawei Noah’s Ark Lab, London, UK \\
\textsuperscript{2}Institute of Automation, Chinese Academy of Sciences (CASIA) \\
}

\iclrfinalcopy 
\begin{document}

\maketitle
\vspace{-0.5cm}
\begin{figure}[h!]
    \centering
    {\scriptsize \textsc{Current Transformer Models}}\par\medskip
    \vspace{-0.2cm}
    \footnotesize
    \begin{minipage}[t]{0.24\textwidth}
        \includegraphics[width=\textwidth]{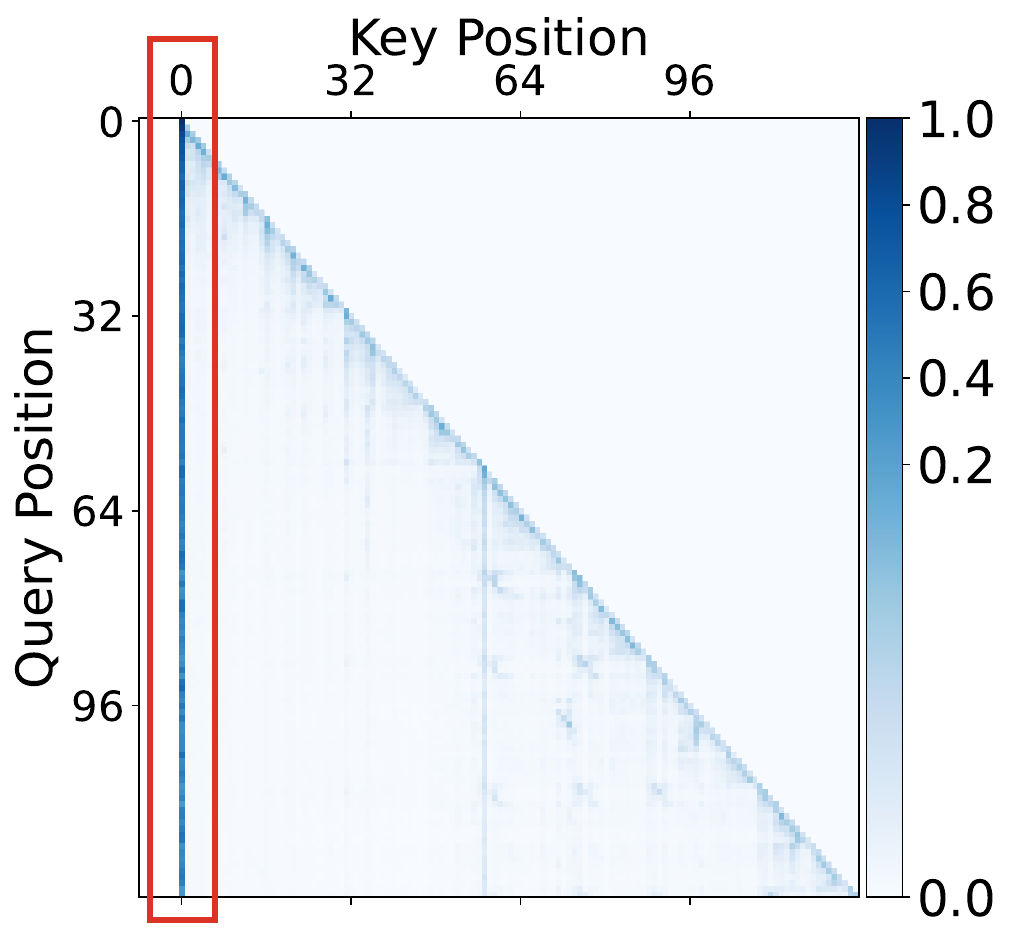}
    \end{minipage}
    \hfill
    \begin{minipage}[t]{0.75\textwidth}
        \includegraphics[width=\textwidth]{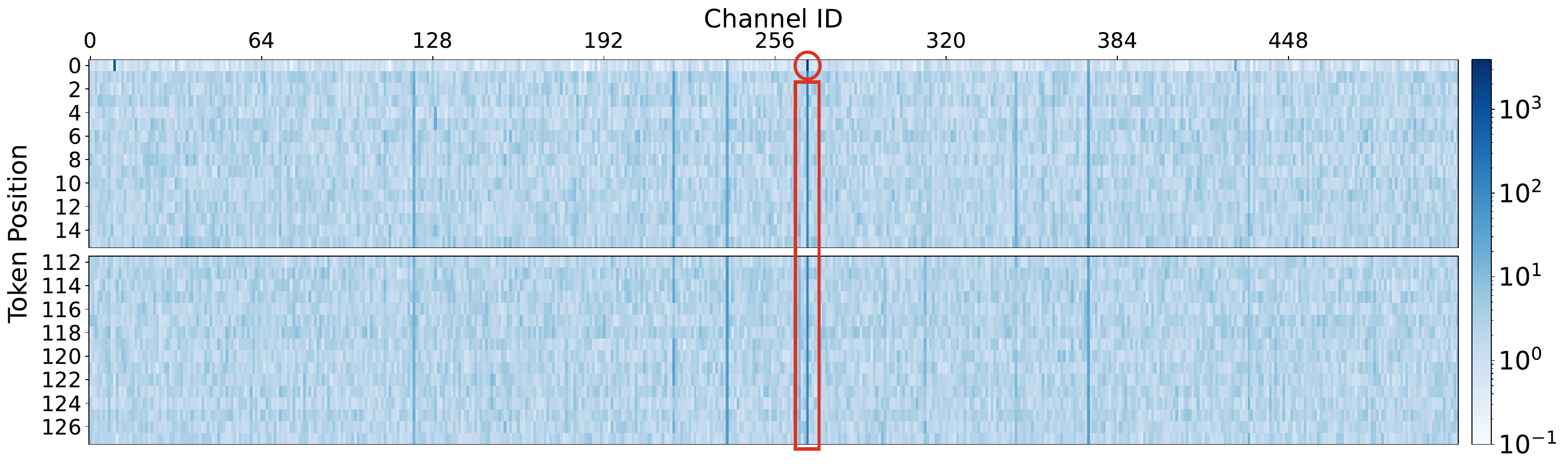}
    \end{minipage}
    \vskip -0.2cm
    {\scriptsize \textsc{\textbf{Our} Transformer Models}}\par
    \centering
    \footnotesize
    \begin{minipage}[t]{0.24\textwidth}
        \includegraphics[width=\textwidth]{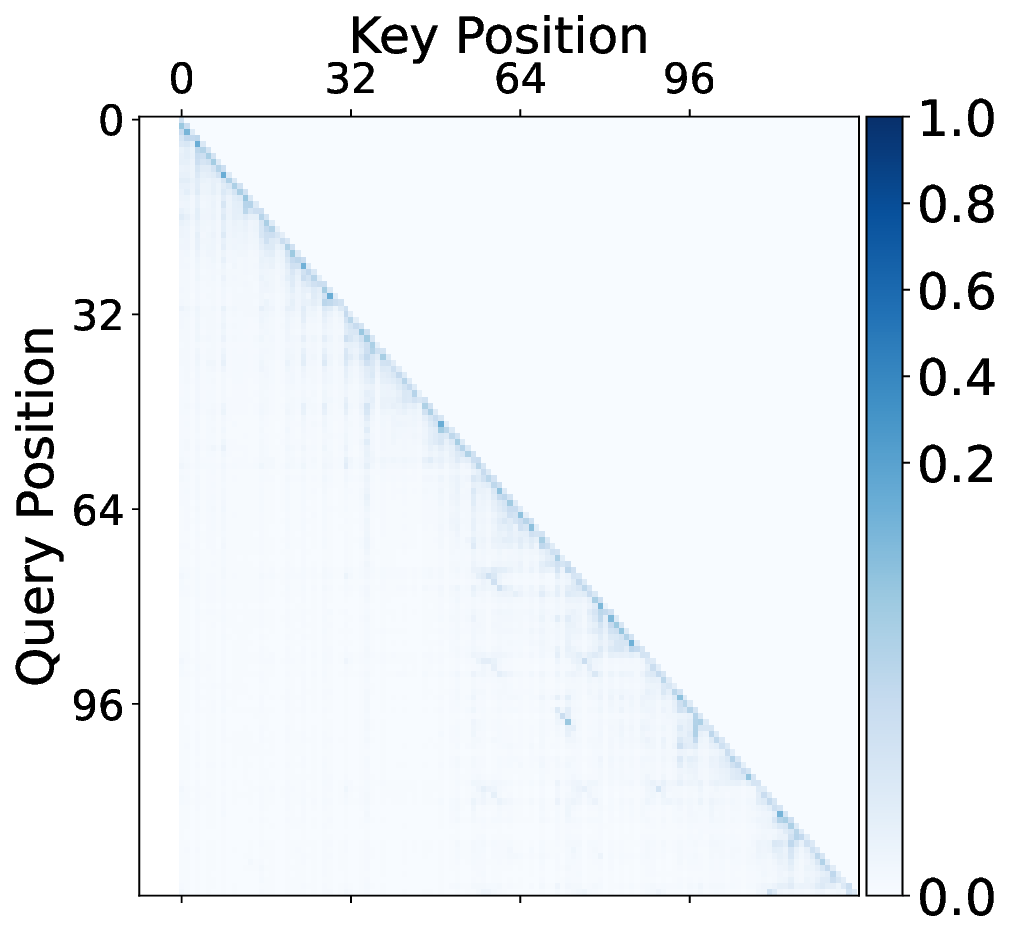}
        \vspace{-0.6cm}
        \subcaption{}\label{fig:teaser_attn}
    \end{minipage}
    \hfill
    \begin{minipage}[t]{0.75\textwidth}
        \includegraphics[width=\textwidth]{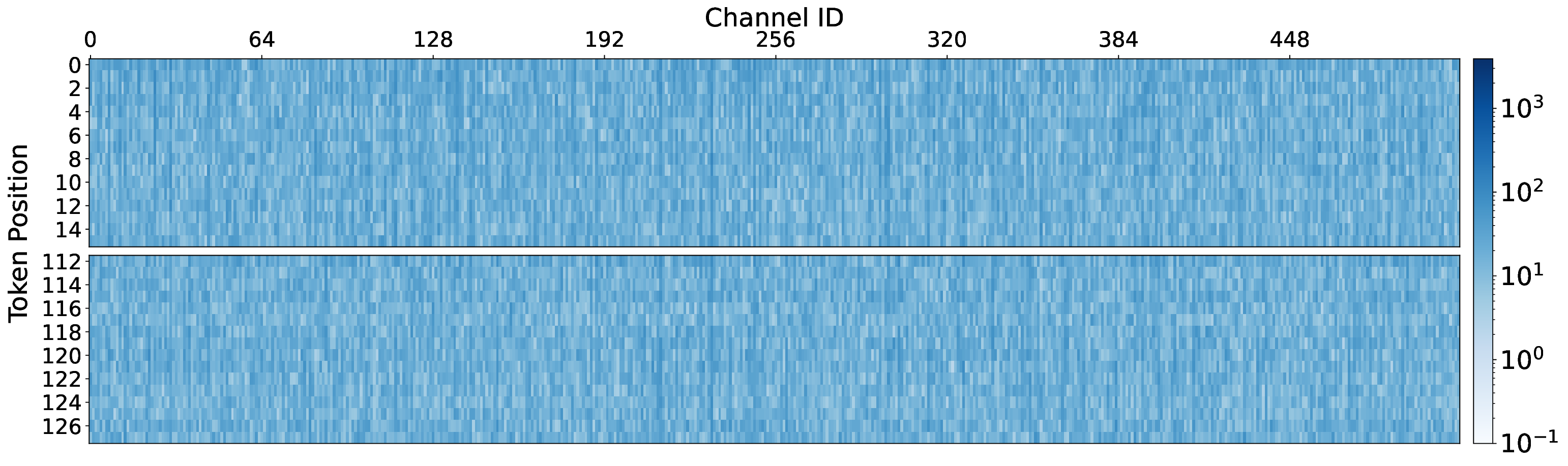}
        \vspace{-0.6cm}
        \subcaption{}\label{fig:teaser_hs}
    \end{minipage}
    \vspace{-0.45cm}
    \caption{
        \textbf{(top)} \textbf{(a)} The mean attention map across all heads and layers of a GPT2-Medium
        model---the first token strangely dominates attention (\textcolor{red}{boxed in red}).
        \textbf{(b)} The mean hidden state across layers of the same model---outlier activations emerge in specific feature dimensions (\textcolor{red}{boxed in red}).
        The first token position exhibits the most extreme outlier activations---(\textcolor{red}{circled in red}).
        \textbf{(bottom)} \textbf{(a)} Replacing the canonical softmax function with our proposed \emph{softmax-1} function eliminates the first token dominance.
        \textbf{(b)} Using our proposed optimiser, \emph{OrthoAdam}, removes outlier activations \emph{without any reduction in model performance}.
    }
    \label{fig:teaser}
\end{figure}

\begin{abstract}
\vspace{-0.1cm}
We study two strange phenomena in auto-regressive Transformers:
(1) the dominance of the first token in attention heads; 
(2) the occurrence of large outlier activations in the hidden states.
We find that popular large language models,
such as Llama attend maximally to the first token in 98\% of attention heads,
a behaviour we attribute to the softmax function. 
To mitigate this issue,
we propose a reformulation of softmax to \emph{softmax-1}.
Furthermore, we identify adaptive optimisers,
\eg~Adam, as the primary contributor to the large outlier activations and
introduce \emph{OrthoAdam}, a novel optimiser that utilises orthogonal matrices to transform gradients,
to address this issue.
Finally, not only do our methods prevent these phenomena from occurring,
but additionally, they enable Transformers to sustain their performance when quantised using basic algorithms,
something that standard methods are unable to do.
In summary, our methods reduce the attention proportion on the first token from $65$\% to $3.3$\%,
the activation kurtosis in the hidden states from $1657$ to $3.1$,
and perplexity penalty under 4-bit weight quantisation from $3565$ to $0.3$.
\end{abstract}
\vspace{-0.2cm}
\section{Introduction}\label{sec:intro}
\vspace{-0.1cm}

Transformers have revolutionised machine learning, achieving state-of-the-art performance across diverse domains,
including natural language processing, computer vision and even protein structure prediction~\citep{openai2023chat,carion2020end,jumper2021highly}.
However, the inner workings of auto-regressive Transformers remain enigmatic.
Recent studies~\citep{elhage2022toy,olsson2022context,bansal2023rethinking} unravelled some of their complexities,
yet our research reveals two surprising phenomena remain pervasive:
\begin{enumerate}[noitemsep,topsep=0pt,leftmargin=15pt]
    \item The strong, consistent dominance of the first token in attention maps---see top of~\cref{fig:teaser_attn}.
    \item The presence of outlier activation values, across sequence position, in specific feature channels of the hidden states
    (the intermediate features of each layer \emph{after} the residual connection)
    that are orders of magnitude larger than other values---see top of~\cref{fig:teaser_hs}.
\end{enumerate}
We ask: What causes these phenomena? Are they essential to performant models? And, if not, how can we mitigate them?

These two phenomena are aesthetically curious, but also have important practical implications.
For instance,
Llama models~\citep{touvron2023llama,dubey2024llama} exhibit the aforementioned first token dominance of attention,
and so requiring complicated attention masking schemes to extend Llama models to tasks with
long sequences~\citep{xiao2024efficient}
\ie~increase the maximum context length used during training.
This is particularly crucial for instruction-tuned models where long conversations are desirable~\citep{wei2022finetuned,ouyang2022training}.
Similarly, the presence of outlier activations leads to challenges in quantising large language models (LLMs).
Large outlier activations increase the required quantisation range (to capture the outliers),
resulting in low effective bits for the non-outlier activations, causing severe performance degradation post-quantisation.
To address this issue, prior work has proposed mixed-precision decomposition of LLMs~\citep{dettmers2022llmint}
or complex scaling of the weights and activations which must be learnt for each model~\citep{xiao2023smoothquant}.
Therefore, our additional motivation is to understand and mitigate these phenomena in a general manner,
such that these issues are resolved \emph{during training}.

We begin by examining the attention mechanism, and surprisingly find, across numerous input sequences,
query tokens attend \emph{most} to the first key token up to 98\% of the time.
This is striking considering the limited semantic information the first token typically
contains---it is often a special token indicating the start of a sequence, such as \texttt{<bos>}.
We explore explanations for this, ruling out positional encodings, non-linearity choice, or feature normalisation.
Ultimately, we identify the softmax function used in the attention mechanism combined with causal masking as the root
cause---excessive attention on the first key token demonstrates an attention head effectively doing nothing~\citep{bondarenko2023removing,clark2019bert}.
The first token is privileged due to causal masking; it is the only key token to which all query tokens can attend.
We propose a straightforward adjustment to softmax as a solution, \emph{softmax-1},
which removes first token dominance in attention~(bottom of \cref{fig:teaser_attn}).

\begin{wraptable}{r}{0.4\textwidth}
    \vspace{-0.5cm}
    \centering
    \footnotesize
    \begin{adjustbox}{width=\linewidth}
    \begin{tabular}{l|c|cc}
    \toprule
    \multirow{2}{*}{Model} & \multirow{2}{*}{\#Parameters} & \multicolumn{2}{c}{PPL}   \\
                           &                               & FP16    & 4-bit Quant     \\ \midrule
    GPT2-Small             & 137M                          & 37.8    &    4456.1       \\
    GPT2-Medium            & 350M                          & 28.8    &    2435.3       \\
    GPT2-Large             & 812M                          & 25.2    &     571.0       \\
    GPT2-XL                & 1.6B                          & 23.2    &    7981.8       \\ \midrule
    Llama2-7B              & 6.7B                          &  7.7    &  191477.5       \\
    Llama3.1-8B            & 8B                            & 10.2    & 2087638.0       \\ \midrule
    \rowcolor[HTML]{C0C0C0}
    GPT2 (Ours)            & 350M                          & 16.3    &     17.1        \\ 
    \rowcolor[HTML]{C0C0C0}
    GPT2 (Ours)            & 1.4B                          & 13.3    &     13.6        \\ \bottomrule
    \end{tabular}
    \end{adjustbox}
    \vspace{-0.3cm}
    \caption{
        Due to surprising phenomena in Transformer models,
        basic zeropoint 4-bit weight quantisation leads to catastrophic performance degradation.
        Our models trained with \emph{softmax-1} and \emph{OrthoAdam} exhibit improved
        robustness to quantisation.
    }\label{tab:teaser}
    \vspace{-0.5cm}
\end{wraptable}

Despite removing first token dominance in attention, using \emph{softmax-1}, we find that the problem of
outlier activations in the hidden states persists.
Once again, we investigate potential causes of this issue and discover the outliers are
primarily caused by the use of adaptive optimisers, \eg~Adam~\citep{kingma2015adam}.
Specifically, our experiments show the exponential decaying averages of first and second moments of gradients result in outlier
activations.
To tackle this, we propose a novel optimiser, \emph{OrthoAdam}, which transforms computed gradients using orthogonal matrices,
thus storing gradients in an alternative basis to the model parameters.
Our results demonstrate this optimiser eliminates the outliers in the hidden states of Transformers~(bottom of \cref{fig:teaser_hs}).

Our research extends beyond aesthetic curiosities.
While LLMs perform well despite first token dominance and outlier activations,
they lead to practical challenges.
Although advanced schemes have been developed to enable quantised LLMs to maintain their performance,
we show our approach enables LLMs to maintain their performance with the most basic quantisation methods,
such as per-tensor 8-bit \emph{absmax} weight/activation quantisation and 4-bit \emph{zeropoint} weight quantisation.
Thus, while our investigation began to better understand Transformers,
our methods offer additional practical benefits.


In summary, our \textbf{contributions} are as follows:
\begin{itemize}[noitemsep,topsep=0pt,leftmargin=15pt]
    \setlength\itemsep{4pt}
    {\item We \textbf{identify} the dominance of the first token in attention and the occurrence of outliers in the
     activations of the hidden states as significant issues in auto-regressive Transformers.}
    
    {\item We \textbf{propose} two simple, effective solutions: a reformulation of the softmax function, \emph{softmax-1},
    to address the former issue, and a novel optimiser, \emph{OrthoAdam}, to tackle the latter.
    Our methods reduce first token attention from $65$\% to $3.3$\% and activation kurtosis from $1657$ to $3.1$.}

    {\item We \textbf{demonstrate} that these proposals not only resolve the identified problems but also lead
     to practical improvements in the performance of Transformers under 8-bit weight/activation and 4-bit weight quantisation.
     Our method reduces the perplexity penalty under 4-bit weight quantisation from $3565$ to $0.3$.}
\end{itemize}
\section{Problem Definition}\label{sec:pd}

This work investigates the two most prominent and strange phenomena of auto-regressive Transformer models:
(1) strong, consistent dominance of the first token in the attention maps;
(2) strong, consistent outlier activations in specific feature channels of the
hidden states (the intermediate features computed \emph{immediately after} the residual connections)---see
top of~\cref{fig:teaser}.
We aim to understand the cause of these phenomena and to propose individual
solutions for each of them.
They have been investigated or commented on previously~\citep{bondarenko2023removing,dettmers2022llmint,xiao2023smoothquant},
but our work reaches different conclusions on the causes and suggests
novel solutions.
We start by describing these two anomalies in detail.

\subsection{First Token Dominance in Attention Maps}\label{ssec:pd-first-token}
The top of~\cref{fig:teaser_attn} shows the attention map, averaged across all layers and heads, of a Transformer model,
specifically a pretrained GPT2-Medium model~\citep{radford2019language},
for a single real natural language sequence.
Strangely, in this average attention map the key corresponding to the first token receives the highest attention across all queries.
Quantitatively, we find the first key token is the most attended to key in $76$\% of (query, head) pairs and
receives $52$\% of all attention, when evaluating on the \texttt{en} validation split of the C4 dataset~\citep{raffel2020exploring,dodge2021documenting}.
This behaviour is consistent across different LLMs, including
the Llama series~\citep{touvron2023llama,dubey2024llama}, DeepSeek~\citep{liu2024deepseek}, and the GPT2 series~\citep{radford2019language}.
See~\cref{sec:pretrained_attention} for detailed examples of attention maps for these models.

Attention is a key component of the Transformer architecture,
and work on the interpretability of LLMs often focuses on
analysing attention~\citep{elhage2021mathematical}.
Moreover, many models, such as Llama2, use a special token for the beginning of a sequence (the \texttt{<bos>} token),
which is \emph{always} the first token in an input sequence.
This makes first token dominance particularly
puzzling, as such models should learn the initial input structure easily.
We hypothesise that this phenomenon in the attention mechanism is a symptom of
a fundamental problem in the Transformer architecture and is not necessary for a performant auto-regressive Transformer.

\subsection{Outlier Activations in the Hidden States}\label{ssec:pd-outliers}
The top of~\cref{fig:teaser_hs} shows the activation magnitude in the hidden states of a pretrained GPT2-Medium model.
We observe the hidden states of the Transformer model exhibit consistent outlier activations
in specific feature channels across all token positions (boxed red),
with the most extreme outliers occurring in the first token position (circled red).
Once again, this behaviour is consistent across different LLMs
and is invariant to the input sequence \ie~the same feature channels \emph{always} exhibit outlier activations.
See~\cref{sec:pretrained_hidden} for examples of hidden states in pretrained models.   

From a practical perspective, these outlier activations are problematic with regards 
to quantising models for deployment~\citep{lin2021towards,dettmers2022llmint}.
However, from a theoretical perspective, the cause of these outlier activations is not well understood.
Previous works, have suggested these outliers are related to
first token domination in attention maps~\citep{xiao2023smoothquant,bondarenko2023removing}.
This is plausible for the most extreme outliers observed in the first token position,
but it does not explain the outlier activations observed across all token positions.
In this work, we show the two phenomena are unrelated and separate solutions are required
to address each.

\section{Method: First Token Dominance of Attention Maps}\label{sec:meth-first-token}

We start by eliminating plausible causes of the first phenomenon of interest: first token dominance of attention maps.
We mainly consider GPT2 as a representative auto-regressive Transformer, because of its simplicity,
but also consider the more recent Llama2 model to narrow down possible causes of this phenomenon.
For all experiments, unless mentioned otherwise, we use a GPT2 model with 130M parameters,
trained on the \texttt{en} split of the C4 dataset.

\subsection{Eliminating Certain Causes of First Token Dominance of Attention Maps}

Both GPT2 and Llama exhibit first token dominance in attention maps.
Thus, we can rule out parts of their architecture that are different:
\begin{itemize}[noitemsep,topsep=0pt,leftmargin=15pt]
    \setlength\itemsep{0pt} 
    \item \textit{Positional encoding.} Llama models use Rotary Positional Encodings (RoPE)~\citep{su2024roformer}, while GPT2 models uses learnt absolute positional encodings~\citep{vaswani2017attention}.
    \item \textit{Initial token.} Llama models use a \texttt{<bos>} token to denote the beginning of a sequence, while GPT2 models do not.
    \item \textit{Activation function.} Llama models use SiLU~\citep{elfwing2018sigmoid} in the feedforward layers, while GPT2 models use GeLU~\citep{hendrycks2016gaussian}.
    \item \textit{Feature Normalisation.} Llama models use RMSNorm~\citep{zhang2019root}, while GPT2 models use LayerNorm~\citep{ba2016layer}.
\end{itemize}

Note that Llama and GPT2 use different positional encoding,
but it is possible that any form of positional encoding might be cause of first token dominance.
To test this possibility,
we train a GPT2 model \textit{without any positional encodings} and observe the attention maps.
We find equivalently trained GPT2 models with/without positional encodings exhibit first token dominance in $33$\%/$20$\% of (query, head) pairs and
allocate $17$\%/$10$\% of all attention to the first token.
Thus, we conclude that positional encodings are not the cause of these anomalies.
The models mentioned here are trained for relatively few steps and first token dominance
is more pronounced in our longer-trained models and in publicly available pretrained models.

\subsection{Removing First Token Dominance of Attention Maps}
After eliminating the above causes,
we have two aspects of Transformers that could cause first token dominance:
(1) causal masking in self-attention; and (2) softmax normalisation in attention heads.

Consider the self-attention mechanism on the initial token in a causal Transformer.
The first \emph{query} token can only attend to its own key token and therefore it in receives an attention score of $1$,
due to softmax normalisation.
Similarly, the second query can only attend to the first two keys, whose attention scores must sum to $1$.
Prior work establishes attention heads specialise to concepts or concept groups~\citep{bansal2023rethinking,elhage2022toy}.
However, given a query irrelevant to the specialisation of an attention head,
it must still allocate attention across the keys summing up to $1$.
Moreover, causal masking privileges the first key token above all others;
it is the only key token to which \textit{all} tokens can attend.
This explains why the \textit{first} token specifically dominates attention maps.

Clearly, a particular attention head should be able to \textit{attend nowhere} if no relevant information is present.
Thus, we modify the softmax function to the following:
\begin{align}
    \text{softmax-1}(x_i) &= \frac{\exp(x_i)}{1 + \sum_{j=1}^{L} \exp(x_j)}; \quad \quad
    \sum_{i=1}^{L} \text{softmax-1}(x_i) < 1
\end{align}
This modification removes the strict enforcement of attention scores summing to $1$,
allowing the model to allocate attention as it sees fit,
including having low attention scores everywhere.
From a registers/attention sink perspective~\citep{darcet2024vision,xiao2024efficient},
the $1$ in the denominator is equivalent to a register/attention sink key token
which has $0$ \emph{dot product} with any query token.

\textbf{Validating the hypothesis.}
We train two GPT2 models, one with canonical softmax and one with softmax-1,
keeping all other variables the same.
The model trained with canonical softmax attention exhibits first token dominance;
the first key token is the most attended to key in $53$\% of (query, head) pairs.
However,
the model trained with softmax-1 lowers this to just $2$\%.
Furthermore, with canonical softmax $46$\% of all attention is received by the first key, while using softmax-1
lowers this to $4\%$, thereby validating our idea.
\begin{wrapfigure}{r}[0pt]{0.3\textwidth}
    \vspace{-0.5cm}
    \centering
    \footnotesize
    \includegraphics[width=\linewidth]{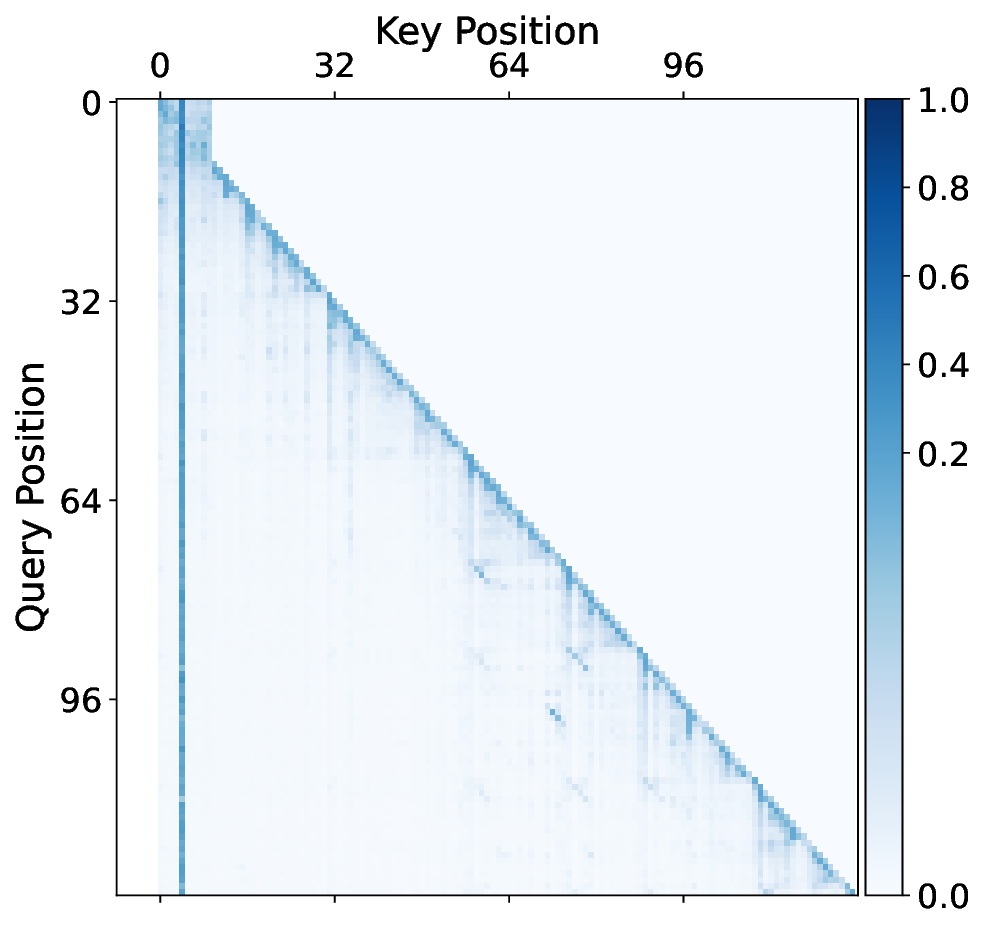}
    \vspace{-0.7cm}
    \caption{Relaxing causal masking leads to attention domination by a different token}\label{fig:causal}
\end{wrapfigure}
The difference in attention maps between canonical softmax and softmax-1 is shown in~\cref{fig:teaser_attn},
which compares the attention maps of two models on the same input sequence.
Furthermore, we find using softmax-1 has no effect on training stability, convergence or model performance
(see~\cref{sec:loss_curves} for the training curves of all our trained models).

\textbf{What if causal masking is relaxed?}
To verify the first token is privileged by causal masking, causing \emph{first} token dominance,
we train a GPT2 model with canonical softmax in which
causal masking is removed for the first 10 tokens (the loss function is appropriately modified).
\Cref{fig:causal} shows a specific token still dominating the attention map,
but it is the \emph{fifth} one, not the first.
\section{Method: Outlier Activations}\label{sec:oa}

To quantitatively establish the extent of outliers in the hidden states,
we use kurtosis.
Kurtosis, in this case, is a measure of tail heaviness of a set of activation values.
Activations which are normally distributed have a kurtosis of $\sim$$3$,
while higher kurtosis indicates a heavier-tailed distribution~(\eg~the exponential distribution)
and lower kurtosis indicates a lighter-tailed distribution~(\eg~the uniform distribution).
Given hidden states $\mathbf{X} \in \mathbb{R}^{M \times L \times D}$ of a Transformer model,
where $M$ is the number of layers, $L$ is the number of tokens and $D$ is the number of feature channels,
we compute the per-layer, per-position kurtosis of the hidden states as:
\begin{align}
    \kappa_{m,l} = \text{Kurt}_{m,l}\left[X_{m,l,d}\right] = \frac{\expect_{d}[(X_{m,l,d} - \mu_{m,l})^4]}{\expect_{d}[(X_{m,l,d} - \mu_{m,l})^2]^2},
        \quad \text{where} \quad \mu_{m,l} = \expect_{d}[\mathbf{X}_{m,l,d}] 
\end{align}
where $X_{m,l,d}$ is the hidden state at layer $m$ at position $l$ for feature $d$,
and $\mu_{m,l}$ is the mean hidden state value at layer $m$ at position $l$. 

\subsection{Eliminating Certain Causes of Outlier Activations}
We start by eliminating certain causes which could lead to the presence of outlier activations.

\textbf{Feedforward Layer Biases.} GPT2 uses biases in all feedforward layers,
while Llama uses none, therefore it is unlikely feedforward layer biases cause of outlier activations.

\textbf{Normalisation Layers.} GPT2 uses LayerNorm~\citep{ba2016layer} while LLama uses RMSNorm~\citep{zhang2019root},
which both learn individual scaling parameters for each feature channel, potentially causing the outlier activations.
To remove such an effect,
we replace LayerNorm in our trained GPT2 models with an RMSNorm version which applies a \emph{single} global scale instead of per-channel scaling,
and call it ``RMSNormSingle''---similar to ``Simple RMSNorm'' from \citet{qin2023scaling} which has no learned parameters.
We find outlier activations persist in the hidden states of a GPT2 model with RMSNormSingle.
In \Cref{tab:ablation_big} we show kurtosis remains high in models trained without biases and/or with RMSNormSingle.

\textbf{Optimiser.}
Most Transformer models are trained with Adam~\citep{kingma2015adam} or a variant.
These optimisers track the first and second moments of the computed gradients using exponential moving averages,
tracking these moments at a parameter level.
The main hyperparameters of Adam-like optimisers are $\beta_1$ and $\beta_2$,
which control the decay rates of the first and second moments, respectively.
If $\beta_2 = 0$, only the first moment of the gradients is tracked, resembling stochastic gradient descent (SGD) with momentum.
Conversely, if $\beta_1 = 0$, only the second moment of the gradients is tracked, resembling RMSProp.
We suspect that given the optimiser tracks moments in the same basis as the model parameters,
it is the most likely cause of the outlier activations in the hidden states auto-regressive Transformer models.

\textbf{Validating the hypothesis.}
We train a series of GPT2 models using Adam, RMSProp, SGD with and without momentum, tuning the learning rate and training schedule to encourage convergence.
The model trained with SGD has the slowest convergence and highest validation perplexity,
while the model trained with Adam converges the fastest and has the lowest perplexity.
However, we find models trained with Adam and RMSProp have high kurtosis, $140$ and $70$, respectively,
while training with SGD gives a kurtosis of $\sim$3.0.
We provide these results in our ablation study (\cref{ssec:ablation}).

\begin{wrapfigure}{r}[0pt]{0.2\textwidth}
    \vspace{-0.2cm}
    \centering
    \footnotesize
    \includegraphics[width=\linewidth]{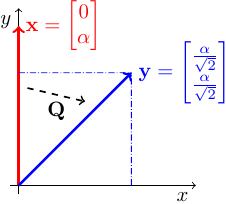}
    \includegraphics[width=\linewidth]{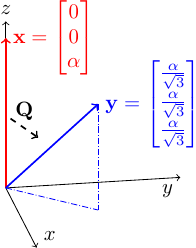}
    \vspace{-0.7cm}
    \caption{Rotating vectors with dominant components leads to a reduction in the maximum absolute value.}\label{fig:rotations}
\end{wrapfigure}
\subsection{OrthoAdam}\label{ssec:orthoadam}
The previous section leaves an important question for training Transformer models:
\textit{``How can we train a model with an optimiser which has the speed and convergence properties of Adam,
but produces activations properties similar to SGD''?}

Optimisers which track exponential decaying averages of the first and/or second moments
of the gradients lead to outlier activations in the hidden states of Transformer models.
Moreover, in the models trained above,
the largest absolute \emph{parameter} values correspond to the features which exhibit outlier activations
in the hidden states, \ie~if outlier activations occur in feature channel $i$ of the hidden states,
the largest model parameter values correspond to specific weights which act on
feature channel $i$ of the hidden states \eg~the $i$\textsuperscript{th} output channel of the output projection weights of the attention/MLP layers.
Therefore, to arrive at these large model parameter values, the optimiser (\eg~Adam) must
provide relatively large updates to these specific parameters and not others.
We note here that Adam and similar optimisers calculate gradient moments in the same basis as the model parameters.
Additionally, given the channels which contain outlier activations appear invariant to the input sequence,
we hypothesise that these channels are an artefact of the optimiser and do not correspond to any meaningful feature in the input
sequence---see~\cref{sec:pretrained_hidden} for plots of the hidden states of pretrained models with different input sequences.
Given these observations, we discuss an idealised case of observed hidden states below,
and show how orthogonal transformations can be used to reduce outlier activations.

Consider a $D$-dimensional vector, $\mathbf{x} = \alpha \mathbf{e}_i + \mathbf{z}$,
where $\mathbf{e}_i$ is the $i$\textsuperscript{th} unit vector in the standard basis,
$\mathbf{x} \in \real^D$, $\alpha \in \real^{+}, \alpha \gg 1$ and $\mathbf{z} \sim \mathcal{N}(\mathbf{0}, \mathbf{I})$.
The first term represents the single outlier activation specific to the $i$\textsuperscript{th} channel
and the second term represents the ``informative'' activations.
The vector $\mathbf{x}$ represents the hidden states of a Transformer model with high kurtosis.
This simplified model makes two assumptions: (1) there is a single outlier activation channel; and
(2) the informative activations are normally distributed.

For values of $D$ similar to that of Transformer models, \ie~$D \approx [10^3, 10^5]$,
$\text{Kurt}[x_j] = O(D)$.
Therefore, we expect larger Transformer models of a given architecture to have larger kurtosis in their hidden states.
Moreover, the ratio of the $\ell_\infty$-norm to the $\ell_2$-norm of the hidden states
in our simplified model, $\frac{\left\lVert\mathbf{x}\right\rVert_{\infty}^2}{\left\lVert\mathbf{x}\right\rVert_{2}^2}$, is close to $1$.
This ratio is another proxy for the extent of outliers.

Now we consider the effect of an appropriate orthogonal transformation on the vector $\mathbf{x}$.
Let $\mathbf{Q} \in \real^{D \times D}$ be an orthogonal matrix, and $\mathbf{y} = \mathbf{Q}\mathbf{x}$.
Under an particular orthogonal transformation,
$\frac{\left\lVert\mathbf{y}\right\rVert_{\infty}^2}{\left\lVert\mathbf{y}\right\rVert_{2}^2} \approx \frac{1}{D}$
and $\text{Kurt}[y_j] = 3$. The orthogonal transformation which achieves this is one which rotates the vector $\mathbf{x}$
such that $\mathbf{Q}\mathbf{e}_i = \frac{1}{\sqrt{D}}\mathbf{1}$.
\Cref{fig:rotations} illustrates this rotation process in 2D and 3D.
The kurtosis and norm ratio results quoted in this section are derived in~\cref{sec:kurtosis} and~\cref{ssec:inf-2norm}, respectively,
and are shown to be empirically valid for models we train from the plots in~\cref{ssec:plots_kurtosis} and~\cref{ssec:linf_l2_ratio}, respectively.

One option is to apply orthogonal transformations directly to the hidden states of the model during
\ie~make $\mathbf{Q}$ part of the model parameters but are kept fixed during training.
Instead, we propose a novel optimizer, \emph{OrthoAdam}, which applies orthogonal transformations to incoming gradients
such that the moment calculations (which our experiments in~\cref{tab:quant_table} show are the key factor in producing outlier activations)
are performed in a different basis to the model parameters to prevent gradient updates to any particular set of parameters which lead to outlier activations.
We provide the full algorithm in~\cref{algo_orthoadam}.

In our experiments, we randomly sample the orthogonal matrix for each parameter (which remains fixed during the training of the model).
We find that using \emph{OrthoAdam} leads to a significant reduction in the kurtosis of hidden states
in Transformer models, effectively eliminating the outlier activations.
This is shown qualitatively at the top of~\cref{fig:teaser_hs}, where feature channels with high absolute activation values in the hidden states
\emph{are no longer present} across all token positions, and quantitatively in~\cref{tab:main_table_small} showing the kurtosis of hidden states
in models trained with \emph{OrthoAdam} is close to $3$, with \emph{no performance penalty}. 

\setlength\fboxsep{1.00pt}
\setlength{\textfloatsep}{0.25cm}
{\begin{algorithm}[t]
\footnotesize
\caption{\footnotesize OrthoAdam, our proposed optimiser for reducing activation outliers. $\mathbf{\bar{g}}^2_t$ is the element-wise square $\mathbf{\bar{g}}_t \odot \mathbf{\bar{g}}_t$.
         With $\beta_1^t$ and $\beta_2^t$ we mean $\beta_1$ and $\beta_2$ taken to the power of $t$.
}\label{algo_orthoadam}
\begin{algorithmic}
\STATE{\textbf{given} learning rate: $\eta=0.001$, first moment decay rate: $\beta_1=0.9$, second moment decay rate: $\beta_2=0.999$,
       numerical epsilon: $\epsilon=10^{-8}$}\label{adam-Given}
\STATE{\textbf{initialise} time step: $t \leftarrow 0$, parameter vector: $\mathbf{\theta}_{t=0} \in \R^n$,  first moment vector: $\mathbf{\bar{m}}_{t=0} \leftarrow \mathbf{0}$,
       second moment vector: $\mathbf{\bar{v}}_{t=0} \leftarrow \mathbf{0}$, schedule multiplier: $\lambda_{t=0} \in \R$,
       \orthoadamtext{unique orthogonal matrix: $\mathbf{Q} \in \mathcal{O}^n$}}\label{adam-Init}
\REPEAT{}
	\STATE{$t \leftarrow t + 1$}
	\STATE{$\nabla f_t(\mathbf{\theta}_{t-1}) \leftarrow  \text{SelectBatch}(\mathbf{\theta}_{t-1})$}\COMMENT{select batch and calculate gradient}
	\STATE{$\mathbf{g}_t \leftarrow \nabla f_t(\mathbf{\theta}_{t-1})$}\COMMENT{store the gradient in model parameter basis}
    \STATE{\orthoadamtext{$\mathbf{\bar{g}_t} \leftarrow \text{MatMul}(\mathbf{Q},\mathbf{g}_t$)}}\COMMENT{transform gradient into unique optimiser basis}
	\STATE{$\mathbf{\bar{m}}_t \leftarrow \beta_1 \mathbf{\bar{m}}_{t-1} + (1 - \beta_1) \mathbf{\bar{g}}_t$}\COMMENT{update biased first moment estimate}\label{adam-mom1} 
	\STATE{$\mathbf{\bar{v}}_t \leftarrow \beta_2 \mathbf{\bar{v}}_{t-1} + (1 - \beta_2) \mathbf{\bar{g}}^2_t$}\COMMENT{update biased second raw moment estimate}\label{adam-mom2} 
	\STATE{$\mathbf{\hat{m}}_t \leftarrow \mathbf{\bar{m}}_t/(1 - \beta_1^t)$}\COMMENT{compute bias-corrected first moment estimate}\label{adam-corr1}
	\STATE{$\mathbf{\hat{v}}_t \leftarrow \mathbf{\bar{v}}_t/(1 - \beta_2^t)$}\COMMENT{compute bias-corrected second raw moment estimate}\label{adam-corr2}
    \STATE{$\mathbf{\bar{s}}_t \leftarrow \hat{\mathbf{m}}_t / (\sqrt{\hat{\mathbf{v}}_t} + \epsilon)$}\COMMENT{calculate the update step in unique optimizer basis}
    \STATE{\orthoadamtext{$\mathbf{s}_t \leftarrow \text{MatMul}(\mathbf{Q}^T,\mathbf{\bar{s}})$}} \COMMENT{transform the update step back to model parameter basis}
	\STATE{$\lambda_t \leftarrow \text{SetScheduleMultiplier}(t)$}	\COMMENT{can be fixed, decay, or also be used for warm restarts}
    \STATE{$\mathbf{\theta}_t \leftarrow \mathbf{\theta}_{t-1} - \lambda_t \eta \mathbf{s}_t$}\COMMENT{apply parameter update}\label{adam-xupdate}
\UNTIL{\emph{stopping criterion is met}}
\RETURN{optimised parameters $\mathbf{\theta}_t$}
\end{algorithmic}
\end{algorithm}
}
\setlength\fboxsep{4pt}
\section{Experiments}\label{sec:exp}
%


\parb{Datasets.}
We train all models on the \texttt{en} training split of the C4 dataset~\citep{dodge2021documenting,raffel2020exploring} and evaluate
on 100000 samples from the validation \texttt{en} split.

\parb{Models.}
For our experiments, we train GPT2 models with $\sim$\{$60$M, $130$M, $350$M, $1.4$B\} parameters
and Llama2 models with $\sim$$130$M parameters.
Apart from changing the softmax function, the only other changes we make to the model architectures are
the use of RMSNormSingle and we do not use biases in feedforward layers.
We ablate these changes in the ablation study at the end of this section.

\parb{Training.}
Unless stated otherwise, we use a batch size of 512 and a cosine learning rate schedule with linear warmup for
\{$1000$, $2000$, $6000$, $10000$\} steps for models with \{$60$M, $130$M, $350$M, $1.4$B\} parameters respectively,
with a maximum learning rate of $10^{-3}$.
We train models with \{$60$M, $130$M, $350$M, $1.4$B\} parameters for \{$160$k, $320$k, $960$k, 600k\} steps respectively.
Note that we use a reduced number of steps for the $1.4$B parameter model due to computational constraints.
In the ablation study, we train GPT2 models with $130$M parameters for $40$k steps only.

\parb{Metrics.}
We evaluate our experiments in the following metrics:
(1) the perplexity (PPL) of models on the validation set;
(2) the mean kurtosis across all layers of the model (evaluated separately for the first token and the remaining tokens);
(3) the maximum absolute activation across all layers of the model (again evaluated separately);
(4) the percentage of (query, head) pairs in which the first key token is the most attended to key token.
We calculate (1) to ensure our method \emph{at least maintains} the vanilla language model performance
\ie~to ensure the model is not harmed by softmax-1 or OrthoAdam.
(2) and (3) show quantitatively the extent to which outlier activations are present in the hidden states.
Finally, (4) shows the extent to which the first token dominates attention in the model.

\subsection{Main Results}\label{ssec:main_results}
We show the results of softmax-1 and OrthoAdam used to train GPT2 and Llama2 models in \Cref{tab:main_table_small}.
We observe that across both model architectures and all sizes,
the evaluated PPL is the same or slightly lower when comparing a model with softmax-1 and
trained with OrthoAdam to the vanilla model with neither,
indicating that our method does not change model performance.
Despite no significant change in PPL,
each of our proposed methods lead to a significant reduction in outlier activations in the hidden states
(shown by a considerably lower mean layer kurtosis
and maximum absolute activation), with the largest reduction observed when both softmax-1 and OrthoAdam are used.
In particular,
for GPT-2 models with $60$M, $130$M, $350$M and $1.4$B parameters,
the kurtosis without our modifications were $77.9$, $141.5$, $161.8$ and $351.0$,
while after our modification they drop to $7$, $7.3$, $3.1$, and $3.0$.
We observe similar results for Llama2-130M where the perplexity is around the same as the original version,
but kurtosis is reduced from $170$ to $6.9$.
Similar to kurtosis, in all cases we see a significant reduction of the mean activation value.
Furthermore, we also observe the drastic drop in first token attention.
While the vanilla versions of the model have maximal first token attention of up to $64.8$\%,
after our modification, it is reduced to $1$-$3$\%.

\begin{table}[t!]
    \centering
    \footnotesize
    \begin{adjustbox}{width=\linewidth}
    \begin{tabular}{c|c|c|c|c|cc|cc|c}
    \toprule
    \multirow{2}{*}{Model}  & \multirow{2}{*}{\#Parameters} & \multirow{2}{*}{Softmax+1?} & \multirow{2}{*}{OrthoAdam?} & \multirow{2}{*}{PPL} & \multicolumn{2}{c|}{Kurtosis}                                                    & \multicolumn{2}{c|}{Activation Value}                                                                 & \multirow{2}{*}{\%First Attn} \\
                            &                               &                             &                             &                      & $\expect_{m}\left[\kappa_{m,1}\right]$ & $\expect_{m}\left[\kappa_{m,>1}\right]$ & $\expect_{m}\left[\left|X_{m,1,d}\right|\right]$ & $\expect_{m}\left[\left|X_{m,>1,d}\right|\right]$ &                               \\ \midrule
    \multirow{12}{*}{GPT2*} & \multirow{4}{*}{60M}          &                             &                             & 31.9                 &  313.8                                 &  77.9                                   &  1856.1                                          &  266.6                                            & 0.489                         \\
                            &                               & \checkmark                  &                             & 31.6                 &  105.6                                 &  81.4                                   &   304.9                                          &  259.0                                            & 0.021                         \\
                            &                               &                             & \checkmark                  & 32.4                 &  260.8                                 &  10.6                                   &  1419.9                                          &  114.7                                            & 0.365                         \\
                            &                               & \checkmark                  & \checkmark                  & 31.8                 &    7.6                                 &   7.0                                   &    92.8                                          &   87.8                                            & 0.019                         \\ \cmidrule(l){2-10} 
                            & \multirow{4}{*}{130M}         &                             &                             & 22.9                 &  514.9                                 & 141.5                                   &  7018.1                                          & 1014.8                                            & 0.527                         \\
                            &                               & \checkmark                  &                             & 22.7                 &  175.4                                 & 144.2                                   &  1134.3                                          &  967.5                                            & 0.024                         \\
                            &                               &                             & \checkmark                  & 23.1                 &  446.4                                 &  20.2                                   &  4285.0                                          &  433.4                                            & 0.424                         \\
                            &                               & \checkmark                  & \checkmark                  & 22.8                 &   10.1                                 &   7.3                                   &   318.1                                          &  261.6                                            & 0.019                         \\ \cmidrule(l){2-10} 
                            & \multirow{2}{*}{350M}         &                             &                             & 16.4                 &  820.3                                 & 161.8                                   & 40196.0                                          & 3801.1                                            & 0.579                         \\
                            &                               & \checkmark                  & \checkmark                  & 16.3                 &    3.1                                 &   3.1                                   &   388.1                                          &  333.3                                            & 0.021                         \\ \cmidrule(l){2-10} 
                            & \multirow{2}{*}{1.4B}         &                             &                             & 13.4                 & 1656.5                                 & 351.9                                   & 56798.3                                          & 7051.2                                            & 0.648                         \\
                            &                               & \checkmark                  & \checkmark                  & 13.3                 &    3.1                                 &   3.0                                   &   181.9                                          &  132.1                                            & 0.033                         \\ \midrule
    \multirow{4}{*}{Llama2} & \multirow{4}{*}{130M}         &                             &                             & 17.4                 &  435.0                                 & 170.0                                   &  4622.7                                          & 1627.4                                            & 0.105                         \\
                            &                               & \checkmark                  &                             & 17.2                 &  208.2                                 & 181.2                                   &  1340.4                                          & 1229.5                                            & 0.016                         \\
                            &                               &                             & \checkmark                  & 17.4                 &  435.8                                 & 169.5                                   &  4685.9                                          & 1629.1                                            & 0.103                         \\
                            &                               & \checkmark                  & \checkmark                  & 17.3                 &    4.2                                 &   6.9                                   &   161.1                                          &  157.0                                            & 0.017                         \\ \bottomrule
    \end{tabular}
    \end{adjustbox}
    \caption{
        Main results showing the impact of \emph{softmax-1} and \emph{OrthoAdam} on trained GPT2 and Llama2 models.
        Utilising \emph{softmax-1} and \emph{OrthoAdam}, significantly reduces the kurtosis and the max activation values of hidden states.
        Using \emph{softmax-1} only is sufficient to reduce first token dominance in attention.
        We generally find that all combinations of \emph{softmax-1} and/or \emph{OrthoAdam} at a given model size lead to similar performance.
        $\expect_{m}\left[\kappa_{m,1}\right]$: mean kurtosis of the first token;
        $\expect_{m}\left[\kappa_{m,>1}\right]$: mean kurtosis of all other tokens;
        $\expect_{m}\left[\left|X_{m,1,d}\right|\right]$: mean max absolute activation value of the first token;
        $\expect_{m}\left[\left|X_{m,>1,d}\right|\right]$: mean max absolute activation value of all other tokens.
        All values are averaged across all layers.
    }\label{tab:main_table_small}
\end{table}

\subsection{Quantisation}\label{ssec:quant}
We quantise trained models
using \emph{Absmax} and \emph{Zeropoint} quantisation.
\emph{Absmax quantisation} scales a given tensor (weight or activation) using the absolute maximum absolute value.
On the other hand,
\emph{Zeropoint quantisation} shifts the quantised tensor such that the minimum tensor value is the minimum representable value.
See~\citet{dettmers2022llmint} for exact details on the quantisation schemes.

\parb{Experimental Setup.}
We quantise the trained models using Absmax quantisation using 8-bit integers and the more powerful
Zeropoint quantisation  using 4-bit integers.
In the case of Absmax quantisation, we use 3 different configurations:
(1) \emph{fine} quantisation, where ``per-channel'' scaling is used for input activations and weights;
(2) \emph{moderate} quantisation, with ``per-tensor'' scaling for input activations and weights;
and (3) \emph{coarse} quantisation, with ``per-tensor'' scaling for input \emph{and output} activations and weights.
In the case of Zeropoint quantisation, we use a single configuration where ``per-channel'' scaling is used for \emph{weights only}.
Note that only linear layers are quantised the embeddings, normalisation layers and softmax functions are not quantised.

\begin{table}[!t]
\centering
\scriptsize
\begin{adjustbox}{width=\textwidth}
\begin{tabular}{c|c|c|ccccccccc}
\toprule
\multirow{2}{*}{Model}  & \multirow{2}{*}{\#Parameters}  & \multirow{2}{*}{OA + S1?} & \multicolumn{9}{c}{PPL}                                                                                                                                                            \\ \cmidrule(l){4-12}
                        &                                &                           & \multicolumn{1}{c|}{full}  & coarse & \multicolumn{1}{c|}{$\Delta$} &  moderate & \multicolumn{1}{c|}{$\Delta$} & fine  & \multicolumn{1}{c|}{$\Delta$} & 4-bit         & $\Delta$ \\ \midrule
\multirow{8}{*}{GPT2}   & \multirow{2}{*}{60M}           & \xmark                    & \multicolumn{1}{c|}{31.88} & 43.53  & \multicolumn{1}{c|}{11.65}    &  34.87    & \multicolumn{1}{c|}{2.99}     & 32.15 & \multicolumn{1}{c|}{0.27}     &     68.5      &     36.6 \\
                        &                                & \checkmark                & \multicolumn{1}{c|}{31.83} & 32.30  & \multicolumn{1}{c|}{ 0.47}    &  32.18    & \multicolumn{1}{c|}{0.35}     & 31.89 & \multicolumn{1}{c|}{0.06}     &     33.9      &      2.1 \\ \cmidrule(l){2-12} 
                        & \multirow{2}{*}{130M}          & \xmark                    & \multicolumn{1}{c|}{22.89} & 46.49  & \multicolumn{1}{c|}{23.60}    &  28.31    & \multicolumn{1}{c|}{5.42}     & 23.07 & \multicolumn{1}{c|}{0.18}     &    679.9      &    657.0 \\
                        &                                & \checkmark                & \multicolumn{1}{c|}{22.78} & 23.21  & \multicolumn{1}{c|}{ 0.43}    &  23.10    & \multicolumn{1}{c|}{0.32}     & 22.83 & \multicolumn{1}{c|}{0.05}     &     24.0      &      1.2 \\ \cmidrule(l){2-12} 
                        & \multirow{2}{*}{350M}          & \xmark                    & \multicolumn{1}{c|}{16.37} & 52.49  & \multicolumn{1}{c|}{36.12}    &  19.92    & \multicolumn{1}{c|}{3.55}     & 16.50 & \multicolumn{1}{c|}{0.13}     & 118507.1      & 118490.7 \\
                        &                                & \checkmark                & \multicolumn{1}{c|}{16.31} & 16.50  & \multicolumn{1}{c|}{ 0.19}    &  16.46    & \multicolumn{1}{c|}{0.15}     & 16.33 & \multicolumn{1}{c|}{0.02}     &     17.1      &      0.8 \\ \cmidrule(l){2-12} 
                        & \multirow{2}{*}{1.4B}          & \xmark                    & \multicolumn{1}{c|}{13.44} & 45.05  & \multicolumn{1}{c|}{31.61}    &  15.19    & \multicolumn{1}{c|}{1.75}     & 13.68 & \multicolumn{1}{c|}{0.24}     &   3577.7      &   3564.3 \\
                        &                                & \checkmark                & \multicolumn{1}{c|}{13.33} & 13.45  & \multicolumn{1}{c|}{ 0.12}    &  13.43    & \multicolumn{1}{c|}{0.10}     & 13.34 & \multicolumn{1}{c|}{0.01}     &     13.6      &      0.2 \\ \midrule
\multirow{2}{*}{Llama2} & \multirow{2}{*}{130M}          & \xmark                    & \multicolumn{1}{c|}{17.39} & 43.61  & \multicolumn{1}{c|}{26.22}    &  24.46    & \multicolumn{1}{c|}{7.07}     & 17.69 & \multicolumn{1}{c|}{0.30}     &     21.5      &      4.1 \\
                        &                                & \checkmark                & \multicolumn{1}{c|}{17.31} & 20.85  & \multicolumn{1}{c|}{ 3.54}    &  20.11    & \multicolumn{1}{c|}{2.80}     & 17.38 & \multicolumn{1}{c|}{0.07}     &     19.7      &      2.4 \\ \bottomrule
\end{tabular}
\end{adjustbox}
\caption{
    Performance of our trained models under various quantisation settings.
    When using \emph{OrthoAdam} and \emph{softmax-1} (OA + S1), the performance penalty due to quantisation is significantly reduced.
    The benefits of our proposed changes are more pronounced under more aggressive quantisation
    settings---\ie~4-bit weight and coarse 8-bit weight/activation quantisation (vanilla models exhibit catastrophic performance degradation).
}\label{tab:quant_table}
\end{table}

\parb{Results.}
In \Cref{tab:quant_table} we show the results of quantising the trained models using Absmax 
and Zeropoint quantisation.
We experimentally confirm that in all cases, models trained with softmax-1 and OrthoAdam are more robust to Absmax quantisation
schemes than models trained with the canonical softmax function and Adam.
The difference in performance is most pronounced when using moderate and coarse quantisation schemes---models
trained with softmax-1 and OrthoAdam are able to maintain performance while models trained with canonical
softmax and Adam suffer a significant degradation in performance.
In particular, in the coarse setting, our method outperforms the baseline by up to $36.12$ points.
For Zeropoint quantisation, we observe that all GPT2 models trained with canonical softmax and Adam
become unusable when using 4-bit integer weight quantisation,
while models trained with softmax-1 and OrthoAdam suffer only a small drop in performance.
Llama2 models in both cases remain usable after quantisation, but the performance drop
is more pronounced when using the canonical softmax function and Adam.

\subsection{Ablation Study}\label{ssec:ablation}

\Cref{tab:ablation_big} shows the results of an ablation study on GPT2 models with $130$M parameters.
As expected from the discussion in~\cref{sec:meth-first-token},
we find removing biases from linear layers and varying the position encodings does
not prevent first token domination---we see a small reduction in first token domination when positional
encodings are removed.

When using softmax-1, we find first token dominance is mitigated with only $\sim$2\% of (query, head) pairs
having the first key token as the most attended to key token.

Switching from LayerNorm to RMSNorm with a learnt scale for each channel (RMSNorm-M, the normalisation used in Llama2) does not reduce
the prevalence of outlier activations in the hidden states.
However, switching to RMSNorm with a single learnt scale (RMSNorm-S) reduces the mean layer kurtosis and max absolute activation by $\sim$$40$\%,
but still remain high.
In all of the above cases in which Adam is used as the optimiser, we observe similar perplexity to the initial model (top row).
Slight exceptions being the use of rotary and no positional encodings,
in which perplexity reduces and increases by $1.3$ and $0.5$, respectively.

Changing the optimiser to RMSProp leads to increased perplexity ($0.5$ compared to the initial model),
reduced mean layer kurtosis and max absolute activation, by $\sim$$50$\% and $\sim$$30$\%, respectively,
when comparing to the equivalent model trained with Adam.
In contrast to all previous cases, using SGD with/without momentum (on a longer schedule to encourage convergence),
leads to a significant decrease in mean layer kurtosis and max absolute activation, by up to 98\% and 97\%,
respectively, when comparing to the equivalent model trained with Adam.
However, using SGD requires a significantly longer training schedule to approach initial model performance.
Additionally, using SGD without momentum leads to a significantly higher perplexity ($6.8$ compared to the initial model).
This finding confirms the importance of the optimiser in causing outlier activations in the hidden states.

Using OrthoAdam yields the desirable results from SGD without momentum---namely
a significant decrease in mean layer kurtosis ($140$ to $3.0$) and max absolute activation ($432$ to $43.5$)
and the desirable results from Adam---namely similar perplexity to a model trained with Adam and therefore
much faster and better convergence than SGD without momentum.

The final three rows of~\cref{tab:ablation_big} show that using OrthoAdam with softmax-1 and RMSNorm-S
leads to the most desirable results, and critically the removal of softmax-1 and the use of LayerNorm or RMSNorm-M
reintroduces first token attention dominance and outlier activations, respectively.

\begin{table}[t!]
    \centering
    \scriptsize
    \begin{adjustbox}{width=\linewidth}
    \begin{tabular}{c|c|c|c|c|cccc}
    \toprule
    Biases     & Position Encoding & Normalisation & Optimizer    & Softmax+1? & PPL    & Kurtosis & \%First Attn & Max Abs. Act? \\ \midrule
    \checkmark & Absolute          & LayerNorm     & Adam         & \xmark     & 26.9   & 291.7    & 0.333        & 1675.9        \\
    \xmark     & Absolute          & LayerNorm     & Adam         & \xmark     & 26.9   & 263.7    & 0.308        & 1104.0        \\ \midrule
    \xmark     & None              & LayerNorm     & Adam         & \xmark     & 27.4   & 283.3    & 0.197        & 1478.7        \\
    \xmark     & Rotary            & LayerNorm     & Adam         & \xmark     & 25.6   & 391.9    & 0.336        & 2577.4        \\ \midrule
    \xmark     & Absolute          & LayerNorm     & Adam         & \checkmark & 26.5   & 244.7    & 0.022        &  648.6        \\
    \xmark     & Absolute          & RMSNorm-M     & Adam         & \checkmark & 26.6   & 230.4    & 0.026        &  628.6        \\
    \xmark     & Absolute          & RMSNorm-S     & Adam         & \checkmark & 26.6   & 140.0    & 0.020        &  432.0        \\ \midrule
    \xmark     & Absolute          & RMSNorm-S     & RMSProp      & \checkmark & 27.4   &  70.5    & 0.021        &  302.2        \\
    \xmark     & Absolute          & RMSNorm-S     & SGD w/mom*   & \checkmark & 25.3   &   5.0    & 0.019        &   17.8        \\
    \xmark     & Absolute          & RMSNorm-S     & SGD w/o mom* & \checkmark & 33.4   &   3.2    & 0.017        &   13.1        \\ \midrule
    \xmark     & Absolute          & RMSNorm-S     & OrthoAdam    & \checkmark & 26.8   &   3.0    & 0.022        &   43.5        \\
    \xmark     & Absolute          & RMSNorm-S     & OrthoAdam    & \xmark     & 27.3   & 323.0    & 0.231        &  726.4        \\
    \xmark     & Absolute          & RMSNorm-M     & OrthoAdam    & \checkmark & 26.7   & 380.9    & 0.025        &  737.2        \\
    \xmark     & Absolute          & LayerNorm     & OrthoAdam    & \checkmark & 26.6   & 188.4    & 0.023        &  514.6        \\ \bottomrule
    \end{tabular}
    \end{adjustbox}
    \caption{
        Ablation study on the impact of various architectural choices on the performance of a GPT2 model with $sim$130M parameter model.
        *SGD models are trained for $8\times$ longer than the others to encourage convergence.
    }\label{tab:ablation_big}
\end{table}
\section{Related Work}\label{sec:rw}
\parb{Language Models.}
Current language models are based on Transformer models~\citep{vaswani2017attention}.
While there are Transformer-based LLMs that used the original encoder-decoder architecture such as T5~\citep{raffel2020exploring},
researchers developed models such as BERT~\citep{devlin2019bert} and RoBERTa~\citep{liu2019roberta}, which are encoder-only.
However, most current LLMs such as the GPT series~\citep{radford2018improving,radford2019language,brown2020language}
and Llama series~\citep{touvron2023llama1,touvron2023llama,dubey2024llama} use a decoder-only architecture.
In our work, we focus on this variant using GPT2 and Llama2 architectures.

\parb{Attention Dominance.}
\citet{bondarenko2023removing} identify the dominance of bland tokens in the attention maps of the BERT encoder-only Transformer,
and suggest complex clipping schemes, additional hyperparameters, and a gating 
mechanism to to mitigate this issue.
Other researchers found the same issue in long-range attention~\citep{xiao2024efficient} and found a workaround using ``attention sinks'' and discontinuous attention masking.
In vision Transformers,~\citet{darcet2024vision} made the same observation and proposed a solution using ``registers''.
In contrast to these works, we first find the root cause of this issue, the softmax mechanism in attention, and then reformulate it to prevent the first token dominance ever happening.

\parb{Outlier Activations.}
Previous works have shown that in certain Transformer models which use post-normalisation
the norm of the \emph{weights} of the learnt model must increase~\citep{arora2018theoretical,soudry2018implicit}.
However the same reasoning does not apply for most recent decoder-only Transformers which use pre-normalisation~\citep{xiong2020layer}~(\ie~normalisation
before the residual connection).
A blog-post by~\citet{elhage2023privileged} discusses the presence of outlier activations
in the hidden states of Transformer models and rules out numerical precision as the cause.
Another blog-post by~\citet{miller2023attention} posits the
activation outliers are caused by the attention mechanism,
however, we find outliers and attention dominance are disjoint phenomena.
\citet{he2024understanding} identify the presence of outliers and propose an ``Outlier Protected Transformer Block'' which makes many architectural changes such as
removing normalisation layers and severely downscaling the activations at the residual connection.
In our contrast, similar to first token dominance, we first find the root cause of this strange behaviour, and then fix it without doing architecture changes.

\parb{Outlier-Aware Quantisation.}
The presence of outliers in the activations of the hidden states has led to
a number of works, such as \texttt{LLM.int8}~\citep{dettmers2022llmint},
per-embedding group quantisation~\citep{bondarenko2021understanding}, and
SmoothQuant~\citep{xiao2023smoothquant} propose varying
quantisation schemes to handle the presence of outliers, which require calibration.
In contrast, we eliminate the presence of outliers in our trained models thus enabling
the use of the most basic quantisation schemes such as Absmax and Zeropoint quantisation.
\section{Conclusion}\label{sec:conc}
In this work, we study two surprising phenomena in large auto-regressive Transformers:
(1) the strong, consistent dominance of the first token in attention maps;
and (2) the presence of outlier activations in the hidden states.
We propose novel solutions:
(1) the softmax-1 function to remove first token dominance;
and (2) the OrthoAdam optimiser which mitigates outlier activations.
By doing so,
we reduce first token dominance of attention maps by up to $95$\% and
the activation kurtosis by up to $99.8$\%.
Furthermore, our work improves our understanding of Transformers but also offer practical benefits in model quantisation,
reducing the quantisation penalty by up to $99.9$\%.

\newpage
\bibliography{iclr2025_conference_pk}
\bibliographystyle{iclr2025_conference}

\newpage
\appendix
\numberwithin{equation}{section}
\newpage
\section{Training Details}\label{sec:training_details}
In this section, we provide details on the training of our models.

In all experiments we use a batch size of 512 and in all experiments
using Adam or OrthoAdam as the optimiser, we use a peak learning rate of $10^{-3}$.
This excludes the experiments in~\cref{ssec:ablation} which use SGD as the optimiser,
which use a peak learning rate of $0.2$.
In all experiments we use a cosine learning rate schedule with linear warmup for
\{1000, 2000, 6000, 10000\} steps for models with \{60M, 130M, 350M, 1.4B\} parameters respectively.
Note that we use a reduced number of steps for the 1.4B model due to computational constraints.
For the main experimental results in~\cref{tab:main_table_small,tab:quant_table},
we train the models with \{60M, 130M, 350M, 1.4B\} parameters for \{160k, 320k, 960k, 600k\} steps respectively.
For the ablation study in~\cref{ssec:ablation},
we train GPT2 models with 130M parameters for 40k steps with 2000 warmup steps.
We use a maximum sequence length of 256 tokens,
which we find is sufficient to observe the anomalies of first token attention dominance and large outlier activations
found in popular pretrained models such as GPT2~\citep{radford2019language}
and Llama~\citep{touvron2023llama1,touvron2023llama,dubey2024llama}.
The result of our training setup is that models trained for the main experimental results with \{60M, 130M, 350M, 1.4B\} parameters
are trained on \{21B, 42B, 126B, 79B\} tokens respectively.
The ablation experiments are trained on 5B tokens.
We train models on 8 NVIDIA 32GB V100 GPUs using the Pytorch deep-learning framework~\citep{paszke2019pytorch}
and the HuggingFace Transformers library~\citep{wolf2020transformers}.

\section{Note on Supplementary Material}\label{sec:supplementary}
The supplementary material contains three folders:
\begin{itemize}[noitemsep,topsep=0pt,leftmargin=15pt]
    \setlength\itemsep{8pt}
    \item \texttt{our\_attention\_maps/}: additional attention map plots for our trained models from~\cref{ssec:main_results}, using softmax-1 removes first token dominance.
    \item \texttt{our\_hidden\_states/}: additional hidden state plots for our trained models from~\cref{ssec:main_results}, using OrthoAdam and softmax-1 together removes outlier activations.
    \item \texttt{our\_output\_weights/}: plots showing the maximum norm of the output features for the final feedforward layer in each attention and MLP sub-block in
    our trained models from~\cref{ssec:main_results}, using OrthoAdam and softmax-1 leads to a large reduction in the maximum norm of the output weights.
\end{itemize}

\newpage
\section{Kurtosis grows with the number of dimensions in Transformers}\label{sec:kurtosis}
In this section, we use some observations from the hidden states of transformer models to
illustrate how the kurtosis of the hidden states grows with the number of dimensions in the hidden states.
This is something we observe empirically in the hidden states of transformer models and is a key motivation for our work.
\cref{tab:main_table_small} shows the kurtosis of the hidden states of transformer models
trained \emph{without softmax-1 or OrthoAdam} grows as the model size increases,
as does the maximum activation value in the hidden states.

To make this mathematically rigorous, we shall consider a simple example,
in which we shall approximate the hidden states of a transformer model at a single token position
as a $D$-dimensional vector comprising of the sum of a scaled one-hot vector and a standard normal vector.

Consider a $D$-dimensional vector $\mathbf{x}$ which is the sum of two $D$-dimensional vectors
$\alpha\mathbf{e}_i$ and $\mathbf{z}$, where $\mathbf{e}_i$ is the $i$th unit vector in the standard basis,
$\mathbf{x} \in \real^D$, $\alpha \in \real$ and $\mathbf{z} \sim \mathcal{N}(\mathbf{0}, \mathbf{I}_{D})$.
Therefore the elements of $\mathbf{x}$ are given by: 
\begin{align}
    x_j &= \alpha \delta_{ij} + z_j \quad \text{for}~j = 1, 2, \ldots, D \nonumber
\end{align}
where $\delta_{ij}$ is the Kronecker delta function.
The mean is given by:
\begin{align}
    \mu = \expect\left[x_j\right] &= \expect\left[\alpha \delta_{ij} + z_j\right] 
                                      = \alpha \expect\left[\delta_{ij}\right] + \expect\left[z_j\right] \nonumber \\
                                      &= \frac{\alpha}{D} + 0 = \frac{\alpha}{D} \quad \text{as}~\expect\left[z_j\right] = 0~\text{by definition of the standard normal distribution} \nonumber \\
                                      \Aboxed{\mu &= \frac{\alpha}{D}}
\end{align}
The variance is given by:
\begin{align}
    \sigma^2 &= \var\left[x_j\right] = \var\left[\alpha \delta_{ij} + z_j\right] \nonumber \\
             &= \alpha^2\var\left[\delta_{ij}\right] + \var\left[z_j\right] \quad \text{as}~\alpha \delta_{ij}~\text{and}~z_j~\text{are independent in our model} \nonumber \\
             &= \alpha^2\var\left[\delta_{ij}\right] + 1 \quad \text{as}~\var\left[z_j\right] = 1~\text{by definition of the standard normal distribution} \nonumber \\
    \var\left[\delta_{ij}\right] &= \expect\left[\delta_{ij}^2\right] - \left(\expect\left[\delta_{ij}\right]\right)^2 = \frac{1}{D}\left(1 - \frac{1}{D}\right) \nonumber \\
\intertext{Therefore:}
    \Aboxed{\sigma^2 &= \frac{\alpha^2}{D}\left(1 - \frac{1}{D}\right) + 1}
\end{align}

The kurtosis of the elements of $\mathbf{x}$ is given by:
\begin{align}
    \text{Kurt}[x_j] &= \expect\left[{\left(\frac{x_j - \mu}{\sigma}\right)}^4\right]
                      = \frac{\expect\left[\left(x_j - \mu\right)^4\right]}{\sigma^4} \nonumber
\end{align}
When $j \neq i$:
\begin{align}
    \expect\left[\left(x_j - \mu\right)^4\right]  &= \expect\left[\left(z_j - \frac{\alpha}{D}\right)^4\right] \nonumber \\
                                                  &= \expect\left[z_j^4 - 4z_j^3\frac{\alpha}{D} + 6z_j^2\left(\frac{\alpha}{D}\right)^2 - 4z_j\left(\frac{\alpha}{D}\right)^3 + \left(\frac{\alpha}{D}\right)^4\right] \nonumber \\
                                                  &= \expect\left[z_j^4\right] - 4\expect\left[z_j^3\right]\frac{\alpha}{D} + 6\expect\left[z_j^2\right]\left(\frac{\alpha}{D}\right)^2 - 4\expect\left[z_j\right]\left(\frac{\alpha}{D}\right)^3 + \left(\frac{\alpha}{D}\right)^4 \nonumber \\
\intertext{As $\expect\left[z_j^3\right] = 0$ and $\expect\left[z_j^4\right] = 3$:}
                                                  &= 3 + 6\left(\frac{\alpha}{D}\right)^2 + \left(\frac{\alpha}{D}\right)^4 \nonumber
\end{align}
When $j = i$:
\begin{align}
    \expect\left[\left(x_j - \mu\right)^4\right] &= \expect\left[\left(\alpha + z_j - \frac{\alpha}{D}\right)^4\right] \nonumber \\
                                                  &= \expect\left[\left(\alpha\left(1 - \frac{1}{D}\right) + z_j\right)^4\right] \nonumber \\
                                                  &= \expect\left[\left(\alpha\left(1 - \frac{1}{D}\right)\right)^4 + 4\left(\alpha\left(1 - \frac{1}{D}\right)\right)^3z_j\right. \nonumber \\
                                                  &\left. \qquad\quad +~6\left(\alpha\left(1 - \frac{1}{D}\right)\right)^2z_j^2 + 4\left(\alpha\left(1 - \frac{1}{D}\right)\right)z_j^3 + z_j^4\right] \nonumber \\
                                                  &= \left(\alpha\left(1 - \frac{1}{D}\right)\right)^4 + 6\left(\alpha\left(1 - \frac{1}{D}\right)\right)^2 + 3 \nonumber
\end{align}
Therefore, the \emph{overall fourth moment} of the elements of $\mathbf{x}$ is given by:
\begin{align}
    \expect\left[\left(x_j - \mu\right)^4\right] &= \frac{1}{D}\left(\left(\alpha\left(1 - \frac{1}{D}\right)\right)^4 + 6\left(\alpha\left(1 - \frac{1}{D}\right)\right)^2 + 3\right) \nonumber \\
                                                 &\quad + \frac{D - 1}{D}\left(3 + 6\left(\frac{\alpha}{D}\right)^2 + \left(\frac{\alpha}{D}\right)^4\right) \nonumber
\end{align}
And the kurtosis of the elements of $\mathbf{x}$ is given by:
\begin{align}
\text{Kurt}\left[x_j\right] &= \frac{\frac{1}{D}\left(\left(\alpha\left(1 - \frac{1}{D}\right)\right)^4 + 6\left(\alpha\left(1 - \frac{1}{D}\right)\right)^2 + 3\right) + \frac{D - 1}{D}\left(3 + 6\left(\frac{\alpha}{D}\right)^2 + \left(\frac{\alpha}{D}\right)^4\right)
}{\left(\frac{\alpha^2}{D}\left(1 - \frac{1}{D}\right) + 1\right)^2} \nonumber \\
\Aboxed{
\text{Kurt}\left[x_j\right] &= \frac{
                                3 + \frac{\alpha^4}{D} + \frac{6\alpha^2}{D} - \frac{4\alpha^4}{D^2} - \frac{6\alpha^2}{D^2} + \frac{6\alpha^4}{D^3} - \frac{3\alpha^4}{D^4}
                            }{
                                1 + \frac{2\alpha^2}{D} - \frac{2\alpha^2}{D^2} + \frac{\alpha^4}{D^2} - \frac{2\alpha^4}{D^3} + \frac{\alpha^4}{D^4}
                            }\label{eq:kurtosis_model}
}
\end{align}
At this point, we can see that Kurtosis is a function of $\alpha$ and $D$, however if
we consider the limit as $D \to \infty$, we can see that $\text{Kurt}[x_j] \to 3$ \ie~the kurtosis of a Gaussian distribution.
However, this neglects the importance of the scaling factor $\alpha$ which we know empirically is larger than the dimensionality of the hidden states.
The table below summarises the maximum activation values (analogous to $\alpha$) and the dimension of the hidden states for the models we trained.
\begin{table}[h!]
    \centering
    \begin{tabular}{@{}l|c|c|c@{}}
    \toprule
    \multicolumn{1}{c|}{Model} & \#Parameters & Model Size ($D$) & Max Activation ($\alpha$) \\ \midrule
    \multirow{4}{*}{GPT2}      & 60M          & 512              & 1856                      \\
                               & 130M         & 768              & 7018                      \\
                               & 350M         & 1024             & 40196                     \\
                               & 1.4B         & 2048             & 56798                     \\ \midrule
    \multicolumn{1}{c|}{Llama} & 130M         & 768              & 4623                      \\ \bottomrule
    \end{tabular}
    \caption{Model sizes and maximum activation values for the models used in our experiments.}\label{tab:alpha_D_table}
\end{table}
Given this empirical information, we make the \emph{conservative} assumption that $\alpha = D$.
Under this assumption which is supported by our empirical observations,~\cref{eq:kurtosis_model} simplifies to:
\begin{align}
    \text{Kurt}\left[x_j\right] &=
    \frac{
        3 + \frac{D^4}{D} + \frac{6D^2}{D} - \frac{4D^4}{D^2} - \frac{6D^2}{D^2} + \frac{6D^4}{D^3} - \frac{3D^4}{D^4}
    }{
        1 + \frac{2D^2}{D} - \frac{2D^2}{D^2} + \frac{D^4}{D^2} - \frac{2D^4}{D^3} + \frac{D^4}{D^4}
    } \nonumber \\
    &= \frac{3 + D^3 + 6D - 4D^2 - 6 + 6D - 3}{1 + 2D - 2 + D^2 - 2D + 1} \nonumber \\
    &= \frac{D^3 - 4D^2 + 12D - 6}{D^2} \nonumber \\
\Aboxed{
\text{Kurt}\left[x_j\right] &= D - 4 + \frac{12}{D} - \frac{6}{D^2} = O(D) \label{eq:kurtosis_model_final}
}
\end{align}
Using our conservative assumption that $\frac{\alpha}{D} = 1$,
we can see that the kurtosis of the hidden states grows linearly with the dimensionality of the hidden states
when $D$ is in the region of $10^3-10^5$ as is the case for transformer models.

This simple example serves as a mathematical illustration of the empirical observations we make in the hidden states of transformer models.
We have shown that the kurtosis of the hidden states is expected to grow linearly with the dimensionality of the hidden states,
and so the issue of outlier activations is expected to grow as the hidden states of transformer models grow in size.
\newpage
\section{Orthogonal transformations and reduction in $\ell_\infty$--norm and Kurtosis}\label{ssec:inf-2norm}
From our simple model in~\cref{sec:kurtosis} we have a simplified model of Transformer hidden states, $\mathbf{x} \in \real^D$,
where the first element is $\alpha$ and the rest are standard normal random variables.
\begin{align}
    \mathbf{x} &= \alpha\mathbf{e}_i + \mathbf{z} \quad \text{where}~z_j \sim \mathcal{N}(0, 1) \nonumber
\end{align}
From this model, we can compute the expected $\ell_2$--norm:
\begin{align}
    \expect\left[\left\lVert\mathbf{x}\right\rVert_2^2\right] &= \sum_{j=1}^D x_j^2 = \alpha^2 + \sum_{j=1}^D z_j^2 = \alpha^2 + D\var\left[z_j\right]
    = \alpha^2 + D
\end{align}
Using the triangle inequality, we can compute a range for the $\ell_\infty$--norm:
\begin{align}
    \expect\left[\left\lVert\mathbf{x}\right\rVert_\infty\right] = \expect\left[\max_{1 \leq j \leq D} \left(\left|\alpha + z_i\right|, \max_{j \neq i} \left|z_j\right|\right)\right] \nonumber \\
\intertext{Given $\alpha \gg 1$, we can drop the terms for $j \neq i$ and compute the expected $\ell_\infty$--norm using the $i$\textsuperscript{th} element:}
    \expect\left[\left\lVert\mathbf{x}\right\rVert_\infty\right] = \expect\left[\left|\alpha + z_i\right|\right] \nonumber \\
    \left|\alpha + z_i\right| \leq \left|\alpha\right| + \left|z_i\right| \nonumber \\
\intertext{Using folded normal distribution properties, $\expect\left[\left|z_i\right|\right] = \sqrt{\frac{2}{\pi}} \ll \alpha$, therefore:}
    \expect\left[\left\lVert\mathbf{x}\right\rVert_\infty\right] \approx \alpha \nonumber
\end{align}
Given that $\alpha \gg 1$, we can safely assume that $\left\lVert\mathbf{x}\right\rVert_\infty^2 \approx \alpha^2$.
Therefore:
\begin{align}
    \expect\left[\frac{\left\lVert\mathbf{x}\right\rVert_\infty}{\left\lVert\mathbf{x}\right\rVert_2}\right] &\approx \frac{\alpha}{\sqrt{D + \alpha^2}} \nonumber
\intertext{Note from~\cref{tab:alpha_D_table} that the maximum activation value, $\alpha$, is generally much larger than the model size, $D$.}
\expect\left[\frac{\left\lVert\mathbf{x}\right\rVert_\infty}{\left\lVert\mathbf{x}\right\rVert_2}\right] &\approx 1 \label{eq:inf-2norm_appendix}
\end{align}
We find this empirically to be the case in the middle layers of the Transformer models we study (see plots in~\cref{ssec:linf_l2_ratio}).

The $\infty$-norm of $\mathbf{x}$ can be thought of as a proxy for the extent of outliers in a vector.
If $\frac{\left\lVert\mathbf{x}\right\rVert_2}{\left\lVert\mathbf{x}\right\rVert_{\infty}} \approx 1$,
then a vector has at least one large outlier and consequently a high kurtosis.

We will now show that applying an orthogonal transformation to a vector can reduce the $\ell_\infty$-norm constrained
to a fixed $\ell_2$-norm. Using the same definition of $\mathbf{x}$ as above, let $\mathbf{Q} \in \real^{D \times D}$ be an orthogonal matrix
and let $\mathbf{y} = \mathbf{Q}\mathbf{x}$.
\begin{align}
    \left\lVert\mathbf{y}\right\rVert^{2}_{2} &= \mathbf{y}^T\mathbf{y} = \mathbf{x}^T\mathbf{Q}^T\mathbf{Q}\mathbf{x} = \mathbf{x}^T\mathbf{x} = \left\lVert\mathbf{x}\right\rVert^{2}_{2} \nonumber \\
    \expect[\left\lVert\mathbf{y}\right\rVert^{2}_{2}] &= \expect[\left\lVert\mathbf{x}\right\rVert^{2}_{2}] = \alpha^2 + D \label{eq:ortho_l2}
\end{align}

This standard proof shows that applying an orthogonal transformation to a vector does not change the $\ell_2$--norm of the vector.
It can however lead to a dramatic reduction in the $\ell_\infty$--norm of the vector.
We will now show that for a vector, $\mathbf{y} \in \real^D$, constrained to have a fixed $\ell_2$--norm, $\sqrt{\alpha^2 + D}$,
the $\ell_\infty$--norm of a vector can be reduced significantly by applying an orthogonal transformation
such that $y_j = \frac{\alpha}{\sqrt{D}} + z', \forall j\in[1,D]$, where $z' \sim \mathcal{N}(0, 1)$.

\begin{align}
    \mathbf{y} &= \mathbf{Q}\mathbf{x} = \mathbf{Q}\left(\alpha\mathbf{e}_i + \mathbf{z}\right) = \alpha\mathbf{Q}\mathbf{e}_i + \mathbf{Q}\mathbf{z} \nonumber \\
\intertext{Select $\mathbf{Q}$ such that $\mathbf{Q}\mathbf{e}_i = \left(\frac{1}{\sqrt{D}}, \frac{1}{\sqrt{D}}, \ldots, \frac{1}{\sqrt{D}}\right)$,
given $\mathbf{Q}$ is orthogonal, $\mathbf{Q}\mathbf{z}~=~\mathbf{z'}~\sim~\mathcal{N}(0, \mathbf{I}_D)$.} \nonumber
    \expect\left[\left\lVert\mathbf{y}\right\rVert_{\infty}\right] &\approx \frac{\alpha}{\sqrt{D}} + \sqrt{2\ln{D}}, \quad \text{using extreme value theory~\citep{cramer1946mathematical}} \nonumber \\
    &= \frac{\alpha + \sqrt{2D\ln{D}}}{\sqrt{D}} \nonumber
\end{align}
The expected ratio of $\ell_\infty$--norm to $\ell_2$--norm is:
\begin{align}
    \expect\left[\frac{\left\lVert\mathbf{y}\right\rVert_{\infty}^2}{\left\lVert\mathbf{y}\right\rVert_{2}^2}\right] &= \frac{\expect\left[\left\lVert\mathbf{y}\right\rVert_{\infty}^2\right]}{\expect\left[\left\lVert\mathbf{y}\right\rVert_{2}^2\right]}
    = \frac{\left(\alpha + \sqrt{2D\ln{D}}\right)^2}{D\left(\alpha^2 + D\right)} \nonumber
\intertext{Using the same \emph{conservative} assumption as in~\cref{sec:kurtosis} that $\alpha = D$,~\cref{tab:alpha_D_table} shows empirically $\alpha > D$:}
    \expect\left[\frac{\left\lVert\mathbf{y}\right\rVert_{\infty}^2}{\left\lVert\mathbf{y}\right\rVert_{2}^2}\right] &=
    \frac{D^2 + 2D\ln{D} + 2D\sqrt{2D\ln{D}}}{D^3 + D^2} = \frac{D + 2\sqrt{2D\ln{D}} + 2\ln{D}}{D^2 + 1} \nonumber
\intertext{As $D$ grows, the last term of the numerator and the $1$ in the denominator become negligible:}
    \expect\left[\frac{\left\lVert\mathbf{y}\right\rVert_{\infty}^2}{\left\lVert\mathbf{y}\right\rVert_{2}^2}\right] &\approx \frac{1}{D} + \frac{D + 2\sqrt{2\ln{D}}}{D^{\frac{3}{2}}} = O\left(\frac{1}{D}\right) \nonumber 
\end{align}
Therefore, under an orthogonal transformation, the $\ell_\infty$--norm to $\ell_2$--norm ratio can be reduced significantly.
It is trivial to show that $\text{Kurt}[y_j] = 3$ and we see many of our experiments
which use OrthoAdam and softmax-1 exhibit this behaviour (see plots in~\cref{ssec:plots_kurtosis}).
\begin{equation*}
\boxed{
    \begin{aligned}
    \mathbf{x} = \alpha\mathbf{e}_i + \mathbf{z}, \quad \expect\left[\frac{\left\lVert\mathbf{x}\right\rVert_{\infty}^2}{\left\lVert\mathbf{x}\right\rVert_{2}^2}\right] \approx 1
\quad &\rightarrow \quad \mathbf{y} = \mathbf{Q}\mathbf{x}, \quad \expect\left[\frac{\left\lVert\mathbf{y}\right\rVert_{\infty}^2}{\left\lVert\mathbf{y}\right\rVert_{2}^2}\right] \approx \frac{1}{D} \\
    \text{Kurt}\left[x_j\right] = D - 4 + \frac{12}{D} - \frac{6}{D^2} = O(D) \quad &\rightarrow \quad \text{Kurt}\left[y_j\right] = 3
    \end{aligned}
}
\end{equation*}

The exact form of $\mathbf{Q}$ can be computed numerically or constructed using appropriately normalised Hadamard matrices~\citep{sylvester1867thoughts}.


\newpage
\section{
    Layer Progression of First Token Attention Dominance,
    Kurtosis,
    $\ell_\infty$-Norm to $\ell_2$-Norm Ratio and Maximum Absolute Activation}\label{sec:layer_progression}
For brevity,
we give metrics for the first token attention dominance,
hidden state kurtosis and absolute maximum activation \emph{averaged over all layers}
in~\cref{tab:main_table_small} which gives the results of the main experiments in our work.

However, the layer-wise progression of these metrics is also of interest and
can provide insights into the behaviour of the model.
Additionally, we provide the same metrics for popular pretrained GPT2 and Llama models
to show the similarity to our models \emph{trained without softmax-1 and OrthoAdam}.

Finally, to establish a relationship between activation kurtosis and the $\ell_\infty$-norm to $\ell_2$-norm ratio,
we calculate the Pearson's correlation coefficients between per-layer kurtosis and per-layer $\ell_\infty$-norm to $\ell_2$-norm ratio
for all models in our main experimental results from~\cref{tab:main_table_small}.

All metrics are computed on the same validation set of the C4 dataset~\citep{raffel2020exploring}
as in the main paper~(\cref{sec:exp}).

\subsection{First Token Attention Dominance}
We begin by examining the progression of first token attention dominance across layers.
We calculate the percentage of (head, query) pairs where the query token attends most to the first (key) token.
Given different models have a different number of layers,
we normalise the layer index to the range $[0, 1]$ for each model.

We find a general trend across our trained models which use the canonical softmax function
where the first token attention dominance begins low in the initial layers where models do
initial processing of all input tokens.
The dominance rises to a peak in the middle layers where heads specialise to specific sub-tasks
and so the first token is attended to as a default ``no-op''~\citep{bondarenko2023removing,clark2019bert}.
Finally, the dominance decreases in the final layers where the model ``detokenises'' the features
back into token space.

\subsubsection{GPT2-60M}
\begin{figure}[H]
    \centering
    \includegraphics[width=1.0\textwidth]{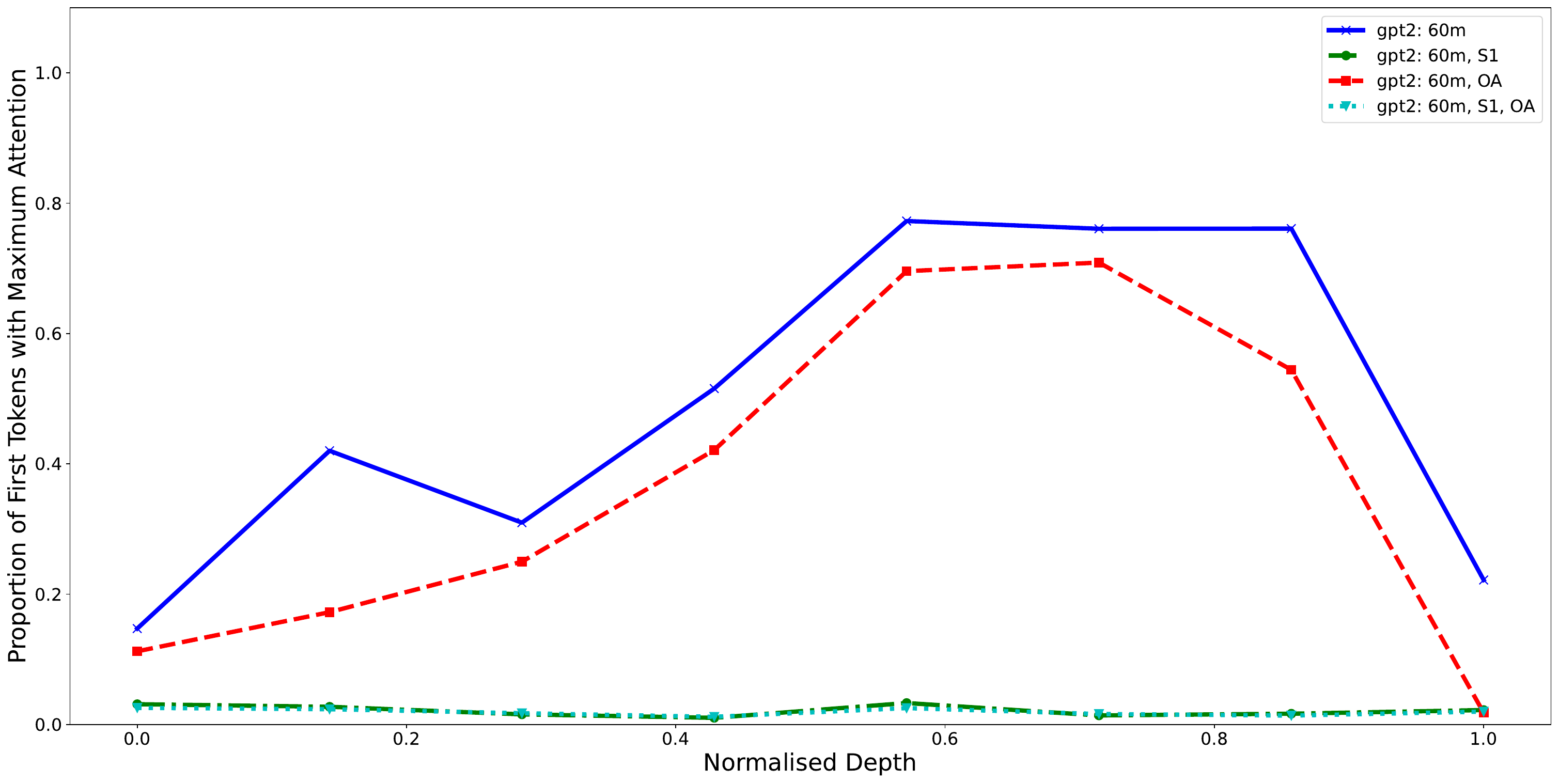}
    \caption{
        Layer-wise progression of first token attention dominance for GPT2-60M.
        The x-axis is normalised to the range $[0, 1]$. S1/OA denote models trained with softmax-1 and/or OrthoAdam.
    }\label{fig:our_num_first_gpt2_60m}
\end{figure}

\subsubsection{GPT2-130M}
\begin{figure}[H]
    \centering
    \includegraphics[width=1.0\textwidth]{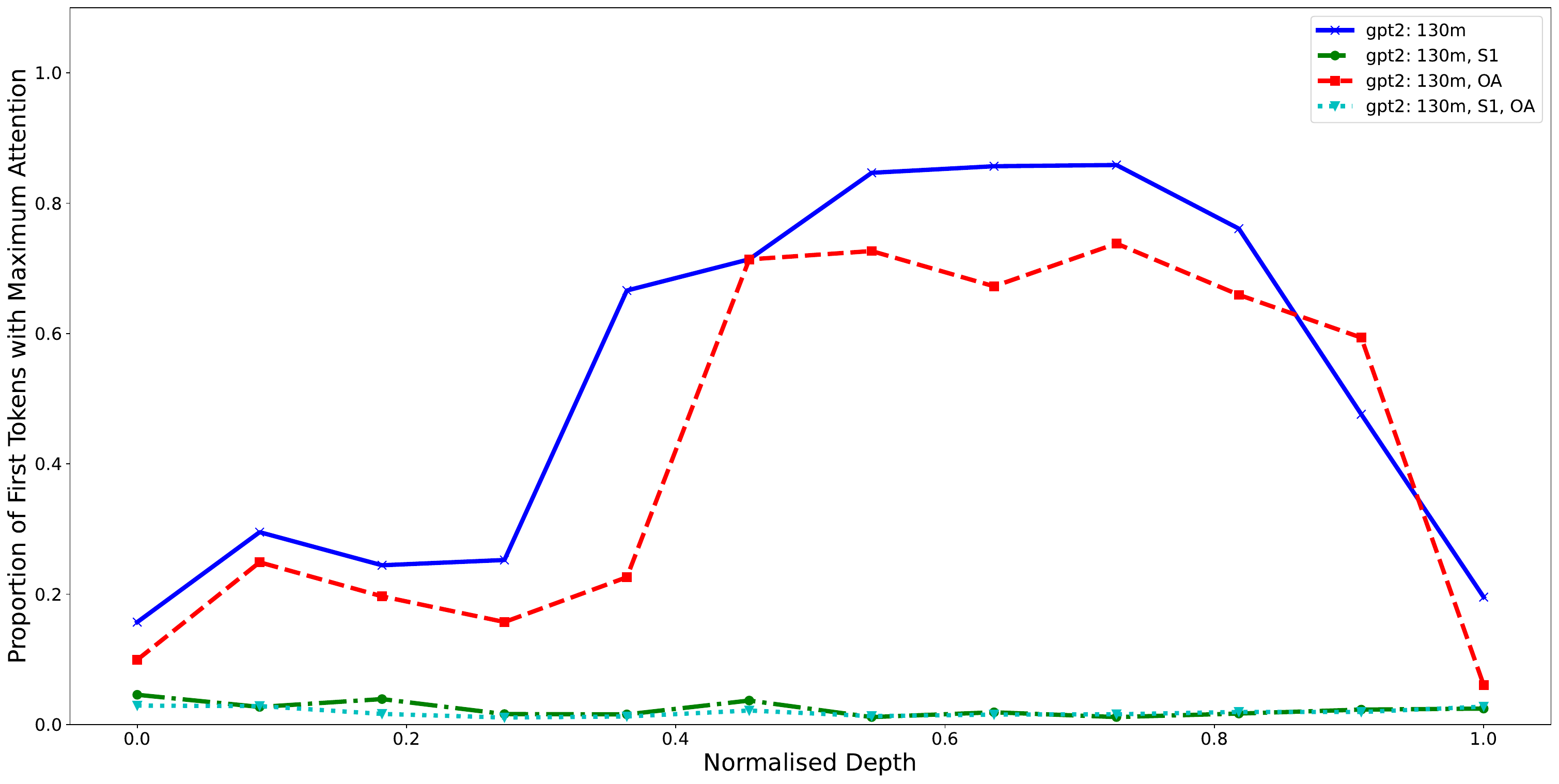}
    \caption{
        Layer-wise progression of first token attention dominance for GPT2-130M.
        The x-axis is normalised to the range $[0, 1]$. S1/OA denote models trained with softmax-1 and/or OrthoAdam.
    }\label{fig:our_num_first_gpt2_130m}
\end{figure}

\subsubsection{GPT2-350M and GPT2-1.4B}
\begin{figure}[H]
    \centering
    \includegraphics[width=1.0\textwidth]{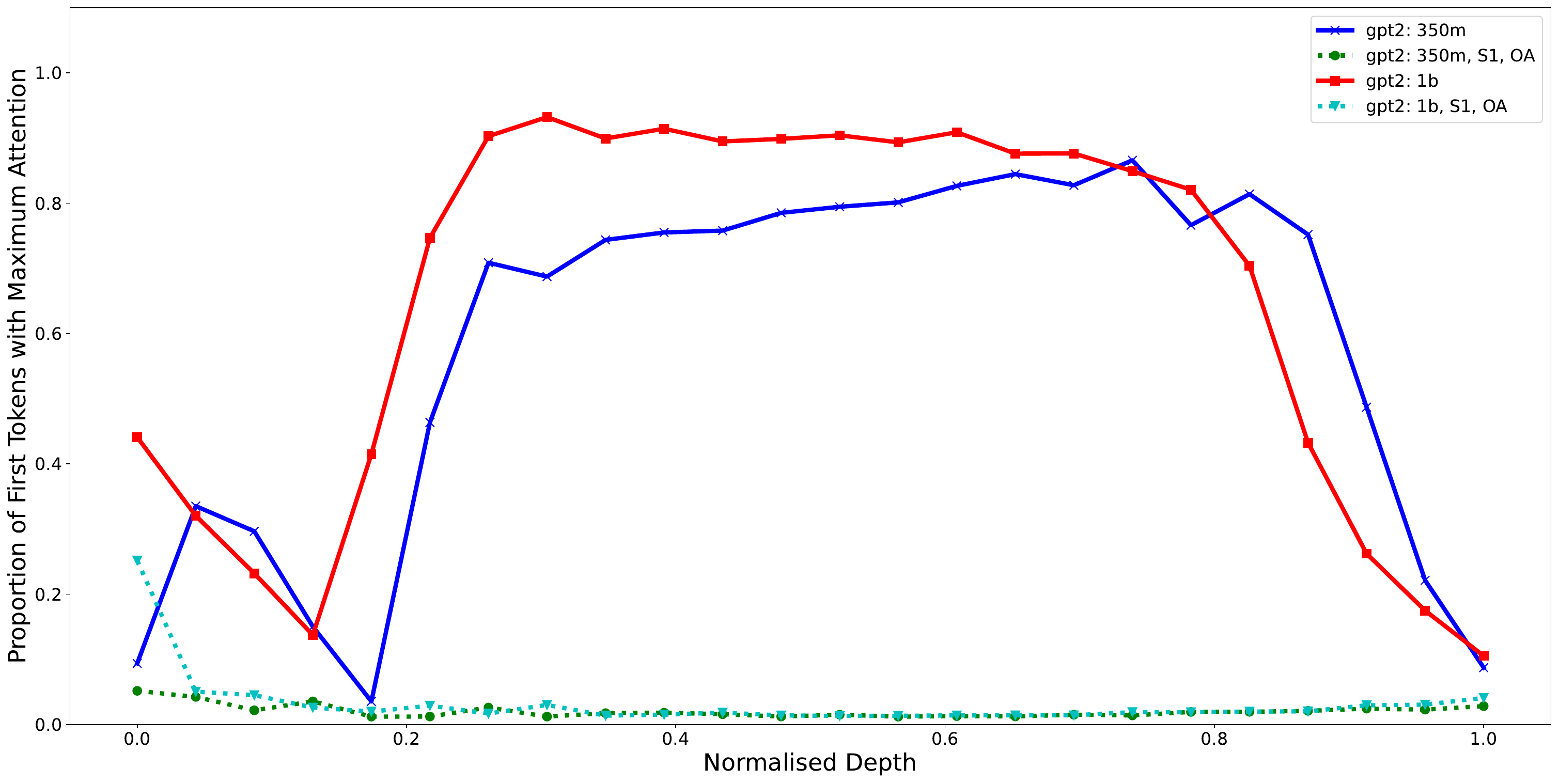}
    \caption{
        Layer-wise progression of first token attention dominance for GPT2-350M and GPT2-1.4B.
        The x-axis is normalised to the range $[0, 1]$. S1/OA denote models trained with softmax-1 and/or OrthoAdam.
    }\label{fig:our_num_first_gpt2_350m}
\end{figure}

\subsubsection{Llama-130M}
\begin{figure}[H]
    \centering
    \includegraphics[width=1.0\textwidth]{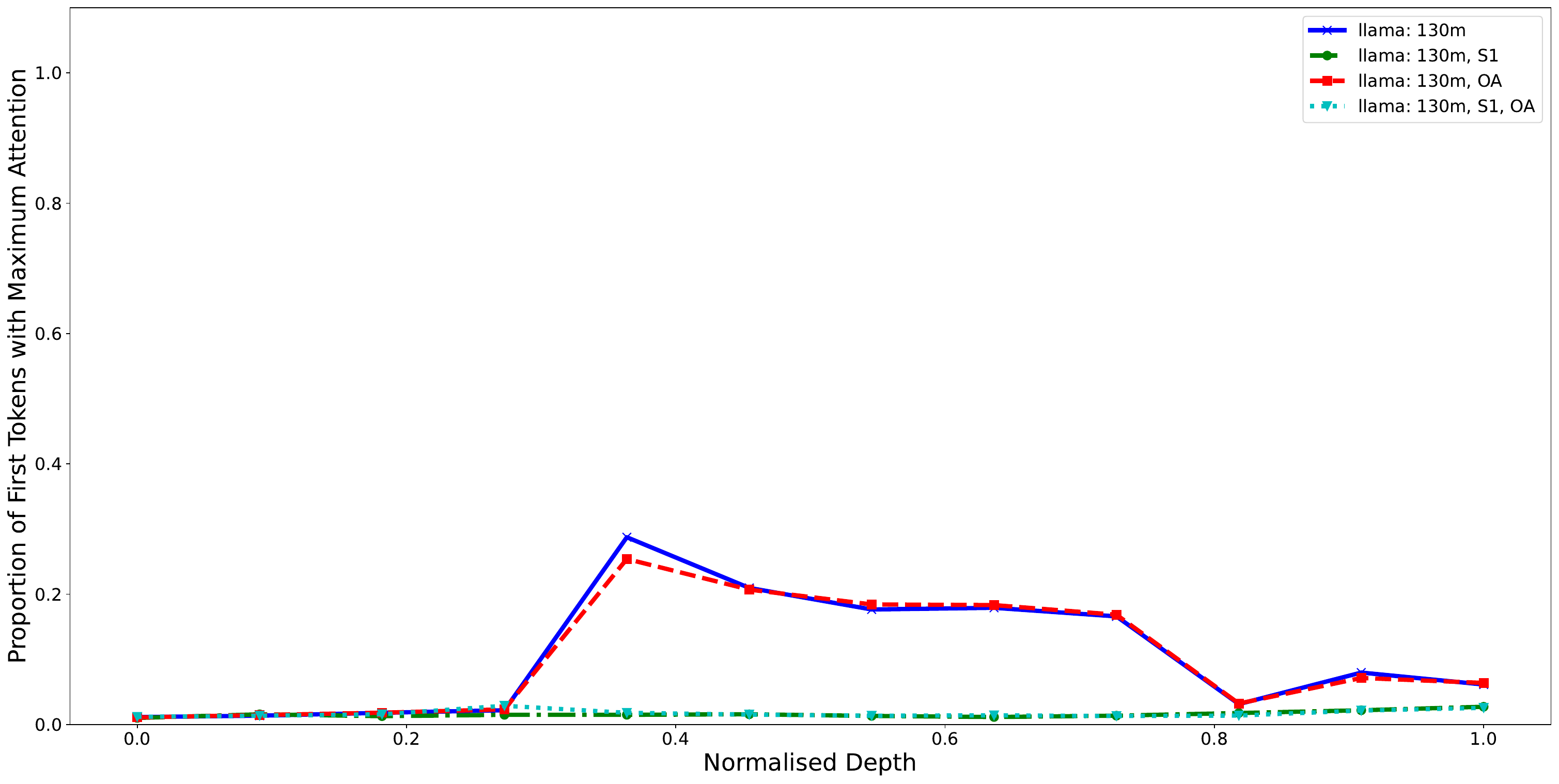}
    \caption{
        Layer-wise progression of first token attention dominance for Llama-130M.
        The x-axis is normalised to the range $[0, 1]$. S1/OA denote models trained with softmax-1 and/or OrthoAdam.
    }\label{fig:our_num_first_llama_130m}
\end{figure}

\subsubsection{Popular Pretrained Models---GPT2 and Llama}
\begin{figure}[H]
    \centering
    \includegraphics[width=1.0\textwidth]{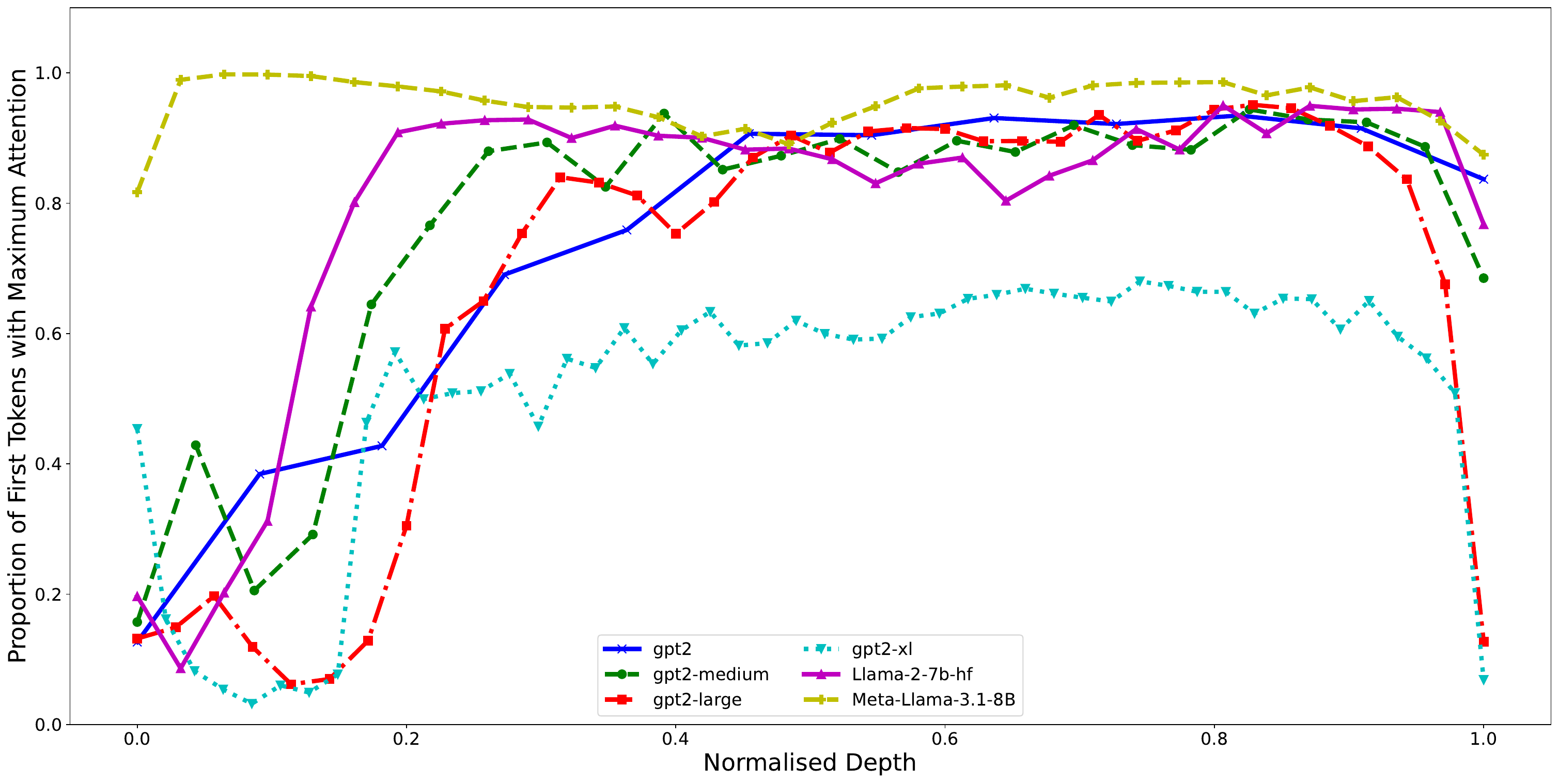}
    \caption{
        Layer-wise progression of first token attention dominance for popular pretrained GPT2 and Llama models.
        The x-axis is normalised to the range $[0, 1]$.
    }\label{fig:pretrained_num_first}
\end{figure}

\subsection{Activation Kurtosis}\label{ssec:plots_kurtosis}
Next, we examine the progression of activation kurtosis across layers.
As observable in~\cref{tab:main_table_small}, the kurtosis of the first hidden state is significantly higher
than the other hidden states and so we plot the kurtosis of the first hidden state only for brevity.

We observe in the plots below that models trained without OrthoAdam exhibit a general trend of increasing kurtosis
as the hidden states progress through the layers. Demonstrating that multiple layers of the model
contribute to the emergence of large activation values.
Models trained with OrthoAdam \emph{but not softmax-1} exhibit a similar trend, but with lower kurtosis values initially.
Finally, models trained with both OrthoAdam and softmax-1 exhibit a consistent small kurtosis across layers---around
the value of 3 which is the kurtosis of a Gaussian distribution.
Interestingly, GPT2-60M and GPT2-130M show small rises in the final layers---the cause of this is left for future work.

We find that some models show a reduction in kurtosis in the final layers, we again
attribute this to the ``detokenisation'' of the features back into token space.

\subsubsection{GPT2-60M}
\begin{figure}[H]
    \centering
    \includegraphics[width=1.0\textwidth]{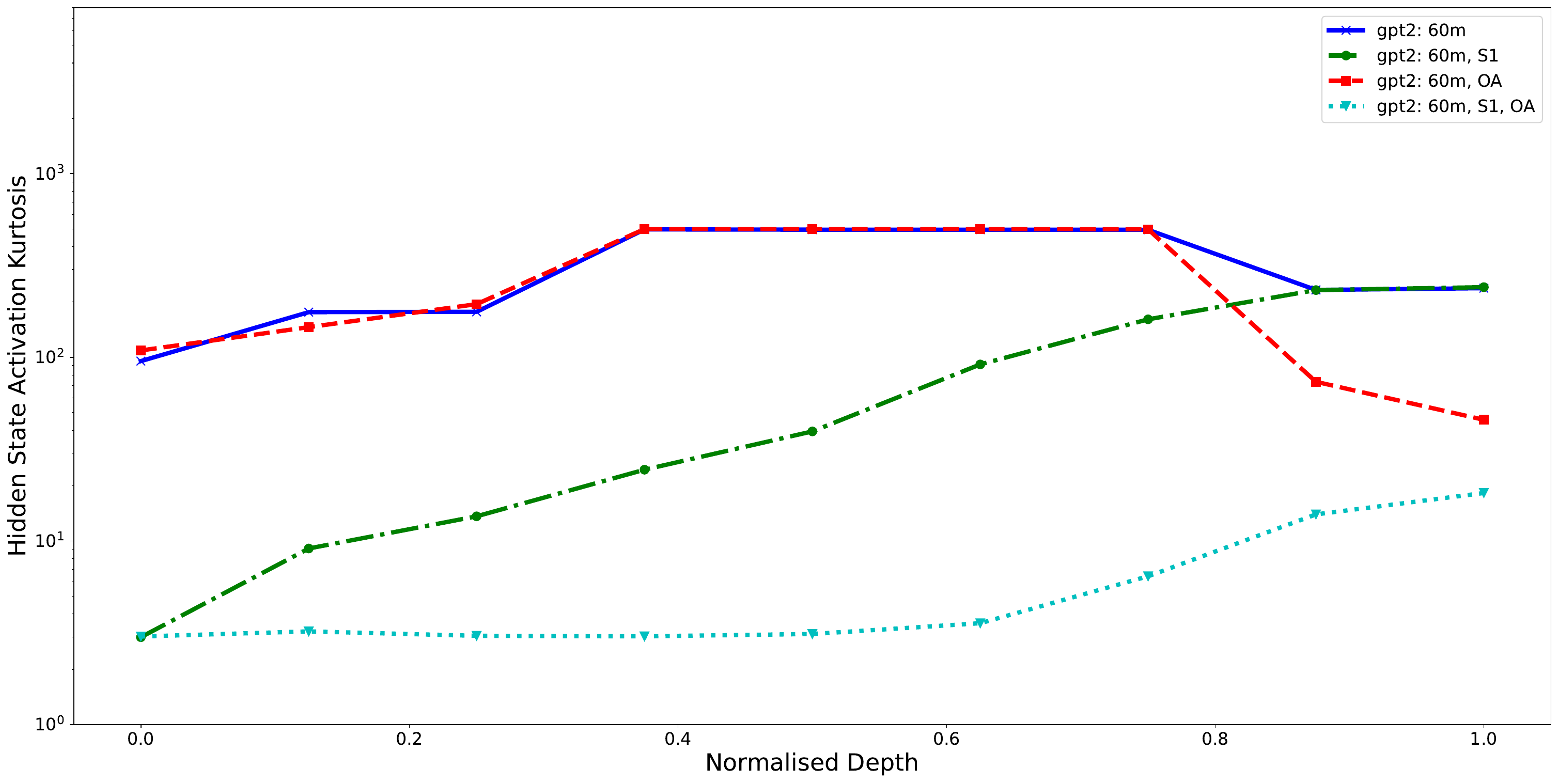}
    \caption{
        Layer-wise progression of activation kurtosis of the first token position for GPT2-60M.
        The x-axis is normalised to the range $[0, 1]$. S1/OA denote models trained with softmax-1 and/or OrthoAdam.
    }\label{fig:our_kurtosis_gpt2_60m}
\end{figure}

\subsubsection{GPT2-130M}
\begin{figure}[H]
    \centering
    \includegraphics[width=1.0\textwidth]{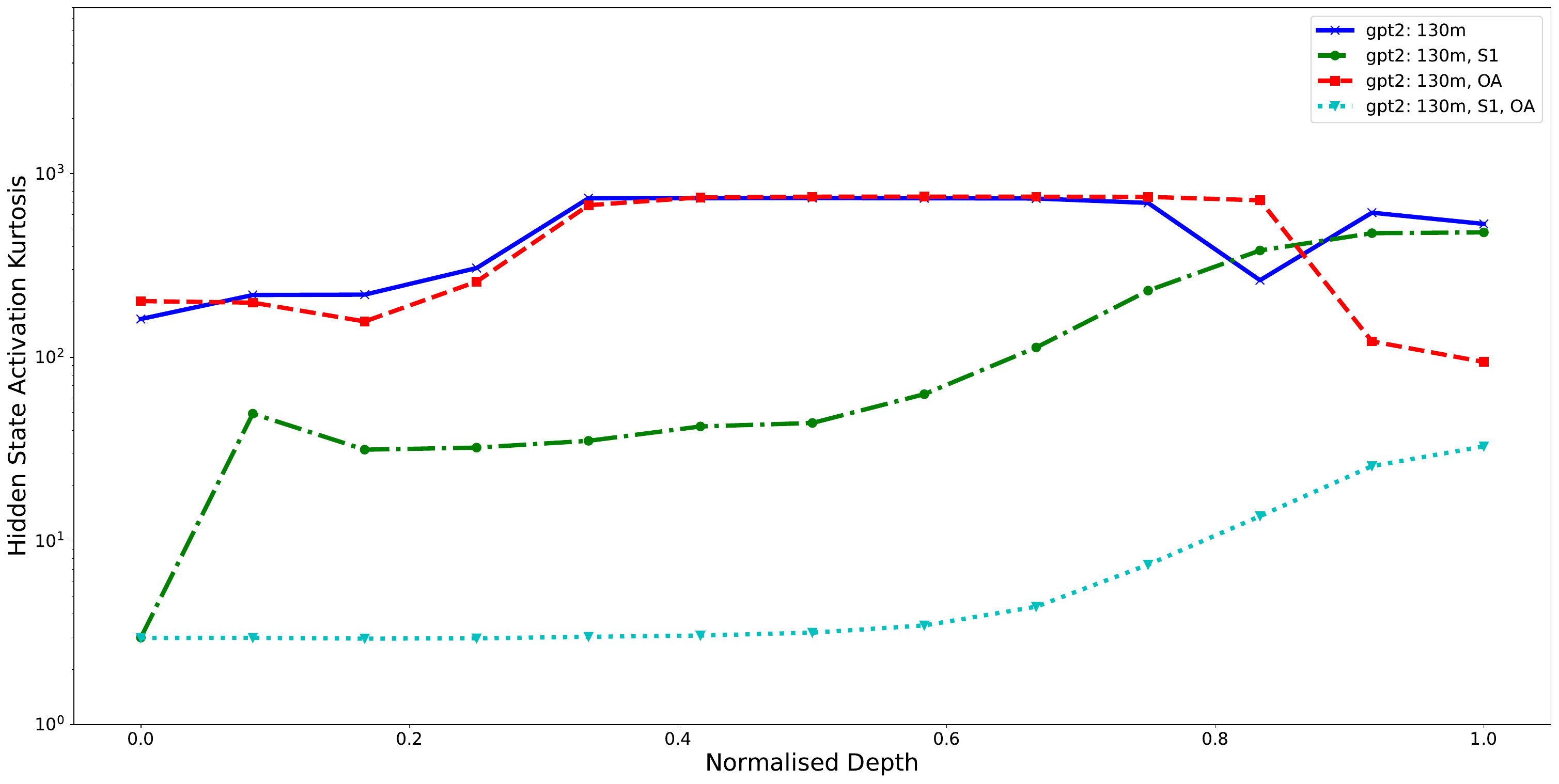}
    \caption{
        Layer-wise progression of activation kurtosis of the first token position for GPT2-130M.
        The x-axis is normalised to the range $[0, 1]$. S1/OA denote models trained with softmax-1 and/or OrthoAdam.
    }\label{fig:our_kurtosis_gpt2_130m}
\end{figure}

\subsubsection{GPT2-350M and GPT2-1.4B}
\begin{figure}[H]
    \centering
    \includegraphics[width=1.0\textwidth]{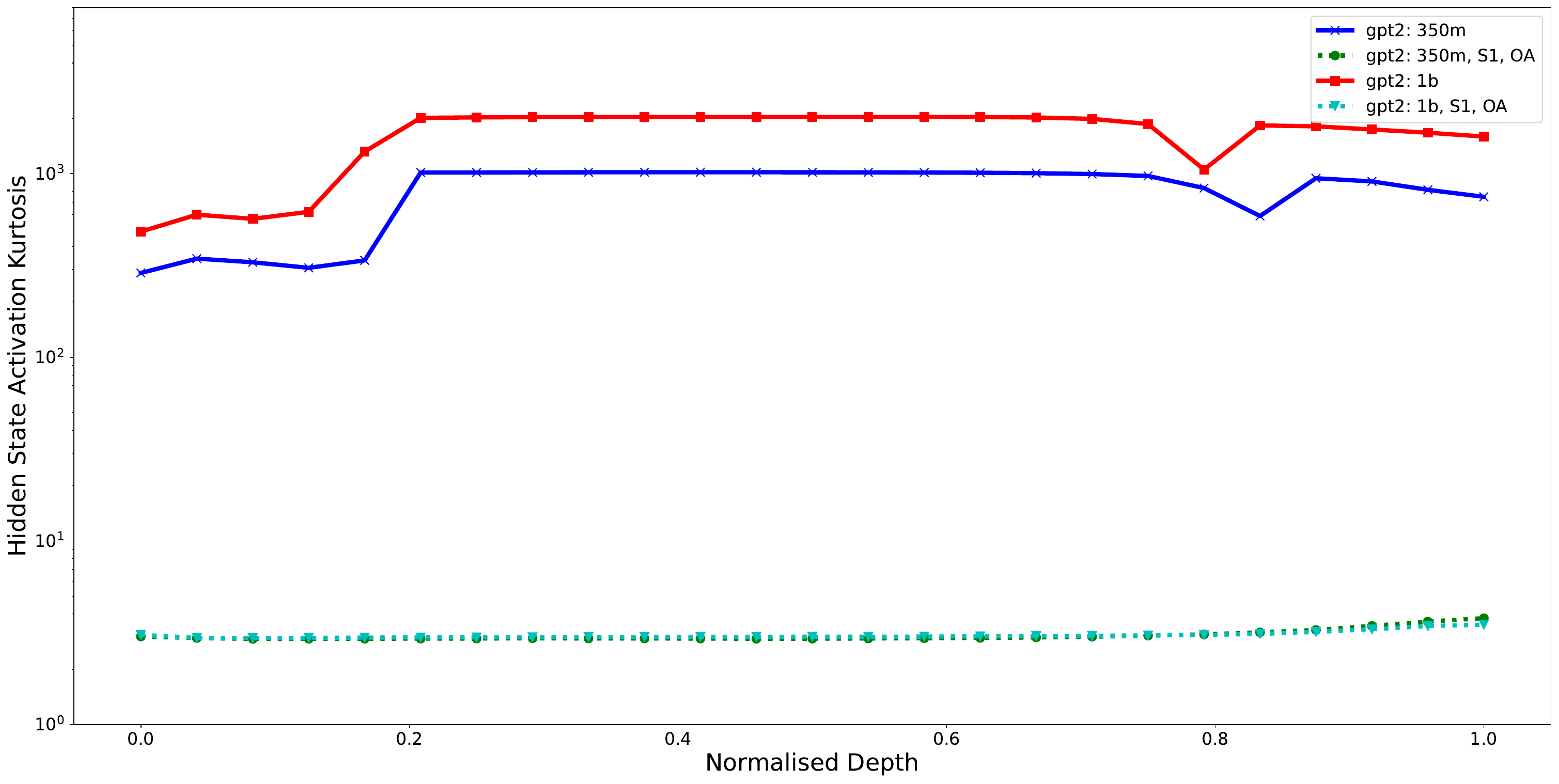}
    \caption{
        Layer-wise progression of activation kurtosis of the first token position for GPT2-350M and GPT2-1.4B.
        The x-axis is normalised to the range $[0, 1]$. S1/OA denote models trained with softmax-1 and/or OrthoAdam.
    }\label{fig:our_kurtosis_gpt2_350m}
\end{figure}

\subsubsection{Llama-130M}
\begin{figure}[H]
    \centering
    \includegraphics[width=1.0\textwidth]{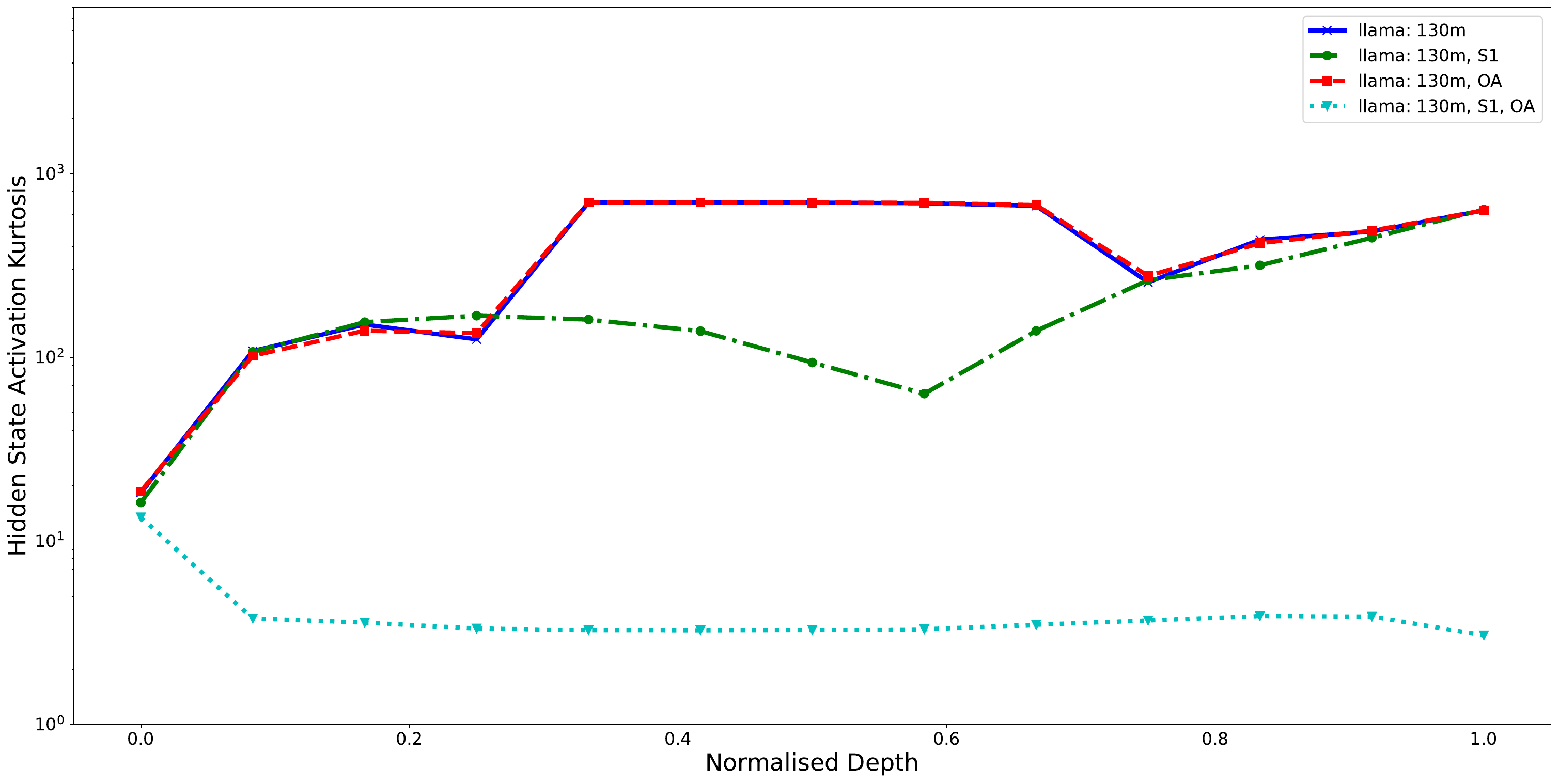}
    \caption{
        Layer-wise progression of activation kurtosis of the first token position for Llama-130M.
        The x-axis is normalised to the range $[0, 1]$. S1/OA denote models trained with softmax-1 and/or OrthoAdam.
    }\label{fig:our_kurtosis_llama_130m}
\end{figure}

\subsubsection{Popular Pretrained Models---GPT2 and Llama}
\begin{figure}[H]
    \centering
    \includegraphics[width=1.0\textwidth]{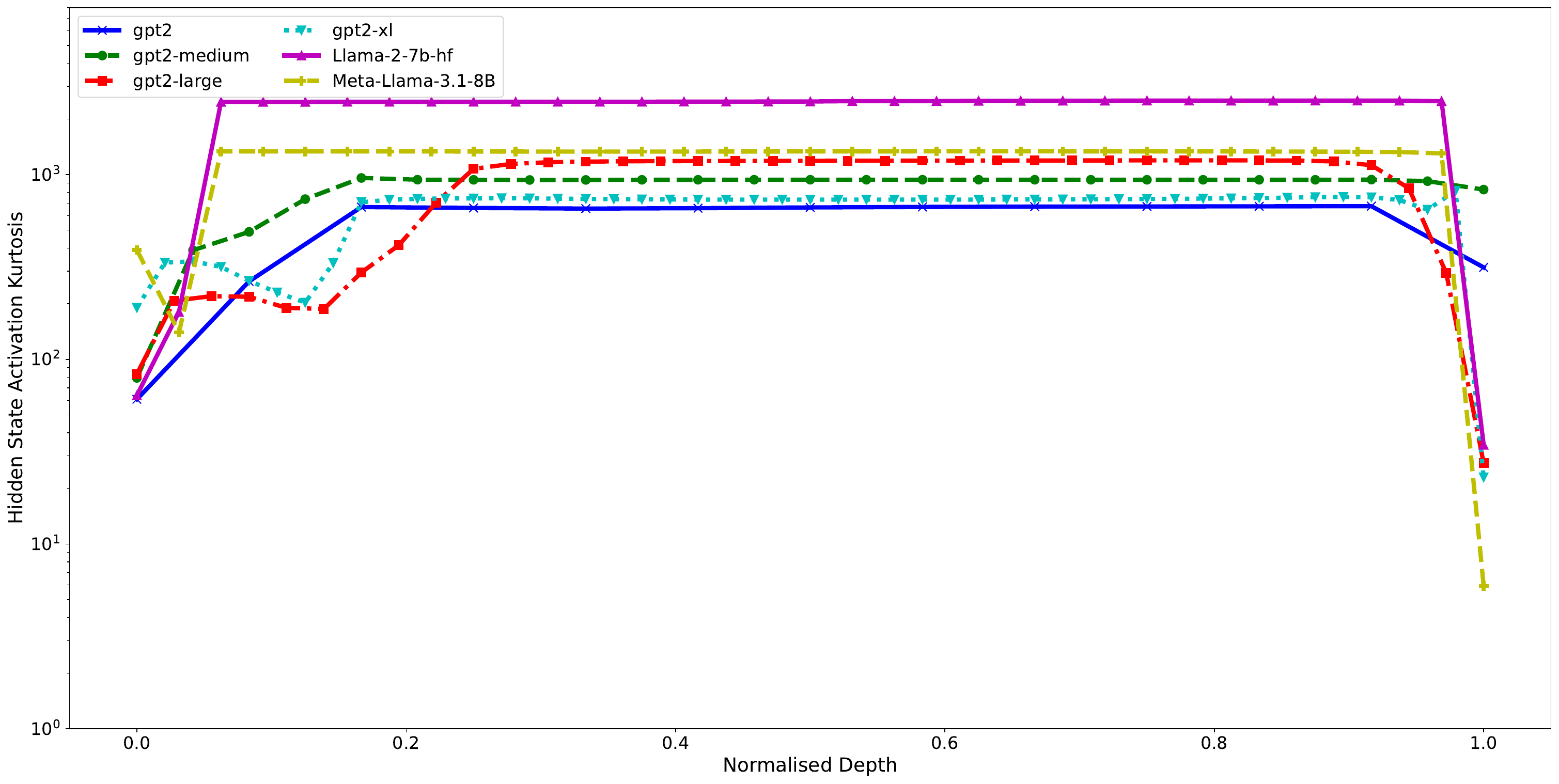}
    \caption{
        Layer-wise progression of activation kurtosis of the first token position for popular pretrained GPT2 and Llama models.
        The x-axis is normalised to the range $[0, 1]$.
    }\label{fig:pretrained_kurtosis}
\end{figure}

\subsection{$\ell_\infty$-Norm to $\ell_2$-Norm Ratio}\label{ssec:linf_l2_ratio}
The plots below show the progression of the $\ell_\infty$-norm to $\ell_2$-norm ratio across layers.
We observe that models trained without OrthoAdam exhibit a general trend of increasing ratio
as the hidden states progress through the layers. Once again as this ratio is maximal in the first hidden state,
we plot the ratio of the first hidden state only for brevity (as done for kurtosis).

The trends are similar to the kurtosis plots and so the same commentary applies.

\subsubsection{GPT2-60M}
\begin{figure}[H]
    \centering
    \includegraphics[width=1.0\textwidth]{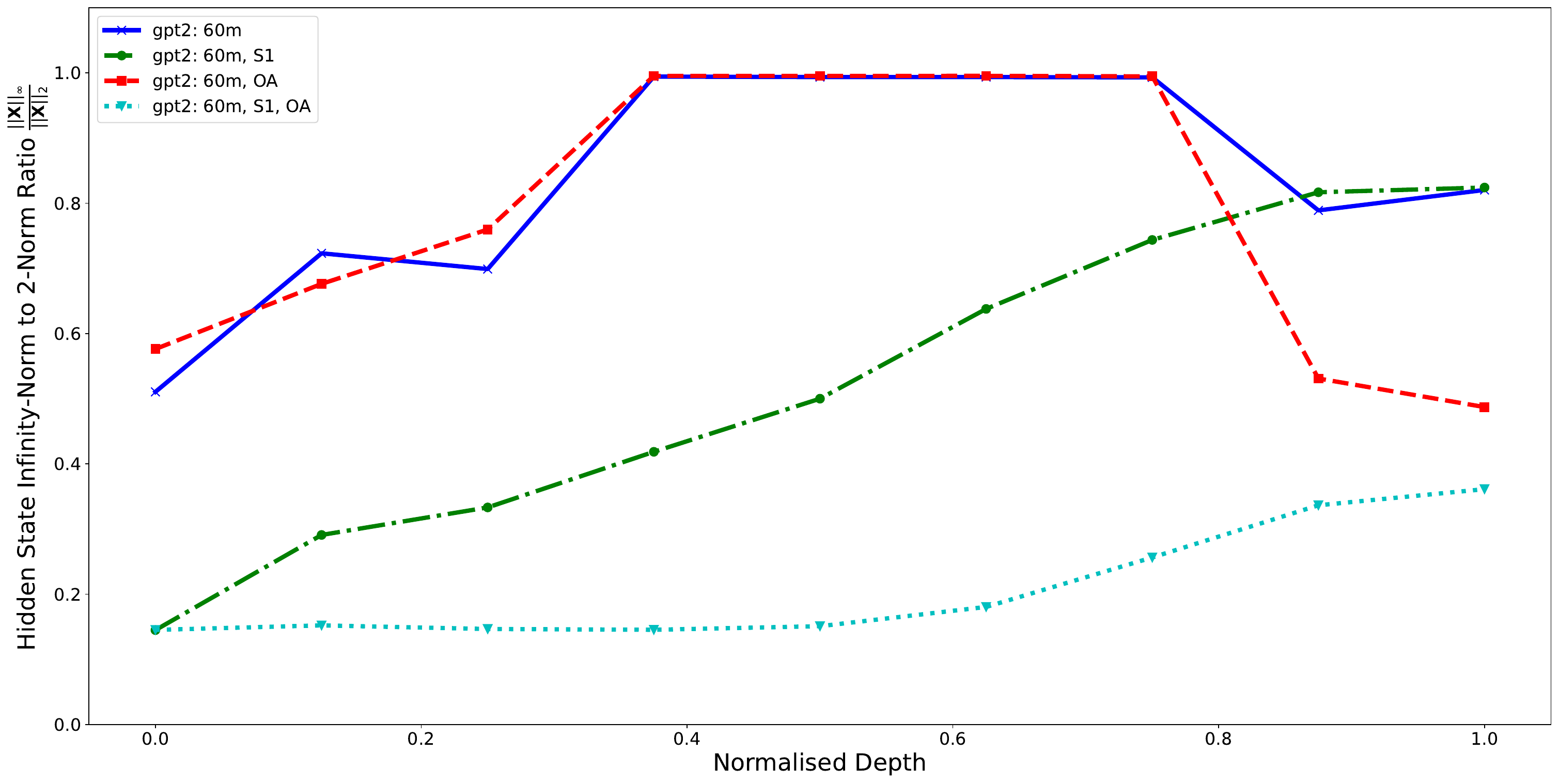}
    \caption{
        Layer-wise progression of the $\ell_\infty$-norm to $\ell_2$-norm ratio in the hidden states of the first token position for GPT2-60M.
        The x-axis is normalised to the range $[0, 1]$. S1/OA denote models trained with softmax-1 and/or OrthoAdam.
    }\label{fig:our_inf_2_gpt2_60m}
\end{figure}

\subsubsection{GPT2-130M}
\begin{figure}[H]
    \centering
    \includegraphics[width=1.0\textwidth]{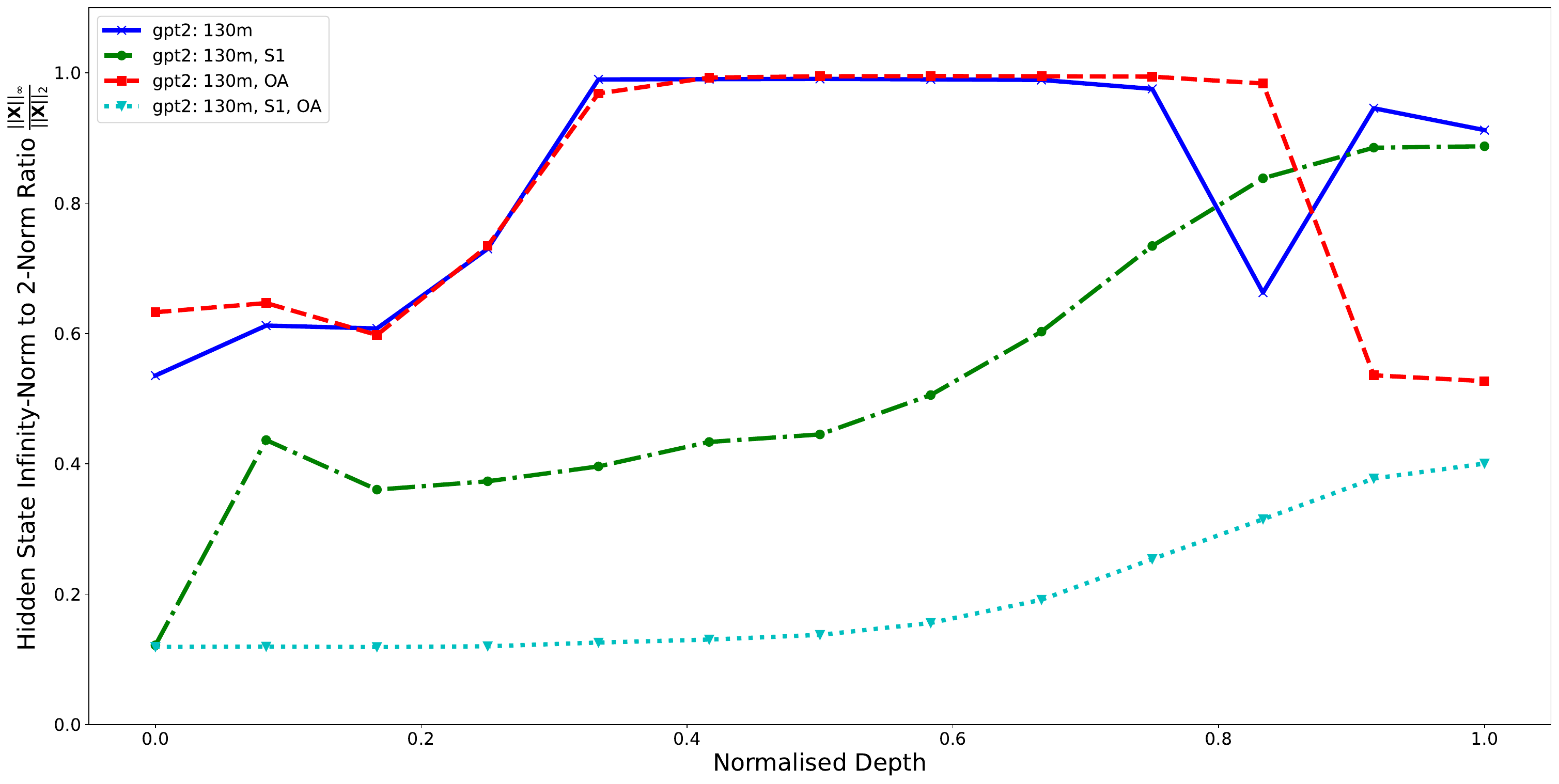}
    \caption{
        Layer-wise progression of the $\ell_\infty$-norm to $\ell_2$-norm ratio in the hidden states of the first token position for GPT2-130M.
        The x-axis is normalised to the range $[0, 1]$. S1/OA denote models trained with softmax-1 and/or OrthoAdam.
    }\label{fig:our_inf_2_gpt2_130m}
\end{figure}

\subsubsection{GPT2-350M and GPT2-1.4B}
\begin{figure}[H]
    \centering
    \includegraphics[width=1.0\textwidth]{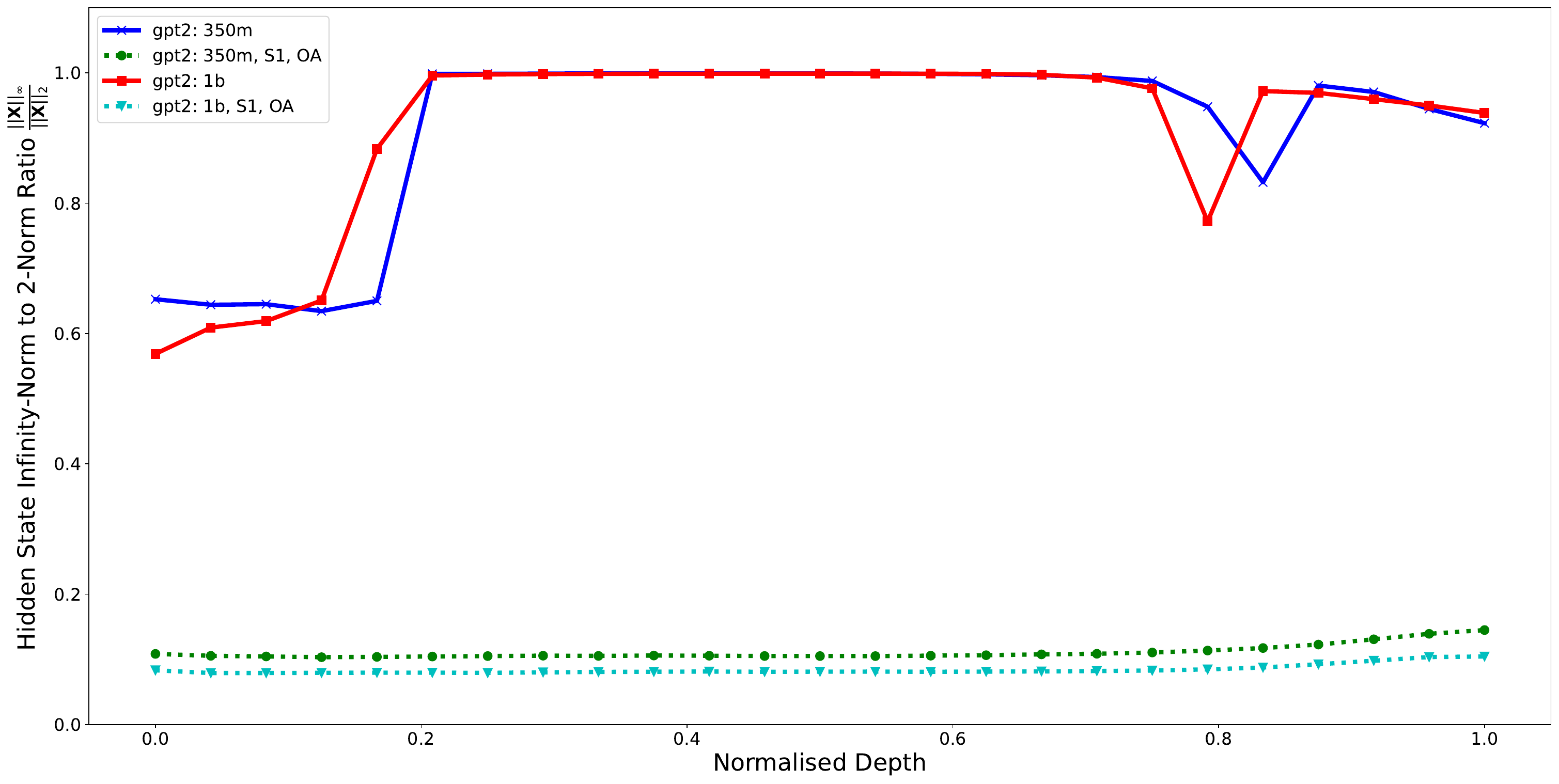}
    \caption{
        Layer-wise progression of the $\ell_\infty$-norm to $\ell_2$-norm ratio in the hidden states of the first token position for GPT2-350M and GPT2-1.4B.
        The x-axis is normalised to the range $[0, 1]$. S1/OA denote models trained with softmax-1 and/or OrthoAdam.
    }\label{fig:our_inf_2_gpt2_350m}
\end{figure}

\subsubsection{Llama-130M}
\begin{figure}[H]
    \centering
    \includegraphics[width=1.0\textwidth]{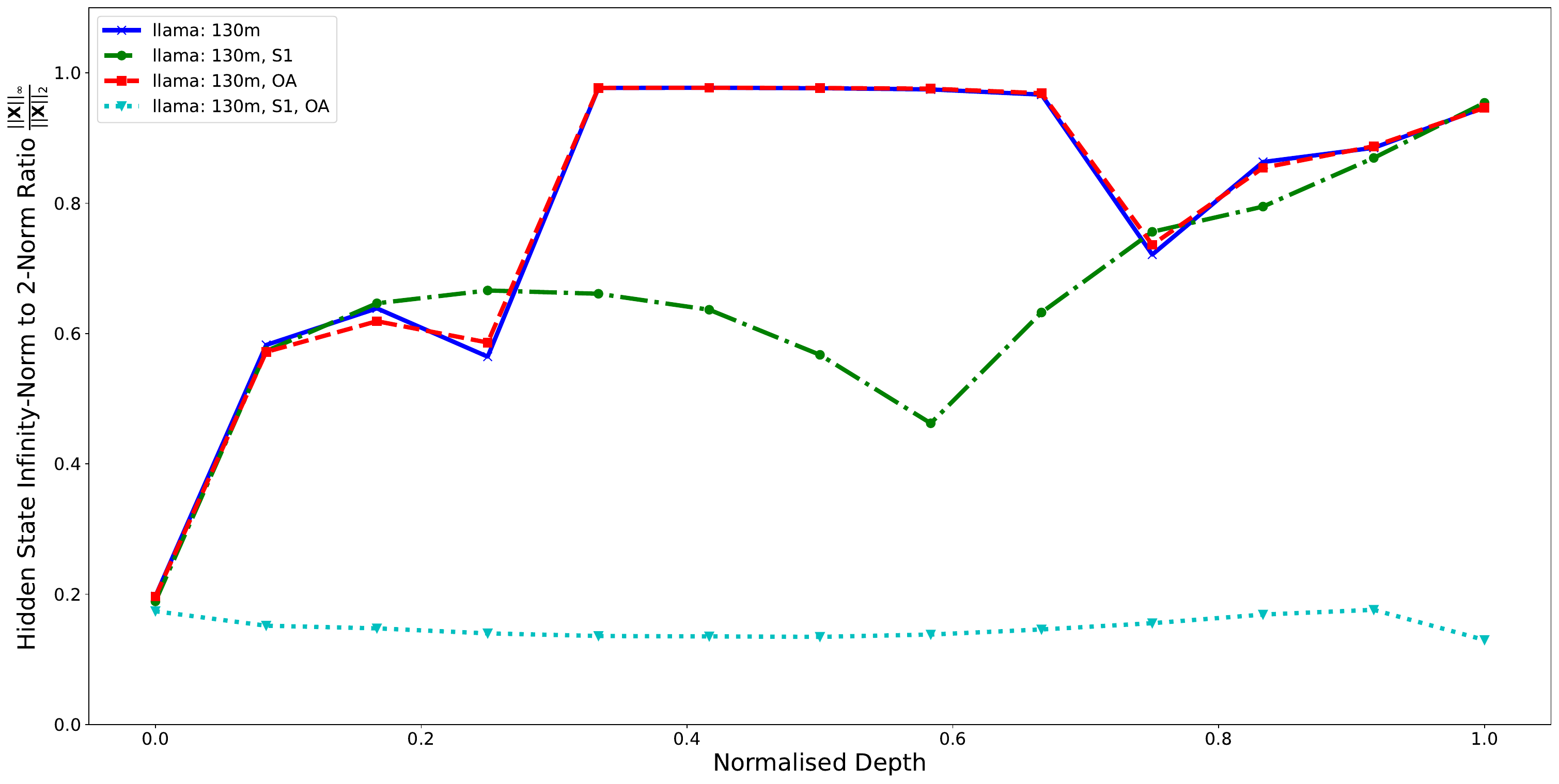}
    \caption{
        Layer-wise progression of the $\ell_\infty$-norm to $\ell_2$-norm ratio in the hidden states of the first token position for Llama-130M.
        The x-axis is normalised to the range $[0, 1]$. S1/OA denote models trained with softmax-1 and/or OrthoAdam.
    }\label{fig:our_inf_2_llama_130m}
\end{figure}

\subsubsection{Popular Pretrained Models---GPT2 and Llama}
\begin{figure}[H]
    \centering
    \includegraphics[width=1.0\textwidth]{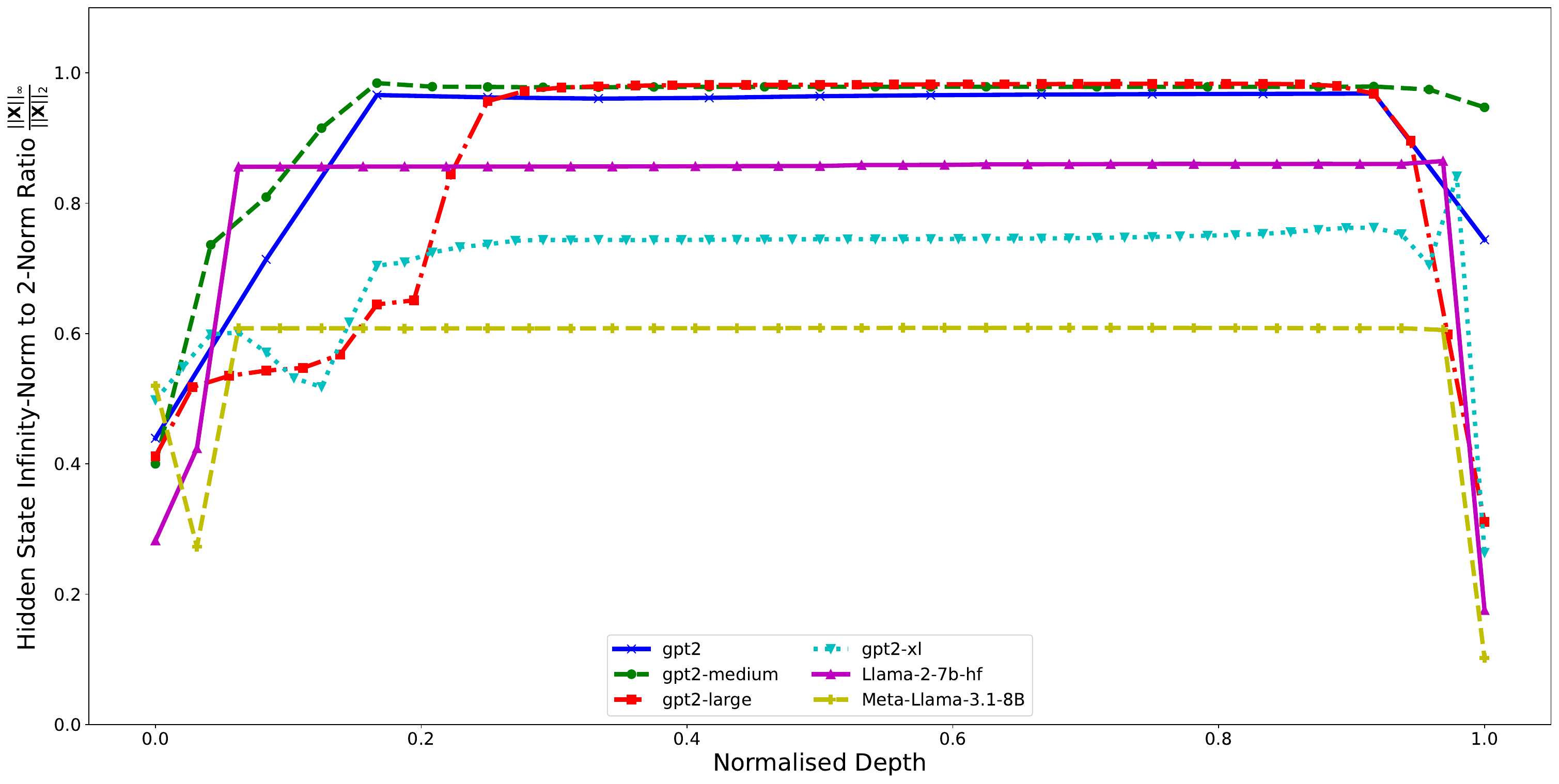}
    \caption{
        Layer-wise progression of the $\ell_\infty$-norm to $\ell_2$-norm ratio in the hidden states of the first token position for popular pretrained GPT2 and Llama models.
        The x-axis is normalised to the range $[0, 1]$.
    }\label{fig:pretrained_norm_ratio}
\end{figure}

To further clarify that in Transformer models the $\ell_\infty$-norm to $\ell_2$-norm ratio is a proxy for activation kurtosis,
we calculate the Pearson's correlation coefficients between the two metrics for all models in our main experimental results
from~\cref{tab:main_table_small} and public GPT2 and Llama models.
The results are shown in~\cref{tab:correlation_table_small}.
We find a strong positive correlation between the two metrics across all models which reinforces our intuition
that using orthogonal matrices to transform the gradients in the optimiser is an effective way to mitigate the emergence of large activation values,
as an orthogonal transformation can reduce the $\ell_\infty$-norm of a vector substantially for a given $\ell_2$-norm.

\begin{table}[h!]
    \centering
    \scriptsize
    \begin{adjustbox}{width=\linewidth}
    \begin{tabular}{c|c|c|c|cc}
    \toprule
    \multirow{2}{*}{Model}         & \multirow{2}{*}{\#Parameters} & \multirow{2}{*}{Softmax+1?} & \multirow{2}{*}{OrthoAdam?} &  \multicolumn{2}{c}{Correlation of Kurtosis to Norm Ratio} \\
                                   &                               &                             &                             &  First Token        & Other Tokens                         \\ \midrule
    \multirow{12}{*}{GPT2 (Ours)}  & \multirow{4}{*}{60M}          &                             &                             &  0.961              & 0.932                                \\
                                   &                               & \checkmark                  &                             &  0.932              & 0.934                                \\
                                   &                               &                             & \checkmark                  &  0.986              & 0.972                                \\
                                   &                               & \checkmark                  & \checkmark                  &  0.968              & 0.970                                \\ \cmidrule(l){2-6} 
                                   & \multirow{4}{*}{130M}         &                             &                             &  0.988              & 0.932                                \\
                                   &                               & \checkmark                  &                             &  0.927              & 0.924                                \\
                                   &                               &                             & \checkmark                  &  0.992              & 0.962                                \\
                                   &                               & \checkmark                  & \checkmark                  &  0.935              & 0.953                                \\ \cmidrule(l){2-6} 
                                   & \multirow{2}{*}{350M}         &                             &                             &  0.990              & 0.929                                \\
                                   &                               & \checkmark                  & \checkmark                  &  0.998              & 0.997                                \\ \cmidrule(l){2-6} 
                                   & \multirow{2}{*}{1.4B}         &                             &                             &  0.988              & 0.952                                \\
                                   &                               & \checkmark                  & \checkmark                  &  0.994              & 0.995                                \\ \midrule
    \multirow{4}{*}{Llama2 (Ours)} & \multirow{4}{*}{130M}         &                             &                             &  0.931              & 0.903                                \\
                                   &                               & \checkmark                  &                             &  0.864              & 0.877                                \\
                                   &                               &                             & \checkmark                  &  0.931              & 0.905                                \\
                                   &                               & \checkmark                  & \checkmark                  &  0.560              & 0.975                                \\ \midrule
    GPT2 (Public)                  & 137M                          &                             &                             &  0.985              & 0.944                                \\
    GPT2-Medium (Public)           & 350M                          &                             &                             &  0.969              & 0.846                                \\
    GPT2-Large (Public)            & 812M                          &                             &                             &  0.985              & 0.896                                \\
    GPT2-XL (Public)               & 1.6B                          &                             &                             &  0.956              & 0.939                                \\
    Llama2-7B (Public)             & 6.7B                          &                             &                             &  0.987              & 0.902                                \\
    Llama3.1-8B (Public)           & 8B                            &                             &                             &  0.928              & 0.915                                \\ \bottomrule
    \end{tabular}
    \end{adjustbox}
    \caption{
        Correlation of the kurtosis and norm-ratio of the hidden states of our trained models
        and popular pretrained models.
    }\label{tab:correlation_table_small}
\end{table}

\subsection{Maximum Absolute Activation}
Finally, we examine the progression of the maximum absolute activation across layers.

\subsubsection{GPT2-60M}
\begin{figure}[H]
    \centering
    \includegraphics[width=1.0\textwidth]{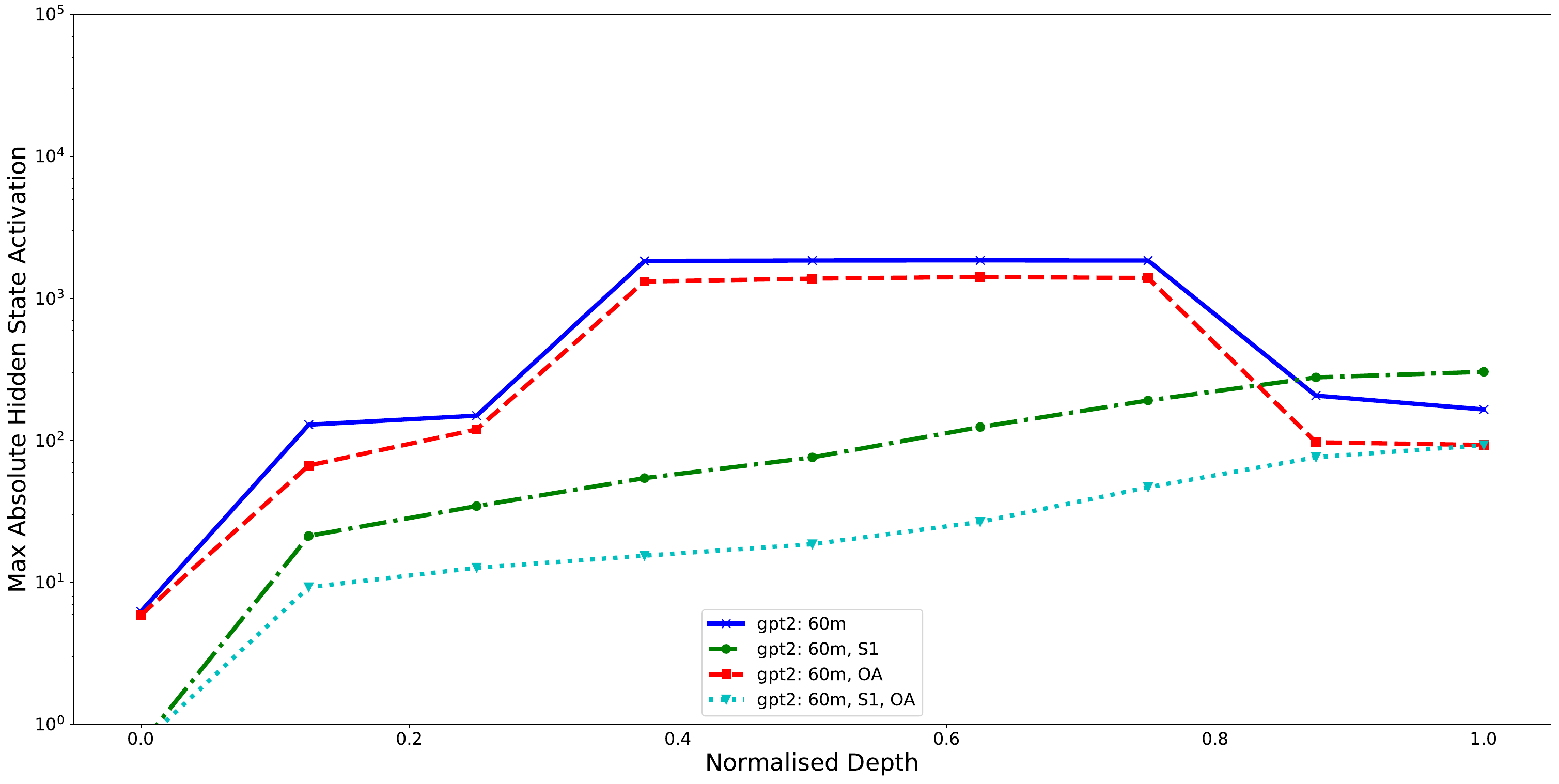}
    \caption{
        Layer-wise progression of the maximum absolute activation in the hidden states of the first token position for GPT2-60M.
        The x-axis is normalised to the range $[0, 1]$. S1/OA denote models trained with softmax-1 and/or OrthoAdam.
    }\label{fig:our_max_gpt2_60m}
\end{figure}

\subsubsection{GPT2-130M}
\begin{figure}[H]
    \centering
    \includegraphics[width=1.0\textwidth]{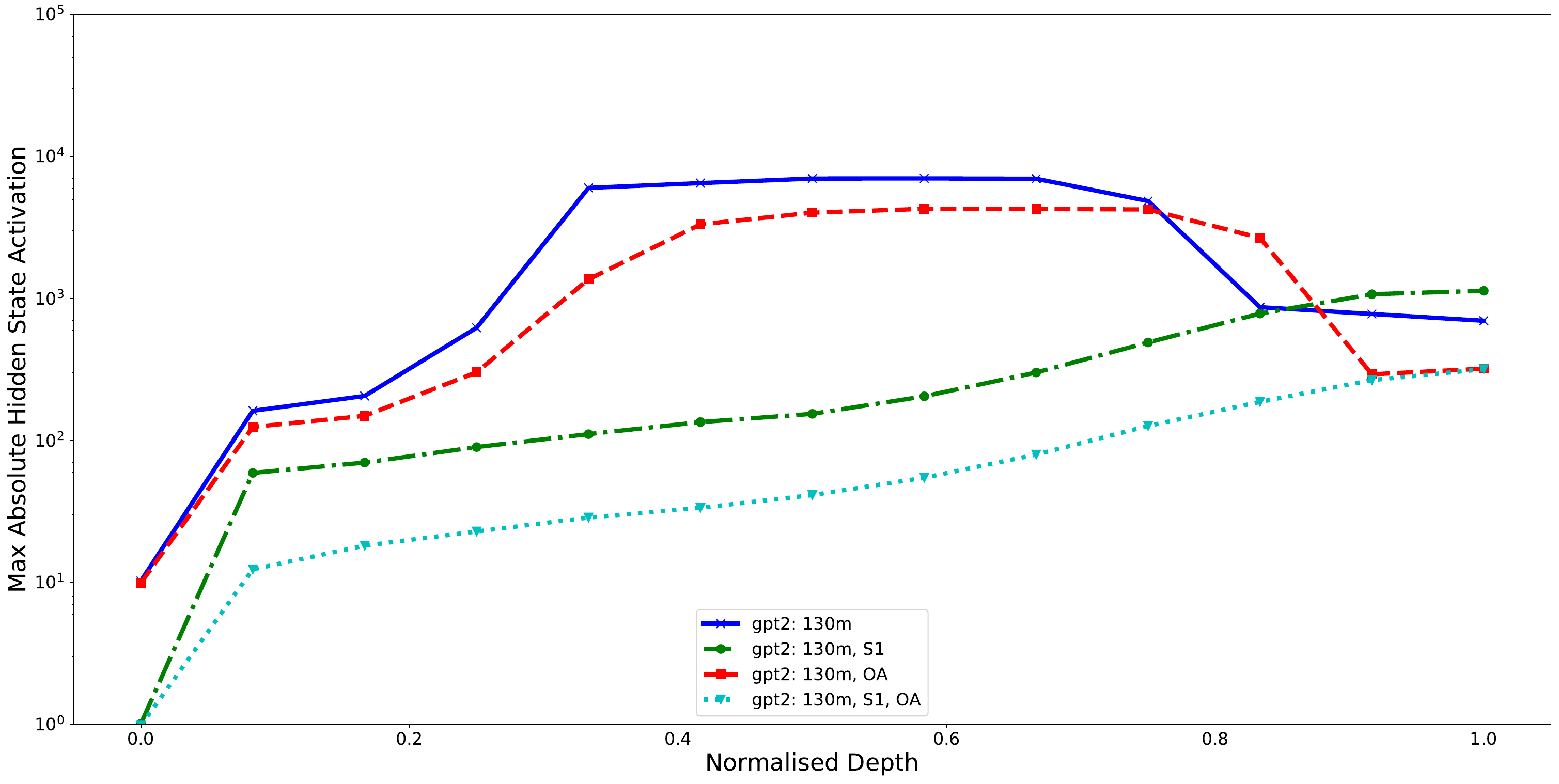}
    \caption{
        Layer-wise progression of the maximum absolute activation in the hidden states of the first token position for GPT2-130M.
        The x-axis is normalised to the range $[0, 1]$. S1/OA denote models trained with softmax-1 and/or OrthoAdam.
    }\label{fig:our_max_gpt2_130m}
\end{figure}

\subsubsection{GPT2-350M and GPT2-1.4B}
\begin{figure}[H]
    \centering
    \includegraphics[width=1.0\textwidth]{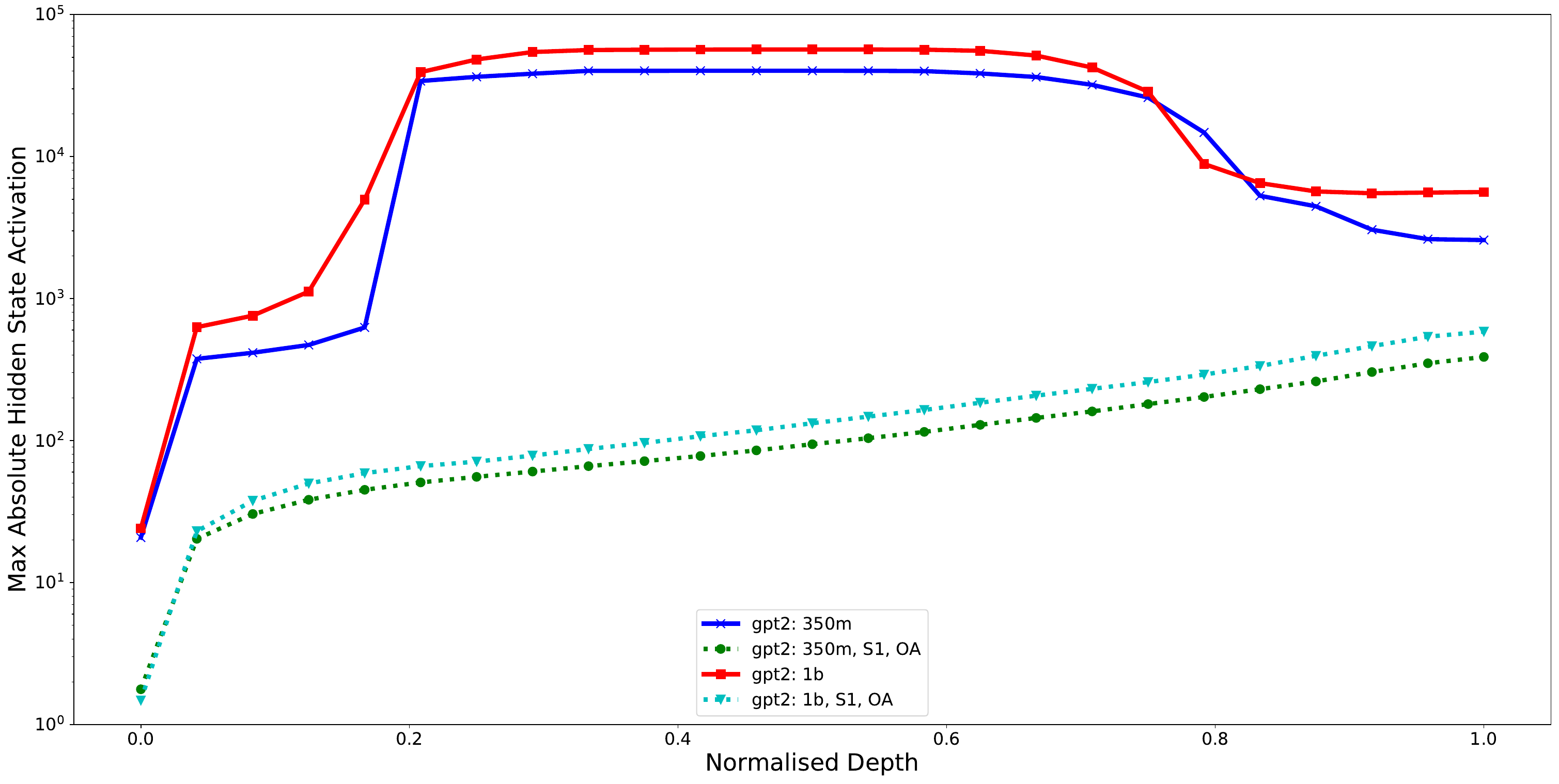}
    \caption{
        Layer-wise progression of the maximum absolute activation in the hidden states of the first token position for GPT2-350M and GPT2-1.4B.
        The x-axis is normalised to the range $[0, 1]$. S1/OA denote models trained with softmax-1 and/or OrthoAdam.
    }\label{fig:our_max_gpt2_350m}
\end{figure}

\subsubsection{Llama-130M}
\begin{figure}[H]
    \centering
    \includegraphics[width=1.0\textwidth]{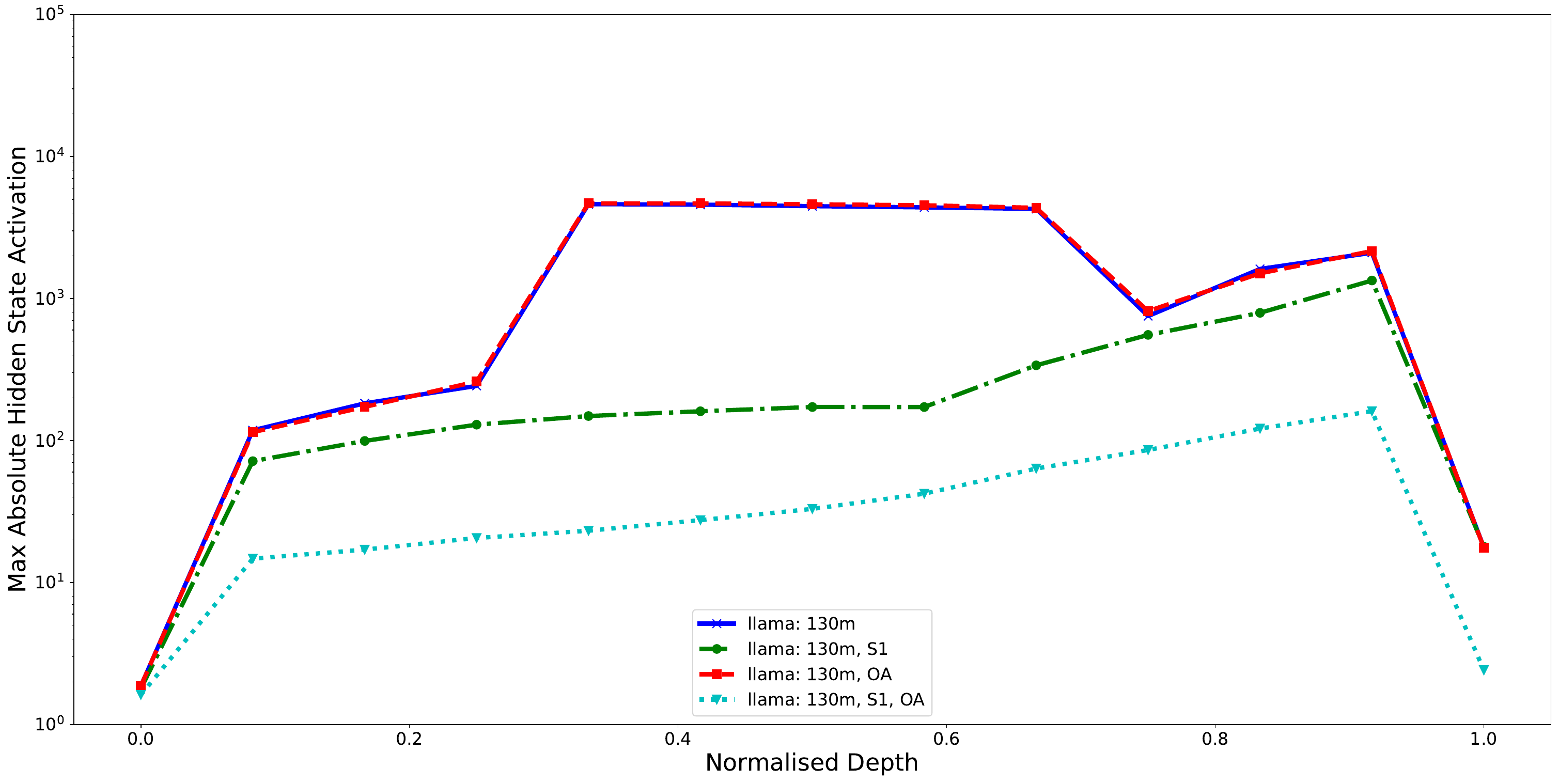}
    \caption{
        Layer-wise progression of the maximum absolute activation in the hidden states of the first token position for Llama-130M.
        The x-axis is normalised to the range $[0, 1]$. S1/OA denote models trained with softmax-1 and/or OrthoAdam.
    }\label{fig:our_max_llama_130m}
\end{figure}

\subsubsection{Popular Pretrained Models---GPT2 and Llama}
\begin{figure}[H]
    \centering
    \includegraphics[width=1.0\textwidth]{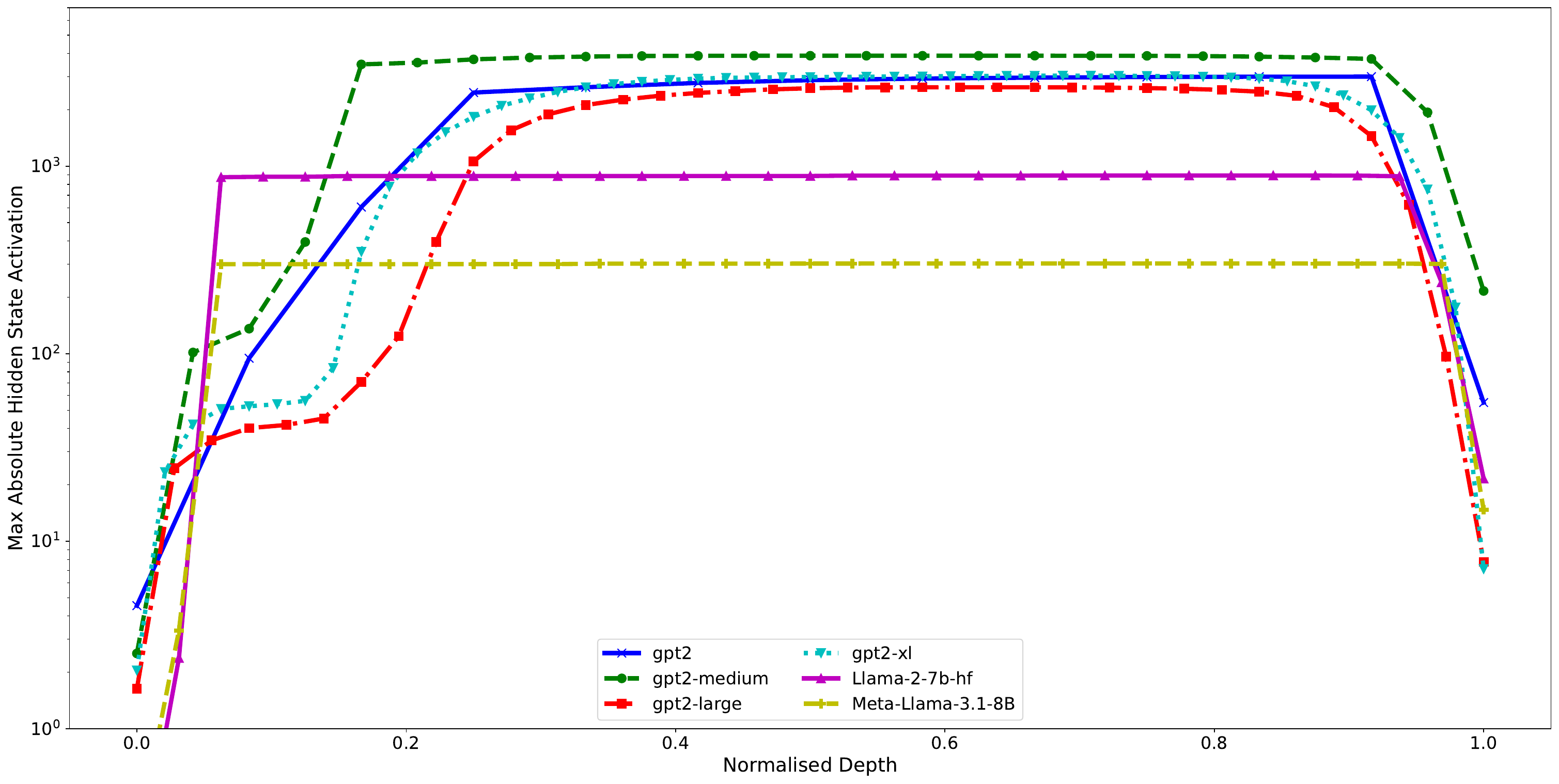}
    \caption{
        Layer-wise progression of the maximum absolute activation in the hidden states of the first token position for popular pretrained GPT2 and Llama models.
        The x-axis is normalised to the range $[0, 1]$.
    }\label{fig:pretrained_max}
\end{figure}
\newpage
\section{Hidden States of Pretrained Models}\label{sec:pretrained_hidden}
In this section, we present the progression of hidden states of popular pretrained models.
This shows how models establish outlier activations and how they persist in the same feature dimensions
across layers. For each model we show the \emph{absolute} activation values in the features containing the
largest activations.
We show the mean across layers, the first layer,~$\frac{1}{4}$ and~$\frac{3}{4}$ of the layers.

\begin{figure}[H]
    \centering
    \includegraphics[width=1.0\textwidth]{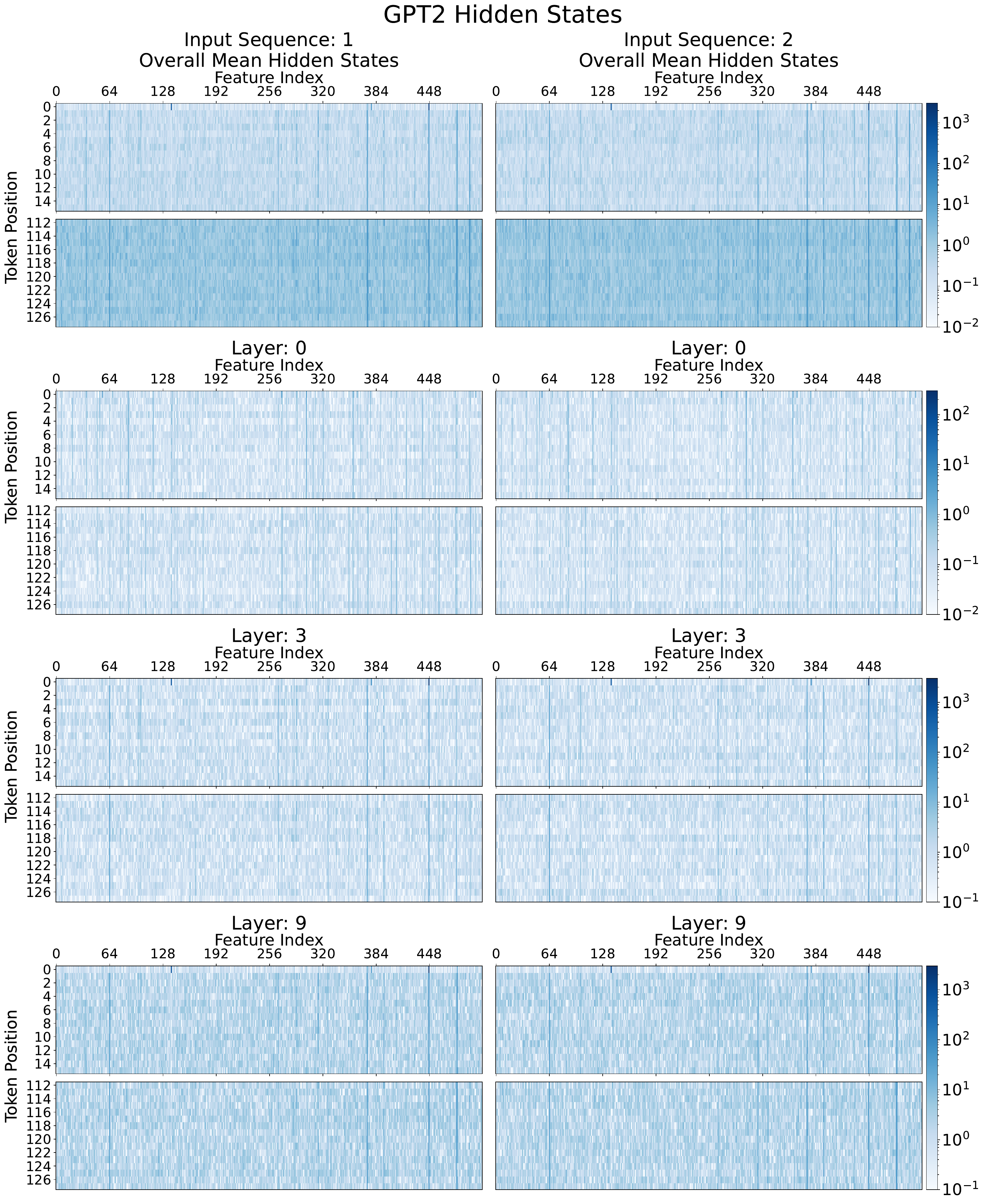}
    \caption{
        Example hidden state plots for a GPT2-Small model.
    }\label{fig:gpt2_hidden_states}
\end{figure}

\begin{figure}[H]
    \centering
    \includegraphics[width=1.0\textwidth]{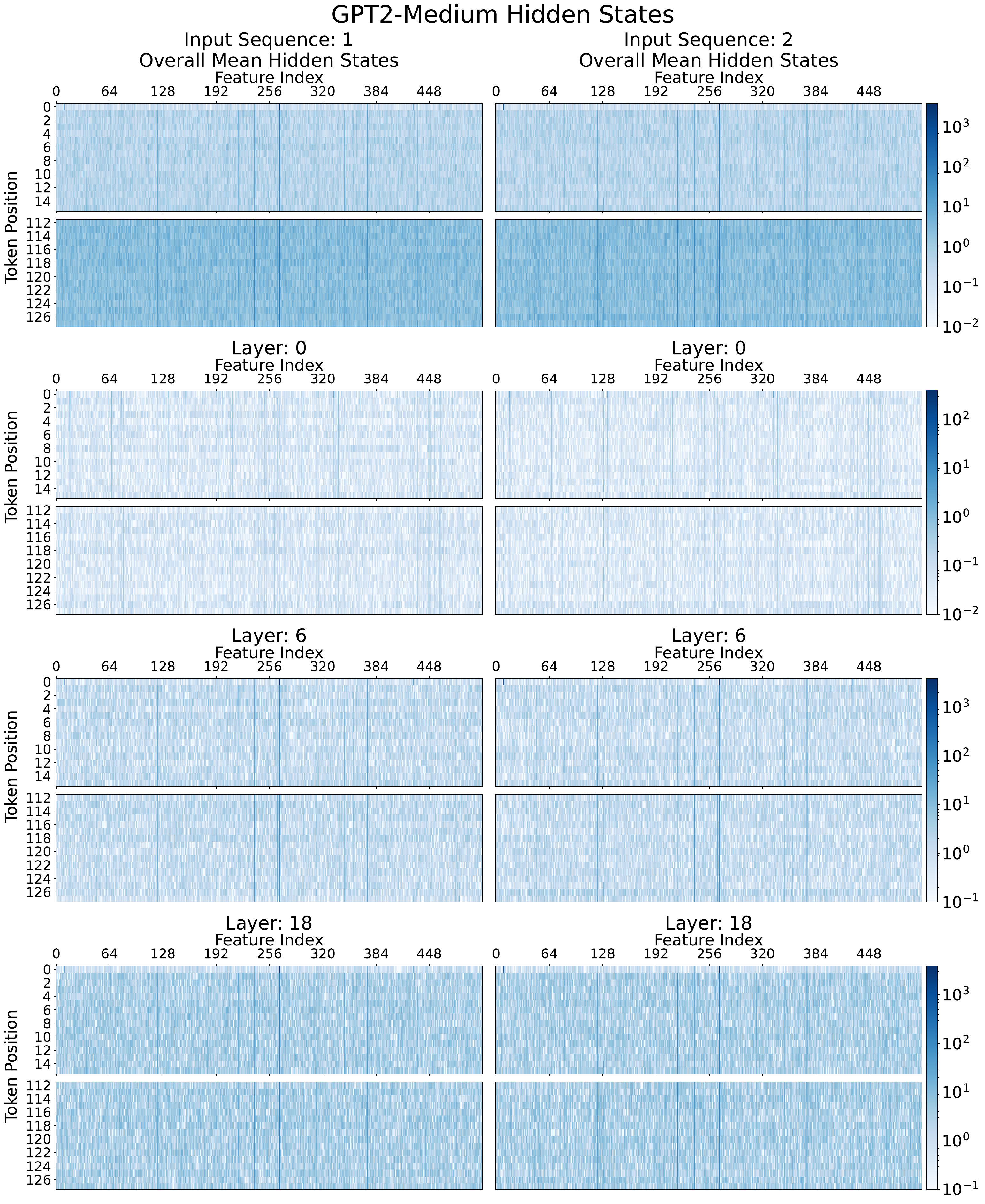}
    \caption{
        Example hidden state plots for a GPT2-Medium model.
    }\label{fig:gpt2-medium_hidden_states}
\end{figure}

\begin{figure}[H]
    \centering
    \includegraphics[width=1.0\textwidth]{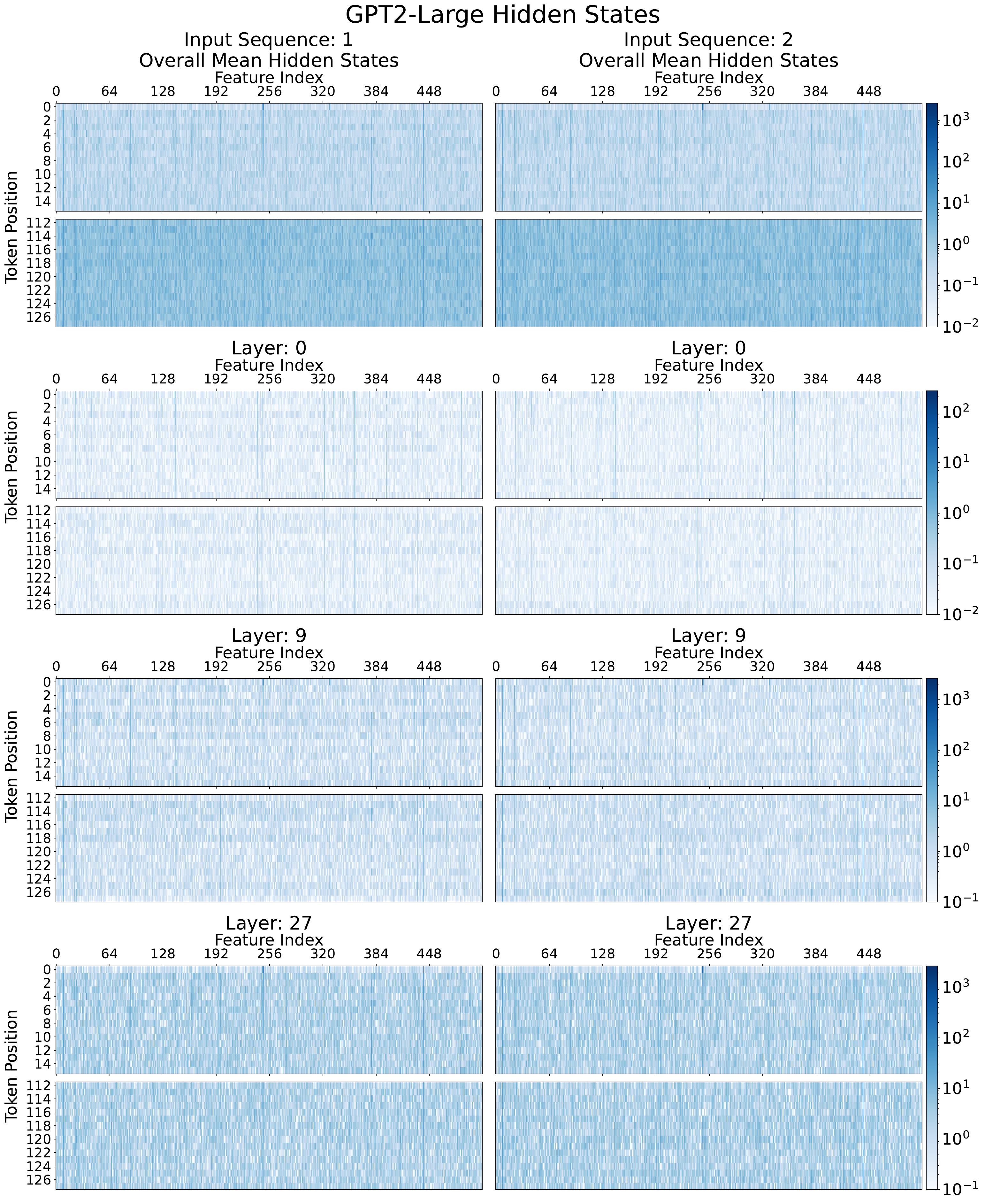}
    \caption{
        Example hidden state plots for a GPT2-Large model.
    }\label{fig:gpt2-large_hidden_states}
\end{figure}

\begin{figure}[H]
    \centering
    \includegraphics[width=1.0\textwidth]{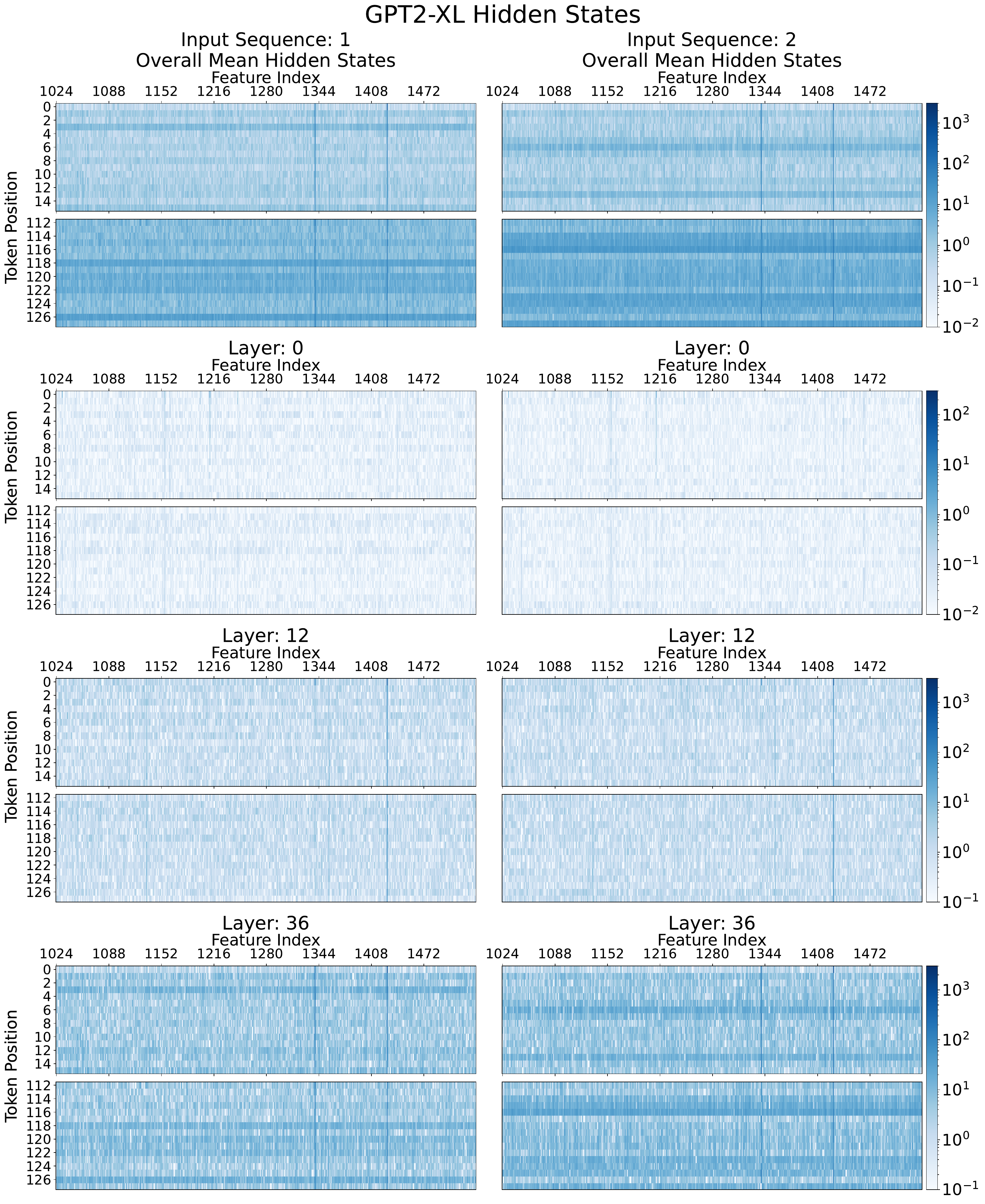}
    \caption{
        Example hidden state plots for a GPT2-XL model.
    }\label{fig:gpt2-xl_hidden_states}
\end{figure}

\begin{figure}[H]
    \centering
    \includegraphics[width=1.0\textwidth]{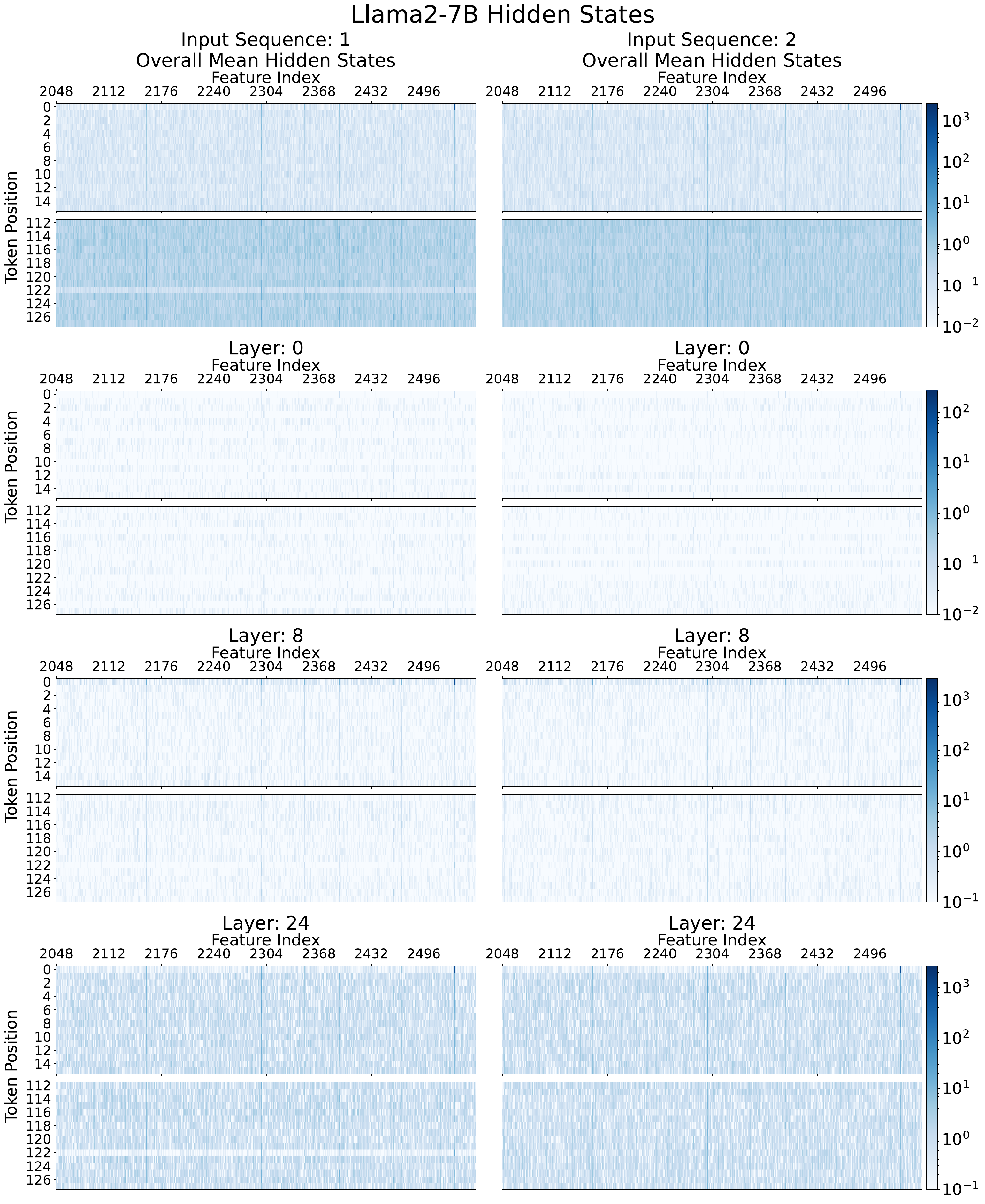}
    \caption{
        Example hidden state plots for a Llama2-7B model.
    }\label{fig:llama2-7b-hf_hidden_states}
\end{figure}

\begin{figure}[H]
    \centering
    \includegraphics[width=1.0\textwidth]{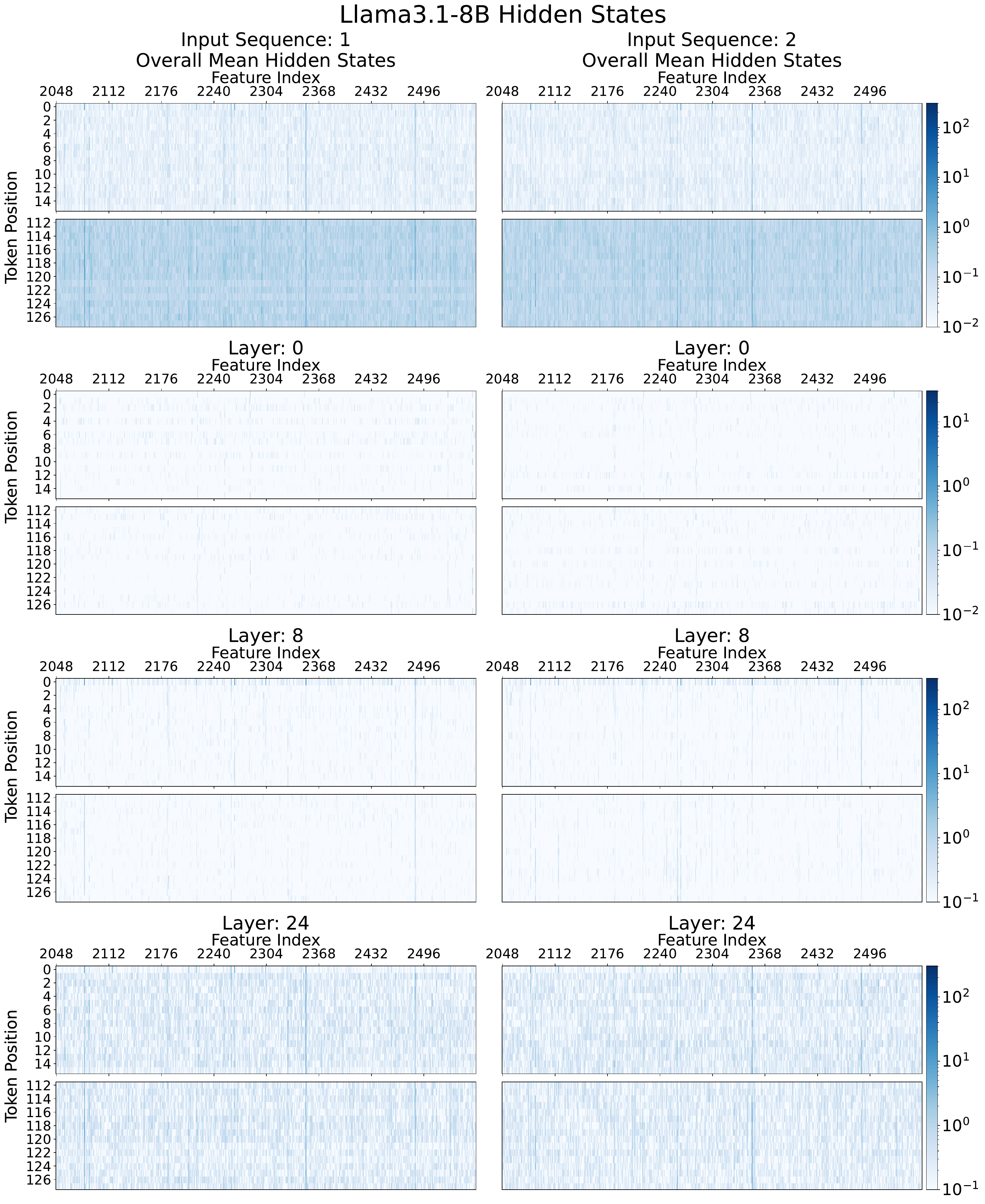}
    \caption{
        Example hidden state plots for a Llama3.1-8B model.
    }\label{fig:llama3-8b_hidden_states}
\end{figure}

\begin{figure}[H]
    \centering
    \includegraphics[width=1.0\textwidth]{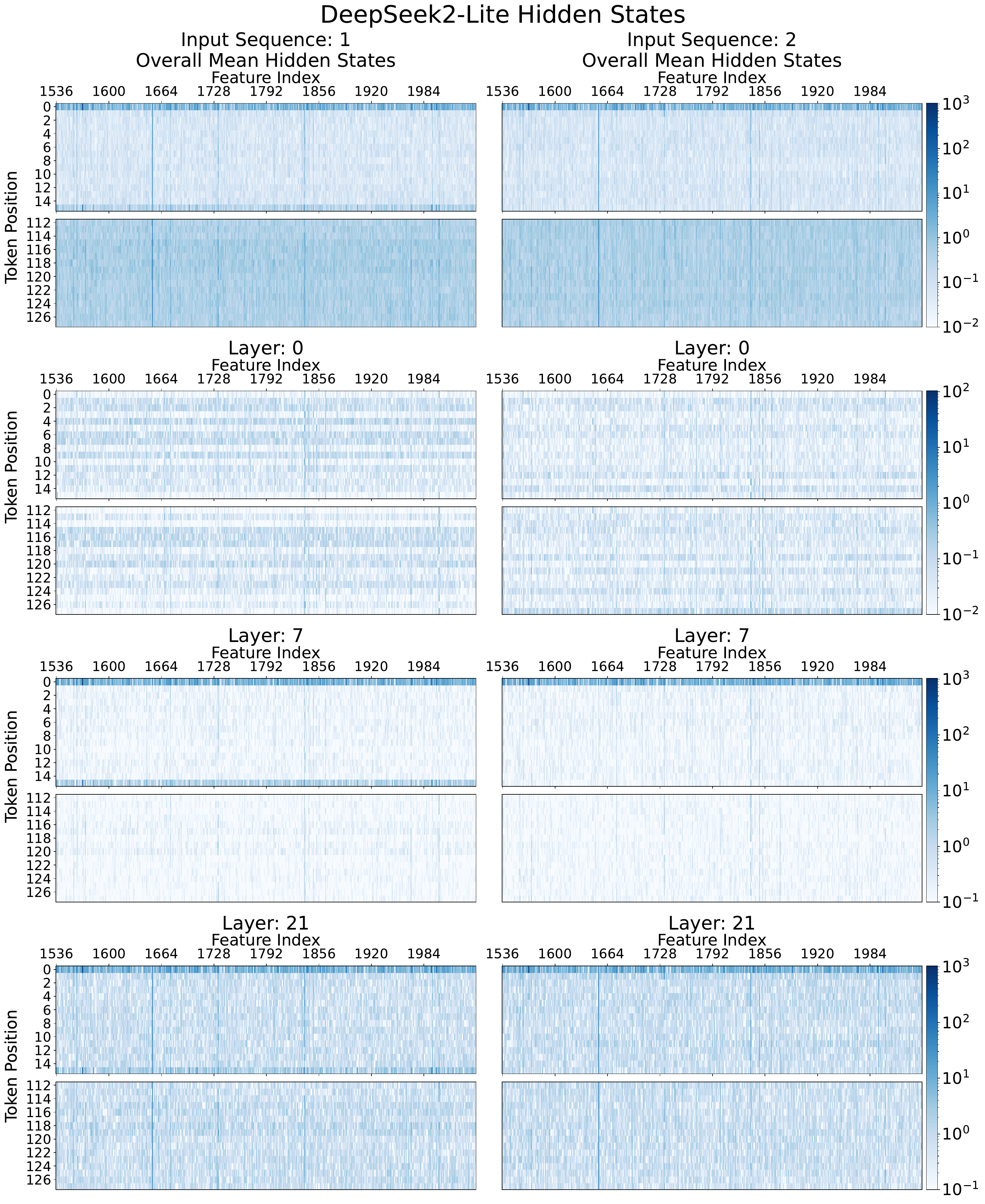}
    \caption{
        Example hidden state plots for a DeepSeekv2-Lite model.
    }\label{fig:deepseek2-lite_hidden_states}
\end{figure}

\newpage
\section{Attention Maps of Pretrained Models}\label{sec:pretrained_attention}
In this section, we present the attention maps of popular pretrained models.
This shows how models establish attention patterns and how they persist after initial layers.
This shows that generally after the first or second layer, first token attention dominance is highly
established and persists across layers.

We show the mean across layers, the first layer,~$\frac{1}{4}$ and~$\frac{3}{4}$ of the
layers---averaging over all heads in each case.

\begin{figure}[H]
    \centering
    \includegraphics[width=0.65\textwidth]{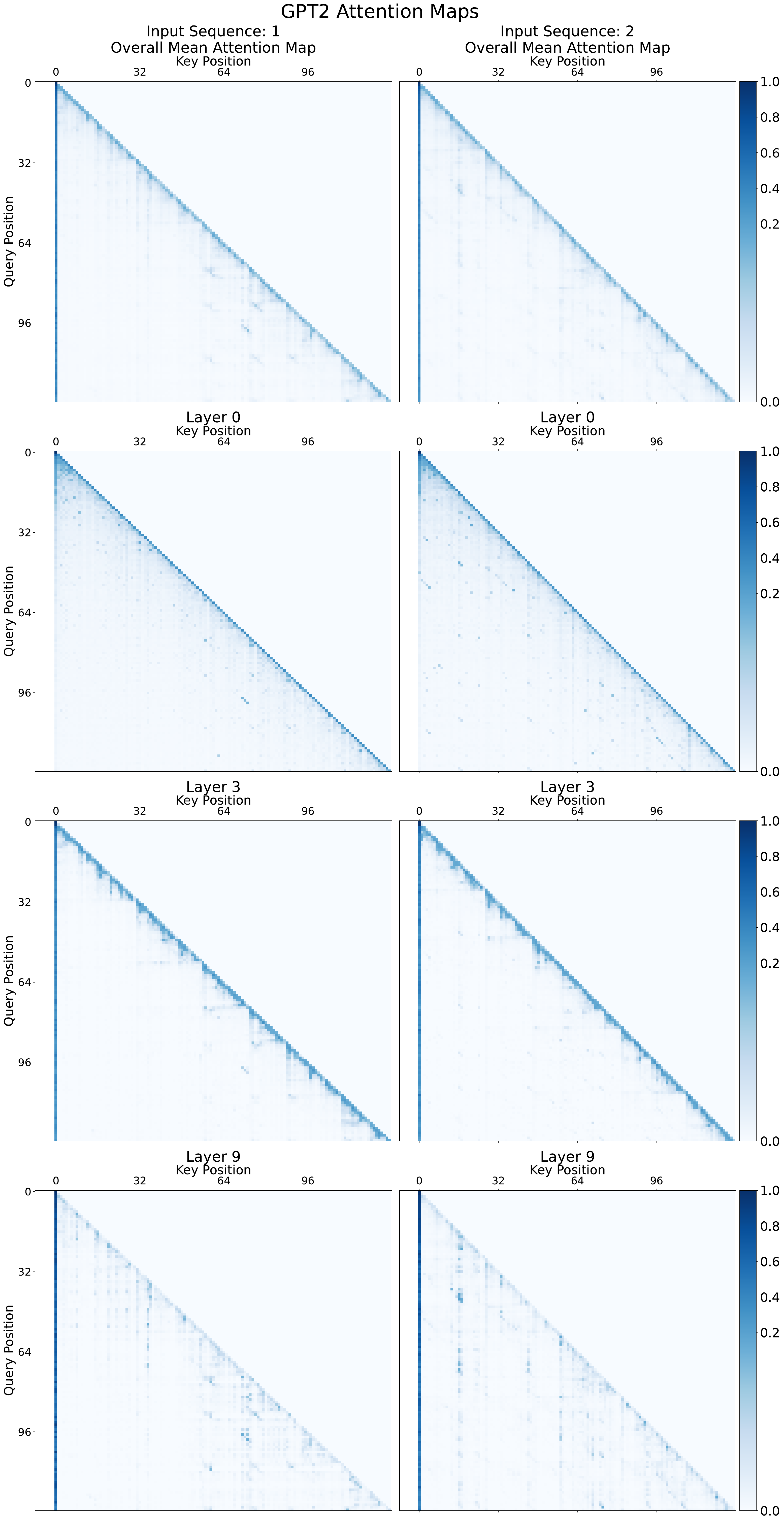}
    \caption{
        Example attention maps for a GPT2-Small model.
    }\label{fig:gpt2_attention_maps}
\end{figure}

\begin{figure}[H]
    \centering
    \includegraphics[width=0.8\textwidth]{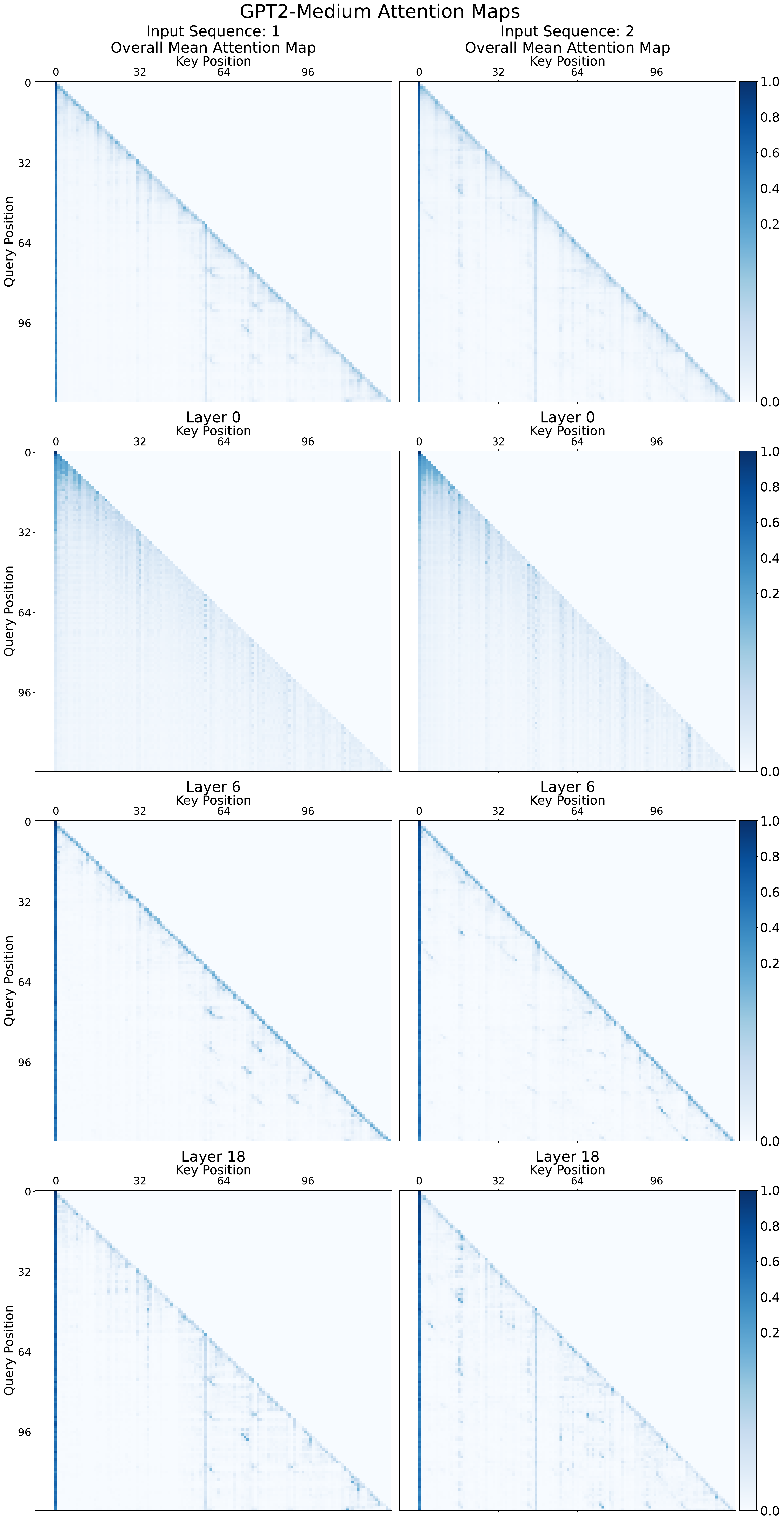}
    \caption{
        Example attention maps for a GPT2-Medium model.
    }\label{fig:gpt2-medium_attention_maps}
\end{figure}

\begin{figure}[H]
    \centering
    \includegraphics[width=0.8\textwidth]{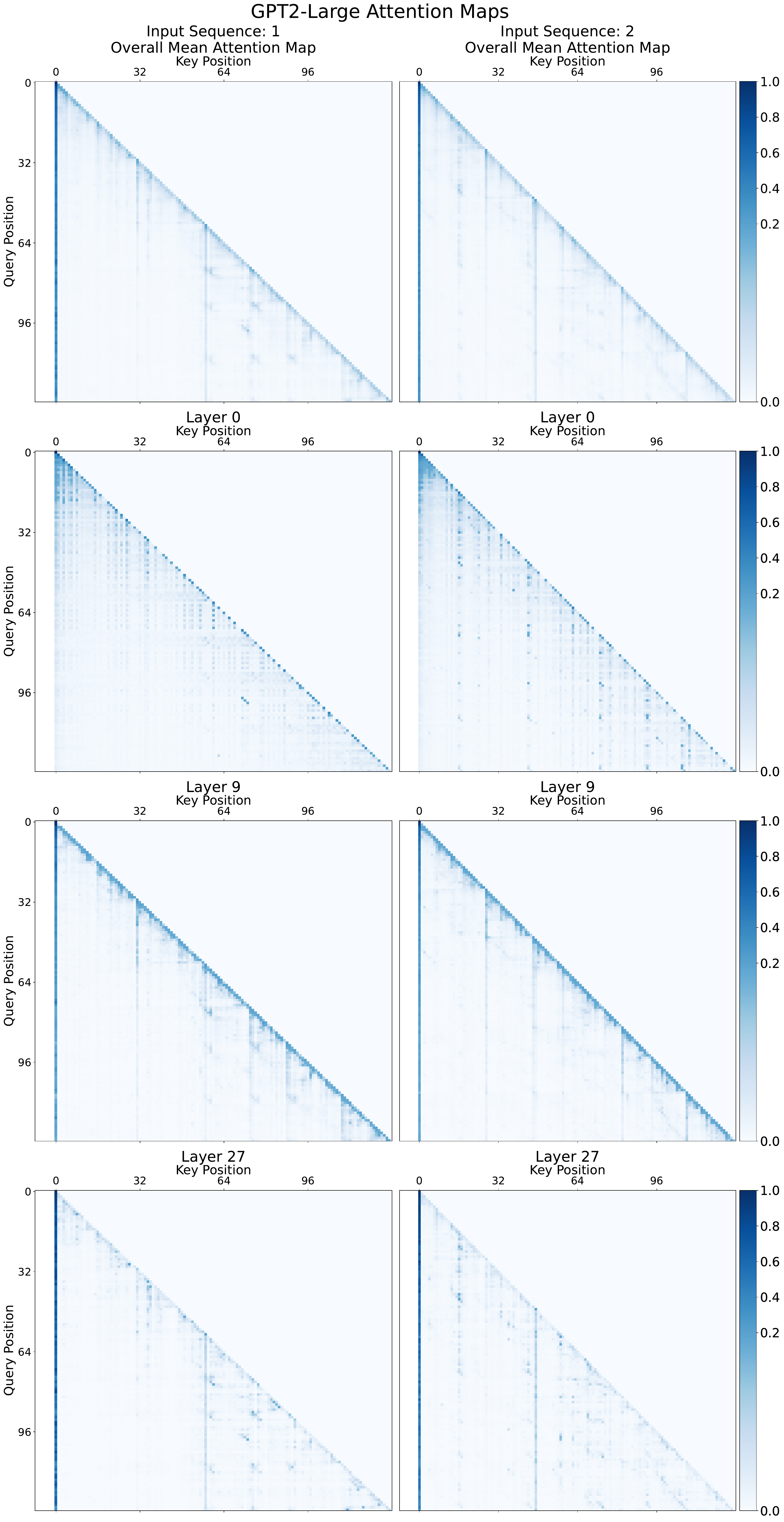}
    \caption{
        Example attention maps for a GPT2-Large model.
    }\label{fig:gpt2-large_attention_maps}
\end{figure}

\begin{figure}[H]
    \centering
    \includegraphics[width=0.8\textwidth]{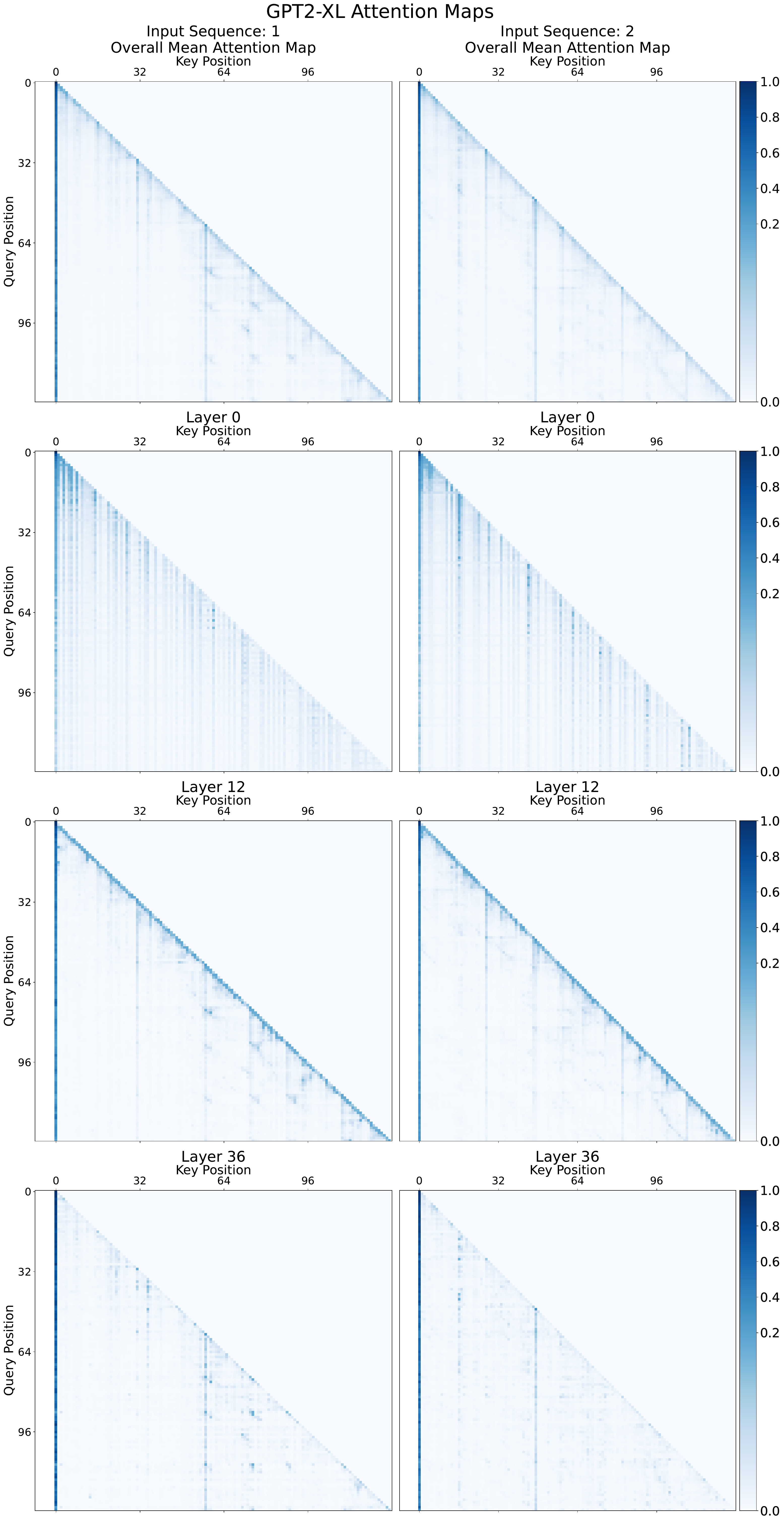}
    \caption{
        Example attention maps for a GPT2-XL model.
    }\label{fig:gpt2-xl_attention_maps}
\end{figure}

\begin{figure}[H]
    \centering
    \includegraphics[width=0.8\textwidth]{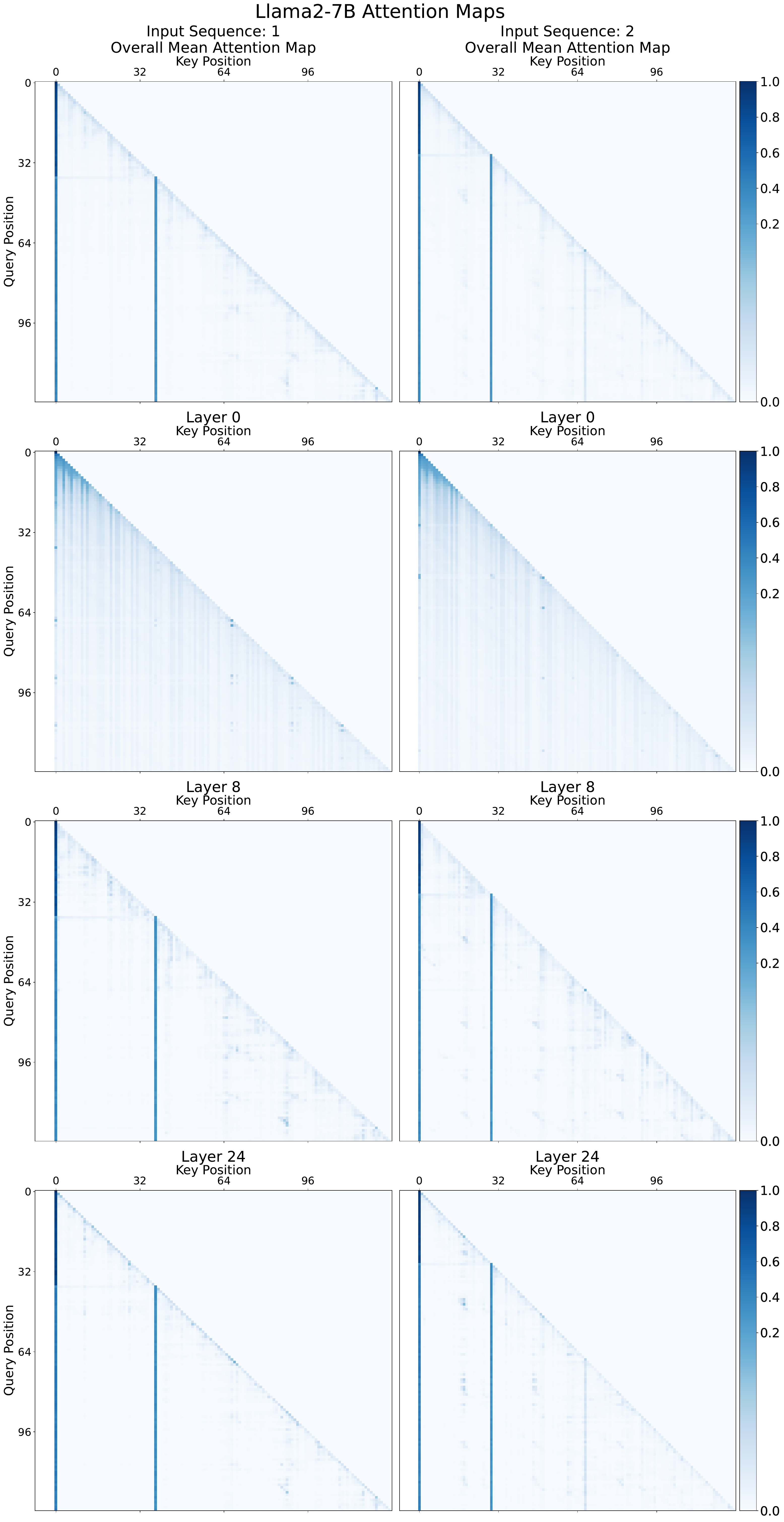}
    \caption{
        Example attention maps for a Llama2-7B model.
    }\label{fig:llama2-7b-hf_attention_maps}
\end{figure}

\begin{figure}[H]
    \centering
    \includegraphics[width=0.8\textwidth]{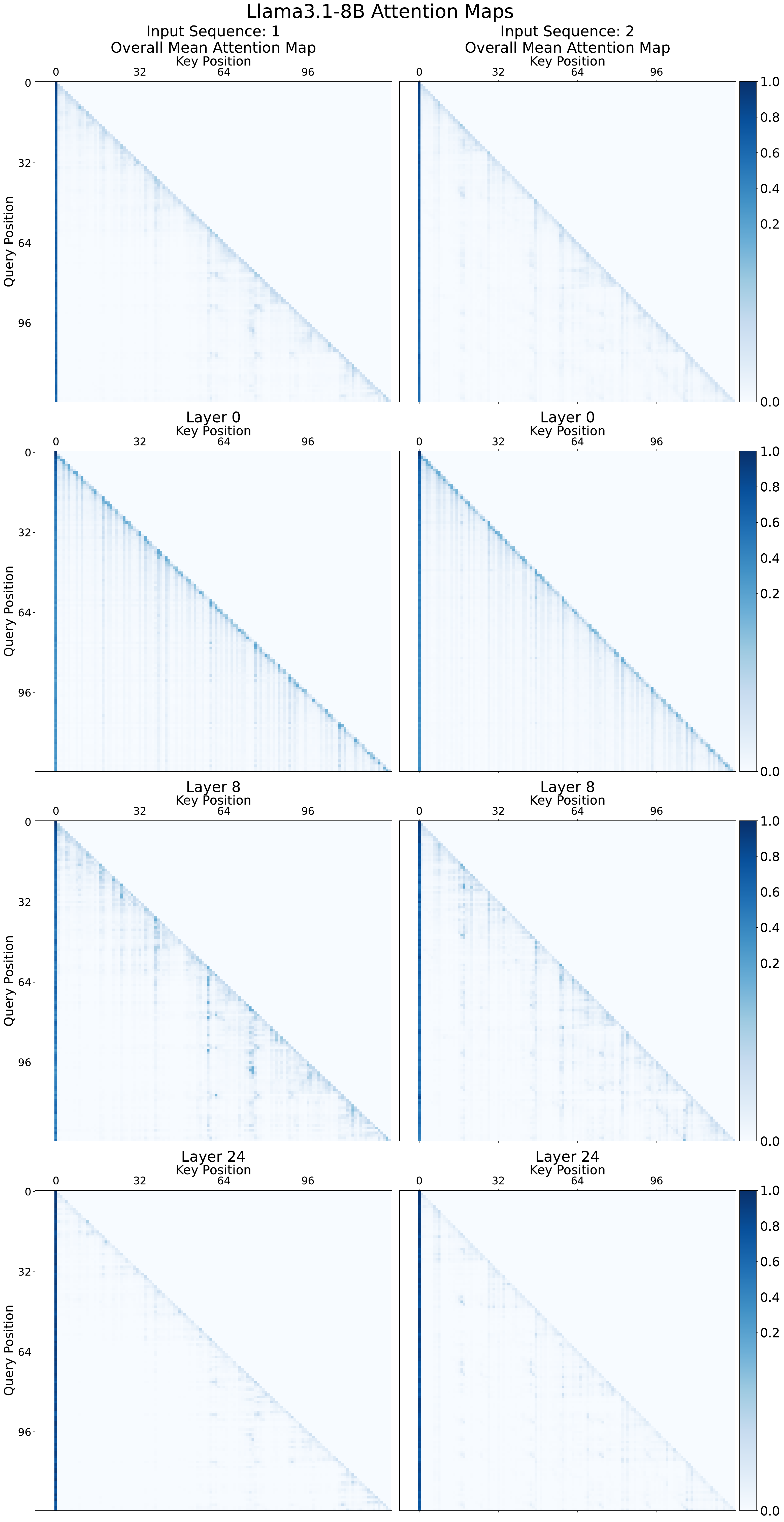}
    \caption{
        Example attention maps for a Llama3.1-8B model.
    }\label{fig:llama3-8b_attention_maps}
\end{figure}

\begin{figure}[H]
    \centering
    \includegraphics[width=0.8\textwidth]{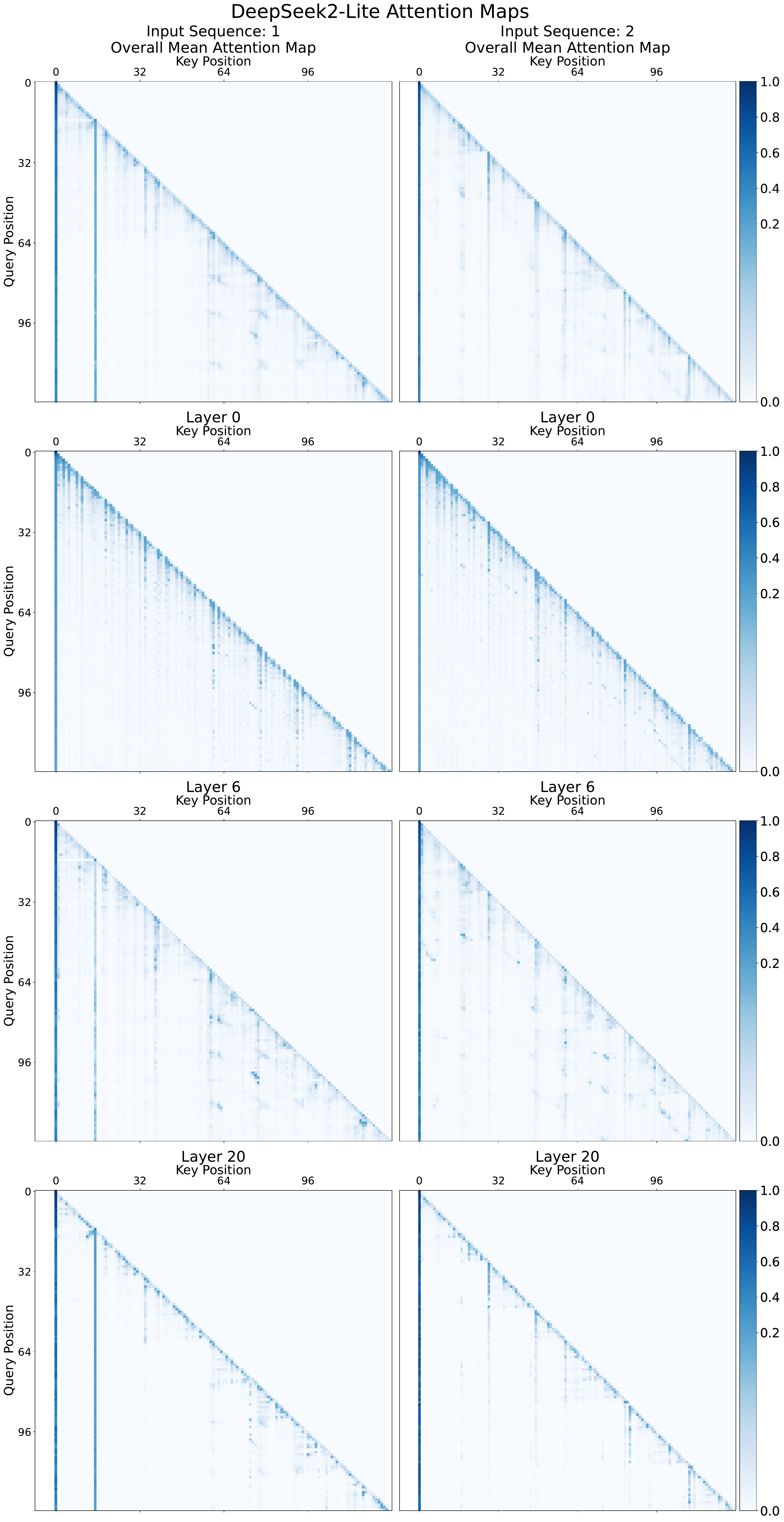}
    \caption{
        Example attention maps for a DeepSeekv2-Lite model.
    }\label{fig:deepseek2-lite_attention_maps}
\end{figure}
\newpage
\section{Training Curves}\label{sec:loss_curves}
To demonstrate that our proposed methods,
\ie~replacing the canonical softmax function with \emph{softmax-1}
and using our proposed optimiser, \emph{OrthoAdam}, do not negatively impact
the training of large language models, we provide the training curves for
our models here.
One can observe that the training curves for models using either or both of our proposed changes
are stable and converge to a similar loss value as the baseline models.

\subsection{GPT2-60M}\label{ssec:loss_gpt2_60m}
\begin{figure}[H]
    \centering
    \includegraphics[width=1.0\textwidth]{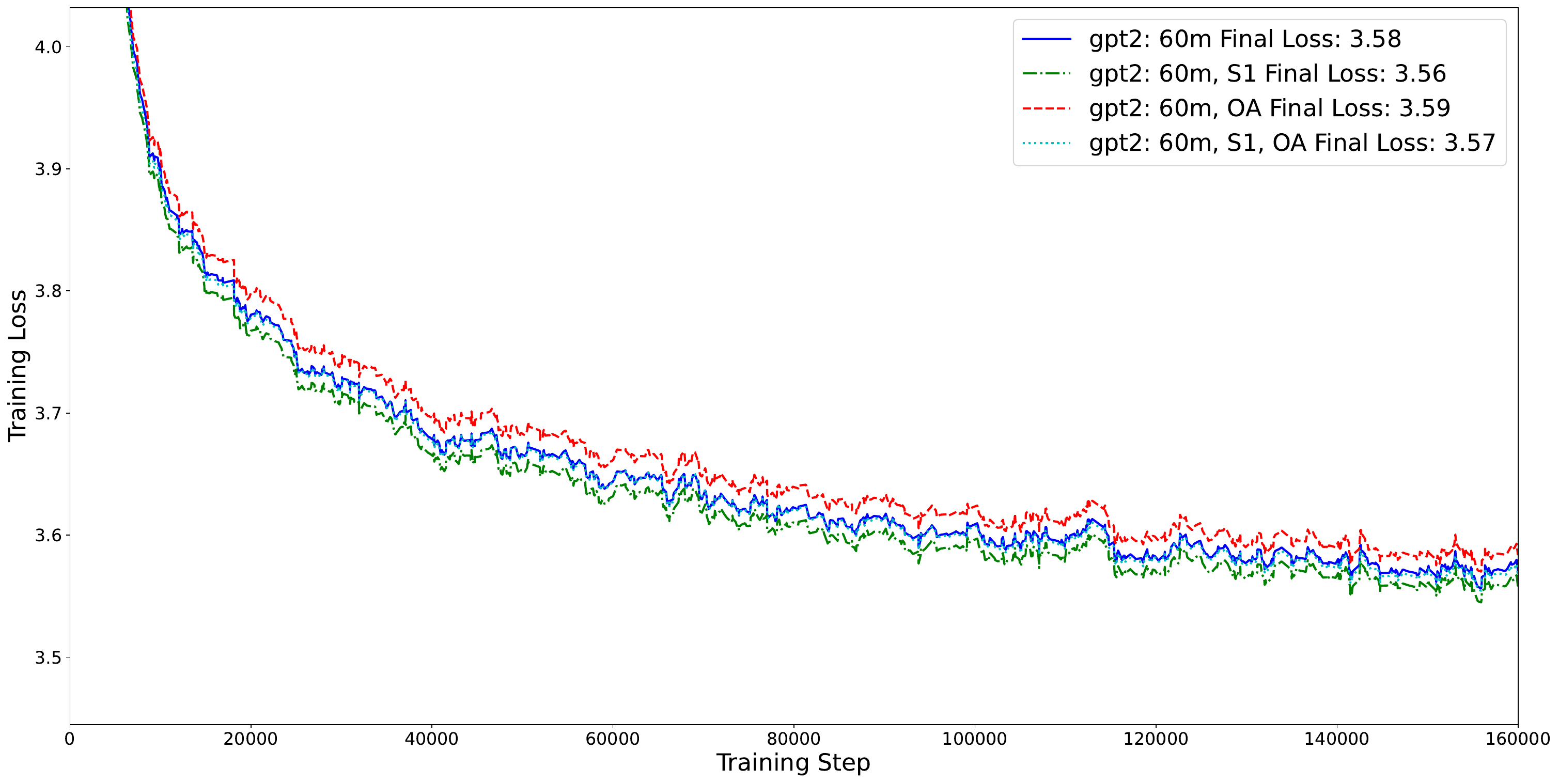}
    \caption{
        Training curves for GPT2-60M models with different optimisers and softmax functions.
        The models using \emph{OrthoAdam} and \emph{softmax-1} are stable and converge to a similar loss value as the baseline models.
        S1/OA denotes the model using softmax-1 and/or OrthoAdam.
    }\label{fig:loss_gpt2_60m}
\end{figure}

\subsection{GPT2-130M}\label{ssec:loss_gpt2_130m}
\begin{figure}[H]
    \centering
    \includegraphics[width=1.0\textwidth]{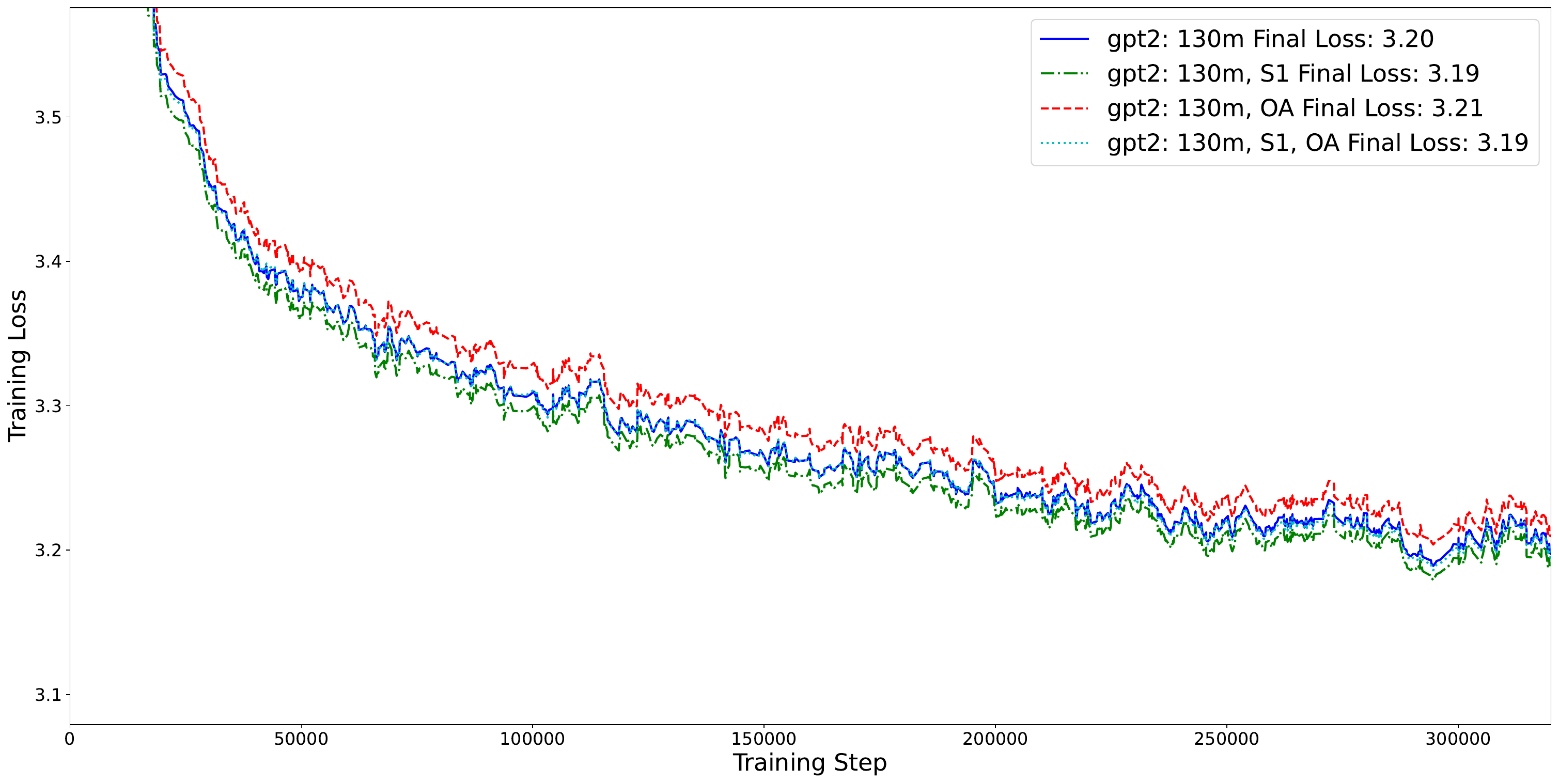}
    \caption{
        Training curves for GPT2-130M models with different optimisers and softmax functions.
        The models using \emph{OrthoAdam} and \emph{softmax-1} are stable and converge to a similar loss value as the baseline models.
        S1/OA denotes the model using softmax-1 and/or OrthoAdam.
    }\label{fig:loss_gpt2_130m}
\end{figure}

\subsection{GPT2-350M and GPT2-1.4B}\label{ssec:loss_gpt2_350m}
\begin{figure}[H]
    \centering
    \includegraphics[width=1.0\textwidth]{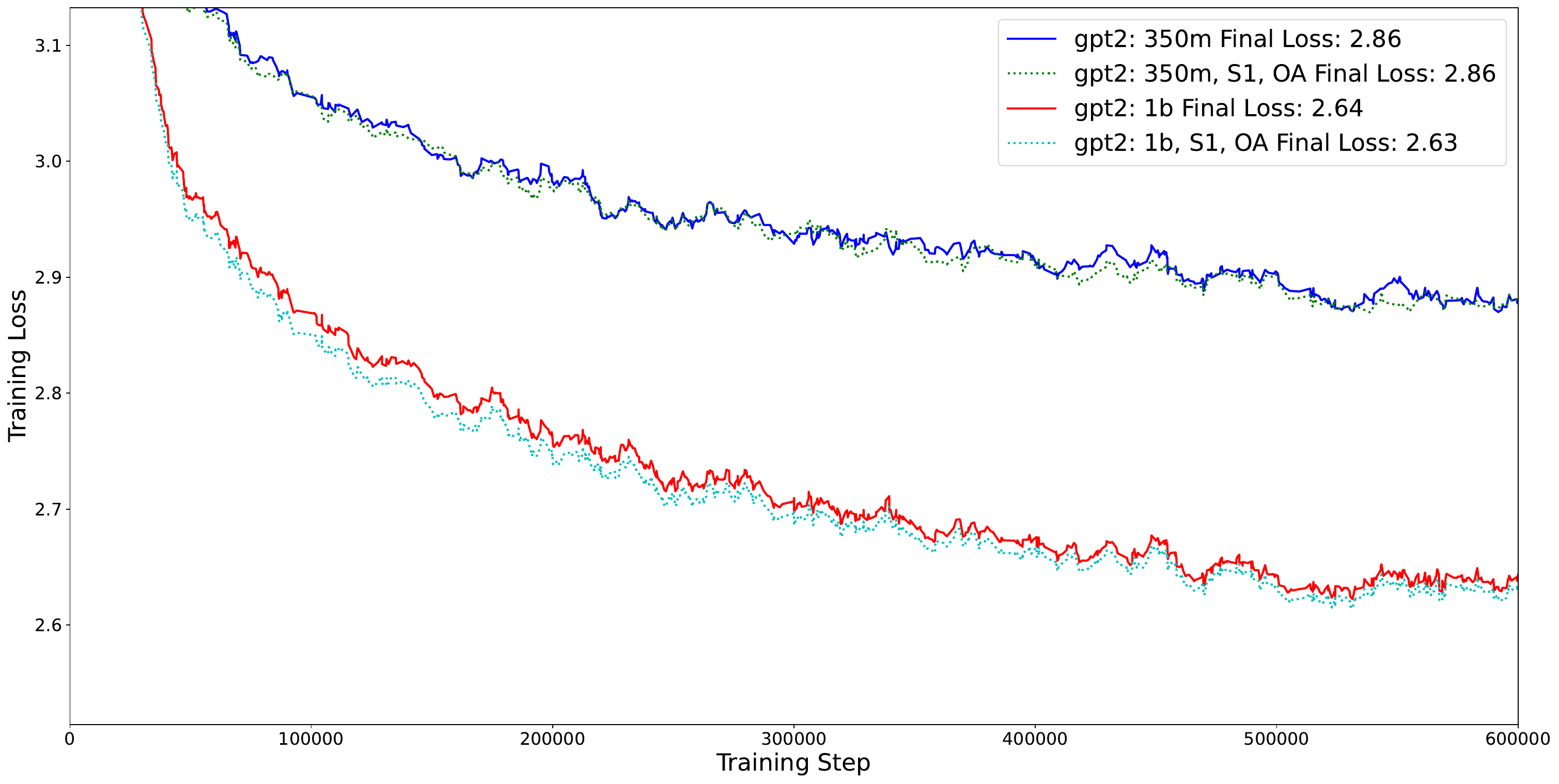}
    \caption{
        Training curves for GPT2-350M and GPT2-1.4B models with different optimisers and softmax functions.
        The models using \emph{OrthoAdam} and \emph{softmax-1} are stable and converge to a similar loss value as the baseline models.
        S1/OA denotes the model using softmax-1 and/or OrthoAdam.
    }\label{fig:loss_gpt2_350m}
\end{figure}

\subsection{Llama-130M}\label{ssec:loss_llama_130m}
\begin{figure}[H]
    \centering
    \includegraphics[width=1.0\textwidth]{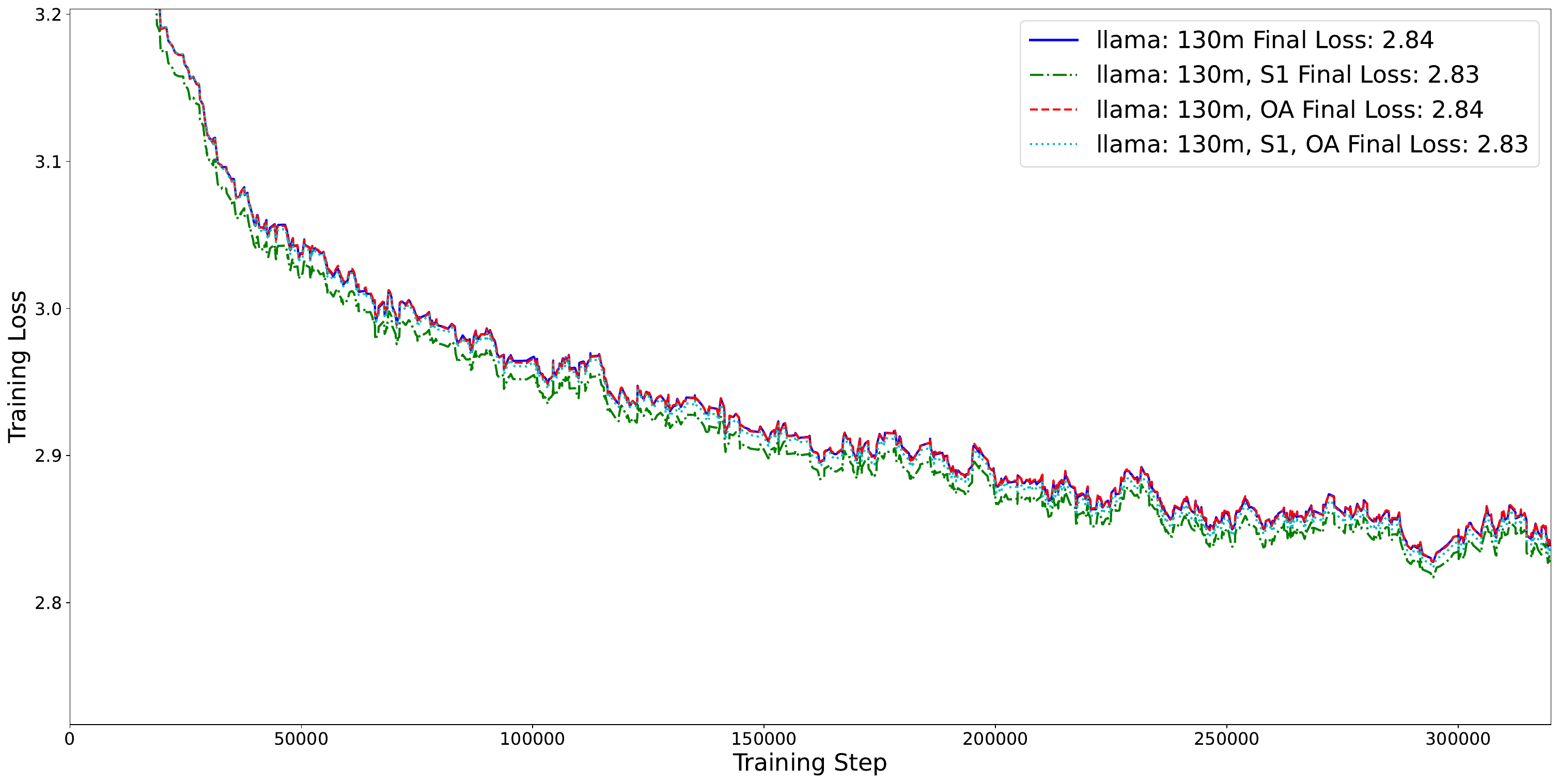}
    \caption{
        Training curves for Llama-130M models with different optimisers and softmax functions.
        The models using \emph{OrthoAdam} and \emph{softmax-1} are stable and converge to a similar loss value as the baseline models.
        S1/OA denotes the model using softmax-1 and/or OrthoAdam.
    }\label{fig:loss_llama_130m}
\end{figure}

\end{document}